\documentclass[10pt,twocolumn,letterpaper]{article}

\PassOptionsToPackage{table,x11names,dvipsnames}{xcolor}

\usepackage{wacv}

\usepackage{subcaption}
\usepackage{pifont}
\usepackage{amsmath}
\usepackage{amssymb}
\usepackage[table,x11names]{xcolor}
\usepackage{tabulary}
\usepackage{colortbl}
\usepackage{booktabs}
\usepackage[font=footnotesize,skip=3pt,subrefformat=parens]{subcaption}
\usepackage{newtxtext}

\usepackage{xfp}
\usepackage{epsfig}
\usepackage{graphicx}
\usepackage{tikz}
\usepackage{pgfplots}
\pgfplotsset{compat=1.18}
\usepackage{pgfplotstable}
\usepackage{caption}
\usepgfplotslibrary{groupplots}
\usetikzlibrary{patterns}

\usepackage{listings}
\usepackage{xcolor}
\usepackage{wrapfig}
\usepackage{tcolorbox}
\usepackage[accsupp]{axessibility}

\definecolor{LGray}{gray}{0.97}
\newcommand{\cmark}{\textcolor{mydarkgreen}{\ding{51}}}
\newcommand{\xmark}{\textcolor{red}{\ding{55}}}

\definecolor{SageGreen}{HTML}{e6f2ce}
\definecolor{DarkSageGreen}{HTML}{5d8412}
\definecolor{DarkRed}{HTML}{a01313}

\definecolor{BackgroundGreen}{HTML}{e6f2ce}
\definecolor{BackgroundPink}{HTML}{f4dcee}
\definecolor{BackgroundBlue}{HTML}{c3e8f4}
\definecolor{BackgroundCoffee}{HTML}{efdabf}
\definecolor{TextGreen}{HTML}{354c08}
\definecolor{TextPink}{HTML}{841c6e}
\definecolor{TextBlue}{HTML}{196577}
\definecolor{TextCoffee}{HTML}{684014}

\definecolor{gt21}{HTML}{c50629}
\definecolor{gt22}{HTML}{3603c9}
\definecolor{gt23}{HTML}{a8f24a}
\definecolor{gt24}{HTML}{f6fa40}
\definecolor{prima22}{HTML}{1804e0}
\definecolor{prima21}{HTML}{f6fa40}
\definecolor{glamm22}{HTML}{1804e0}
\definecolor{glamm21}{HTML}{c50629}
\definecolor{lisa21}{HTML}{84eef4}
\definecolor{lisa22}{HTML}{bc0082}

\definecolor{gt31}{HTML}{f6fa40}
\definecolor{gt32}{HTML}{3603c9}
\definecolor{gt33}{HTML}{a8f24a}
\definecolor{gt34}{HTML}{bc0082}
\definecolor{prima31}{HTML}{84eef4}
\definecolor{prima32}{HTML}{3603c9}
\definecolor{prima33}{HTML}{930f0d}
\definecolor{prima34}{HTML}{6e0d70}
\definecolor{glamm31}{HTML}{bc0082}
\definecolor{glamm32}{HTML}{84eef4}

\definecolor{gt51}{HTML}{3603c9}
\definecolor{gt52}{HTML}{f6fa40}
\definecolor{prima51}{HTML}{c50629}
\definecolor{prima52}{HTML}{3603c9}
\definecolor{glamm51}{HTML}{c50629}
\definecolor{glamm52}{HTML}{3603c9}

\definecolor{gt61}{HTML}{bc0082}
\definecolor{gt62}{HTML}{930f0d}
\definecolor{gt63}{HTML}{1804e0}
\definecolor{prima61}{HTML}{bc0082}
\definecolor{prima62}{HTML}{1804e0}
\definecolor{glamm61}{HTML}{1804e0}
\definecolor{glamm62}{HTML}{c50629}
\definecolor{glamm63}{HTML}{bc0082}
\definecolor{glamm64}{HTML}{3603c9}
\definecolor{lisa61}{HTML}{c50629}
\definecolor{lisa62}{HTML}{930f0d}

\definecolor{gt71}{HTML}{cbfca7}
\definecolor{gt72}{HTML}{bc0082}
\definecolor{prima71}{HTML}{3603c9}
\definecolor{prima72}{HTML}{c50629}
\definecolor{prima73}{HTML}{cbfca7}
\definecolor{prima74}{HTML}{bc0082}
\definecolor{glamm71}{HTML}{f6fa40}
\definecolor{glamm72}{HTML}{bc0082}
\definecolor{glamm73}{HTML}{3603c9}
\definecolor{glamm74}{HTML}{1804e0}
\definecolor{glamm75}{HTML}{930f0d}
\definecolor{glamm76}{HTML}{84eef4}
\definecolor{glamm77}{HTML}{6e0d70}
\definecolor{glamm78}{HTML}{c50629}
\definecolor{glamm79}{HTML}{cbfca7}
\definecolor{lisa71}{HTML}{bc0082}
\definecolor{lisa72}{HTML}{cbfca7}
\definecolor{lisa73}{HTML}{1804e0}
\definecolor{lisa74}{HTML}{3603c9}
\definecolor{lisa75}{HTML}{c50629}
\definecolor{lisa76}{HTML}{f6fa40}
\definecolor{lisa77}{HTML}{bc0082}
\definecolor{lisa78}{HTML}{a8f24a}
\definecolor{lisa79}{HTML}{930f0d}

\definecolor{gt81}{HTML}{930f0d}
\definecolor{gt82}{HTML}{5d0b6d}
\definecolor{gt83}{HTML}{3603c9}
\definecolor{gt84}{HTML}{f6fa40}
\definecolor{prima81}{HTML}{930f0d}
\definecolor{prima82}{HTML}{84eef4}
\definecolor{prima83}{HTML}{1804e0}
\definecolor{prima84}{HTML}{6e0d70}
\definecolor{glamm81}{HTML}{3603c9}
\definecolor{glamm82}{HTML}{cbfca7}
\definecolor{glamm83}{HTML}{f6fa40}
\definecolor{glamm84}{HTML}{a8f24a}
\definecolor{lisa81}{HTML}{c50629}
\definecolor{lisa82}{HTML}{cbfca7}
\definecolor{lisa83}{HTML}{bc0082}

\definecolor{gt91}{HTML}{6e0d70}
\definecolor{gt92}{HTML}{c50629}
\definecolor{gt93}{HTML}{1804e0}
\definecolor{gt94}{HTML}{84eef4}
\definecolor{prima91}{HTML}{a8f24a}
\definecolor{prima92}{HTML}{84eef4}
\definecolor{prima93}{HTML}{1804e0}
\definecolor{prima94}{HTML}{6e0d70}
\definecolor{glamm91}{HTML}{a8f24a}
\definecolor{glamm92}{HTML}{bc0082}
\definecolor{glamm93}{HTML}{3603c9}

\definecolor{gt101}{HTML}{930f0d}
\definecolor{gt102}{HTML}{1804e0}
\definecolor{gt103}{HTML}{c50629}
\definecolor{prima101}{HTML}{bc0082}
\definecolor{prima102}{HTML}{f6fa40}
\definecolor{prima103}{HTML}{cbfca7}
\definecolor{glamm101}{HTML}{c50629}
\definecolor{glamm102}{HTML}{1804e0}
\definecolor{glamm103}{HTML}{a8f24a}
\definecolor{glamm104}{HTML}{930f0d}
\definecolor{lisa101}{HTML}{6e0d70}
\definecolor{lisa102}{HTML}{c50629}
\definecolor{lisa103}{HTML}{cbfca7}

\definecolor{gt111}{HTML}{1804e0}
\definecolor{gt112}{HTML}{bc0082}
\definecolor{gt113}{HTML}{f6fa40}
\definecolor{prima111}{HTML}{84eef4}
\definecolor{prima112}{HTML}{6e0d70}
\definecolor{prima113}{HTML}{cbfca7}
\definecolor{glamm111}{HTML}{6e0d70}
\definecolor{glamm112}{HTML}{1804e0}
\definecolor{glamm113}{HTML}{cbfca7}
\definecolor{glamm114}{HTML}{930f0d}
\definecolor{lisa111}{HTML}{1804e0}
\definecolor{lisa112}{HTML}{bc0082}
\definecolor{lisa113}{HTML}{6e0d70}
\definecolor{lisa114}{HTML}{3603c9}
\definecolor{lisa115}{HTML}{c50629}
\definecolor{lisa116}{HTML}{cbfca7}
\definecolor{lisa117}{HTML}{a8f24a}
\definecolor{lisa118}{HTML}{5d0b6d}

\newcommand{\hallu}[1]{\colorbox{black}{\color{white} #1}}

\definecolor{data11}{HTML}{84eef4}
\definecolor{data21}{HTML}{1804e0}
\definecolor{data22}{HTML}{bc0082}
\definecolor{data23}{HTML}{930f0d}
\definecolor{data24}{HTML}{f6fa40}
\definecolor{data31}{HTML}{cbfca7}
\definecolor{data32}{HTML}{bc0082}
\definecolor{data41}{HTML}{f6fa40}
\definecolor{data42}{HTML}{c50629}
\definecolor{data43}{HTML}{5d0b6d}
\definecolor{data44}{HTML}{930f0d}
\definecolor{data45}{HTML}{cbfca7}
\definecolor{data51}{HTML}{930f0d}
\definecolor{data52}{HTML}{3603c9}
\definecolor{data53}{HTML}{5d0b6d}
\definecolor{data54}{HTML}{a8f24a}
\definecolor{data55}{HTML}{bc0082}
\definecolor{data56}{HTML}{c50629}
\definecolor{data57}{HTML}{84eef4}
\definecolor{data61}{HTML}{1804e0}
\definecolor{data62}{HTML}{5d0b6d}
\definecolor{data71}{HTML}{f6fa40}
\definecolor{data72}{HTML}{c50629}
\definecolor{data73}{HTML}{5d0b6d}
\definecolor{data81}{HTML}{c50629}
\definecolor{data82}{HTML}{cbfca7}
\definecolor{data83}{HTML}{bc0082}
\definecolor{data84}{HTML}{f6fa40}
\definecolor{data85}{HTML}{1804e0}

\definecolor{data91}{HTML}{bc0082}
\definecolor{data92}{HTML}{f6fa40}
\definecolor{data101}{HTML}{5d0b6d}
\definecolor{data102}{HTML}{f6fa40}
\definecolor{data103}{HTML}{84eef4}
\definecolor{data111}{HTML}{1804e0}
\definecolor{data112}{HTML}{f6fa40}
\definecolor{data113}{HTML}{6e0d70}
\definecolor{data121}{HTML}{bc0082}
\definecolor{data122}{HTML}{84eef4}
\definecolor{data123}{HTML}{c50629}
\definecolor{data131}{HTML}{1804e0}
\definecolor{data132}{HTML}{c50629}
\definecolor{data133}{HTML}{5d0b6d}
\definecolor{data134}{HTML}{84eef4}
\definecolor{data135}{HTML}{f6fa40}
\definecolor{data136}{HTML}{cbfca7}
\definecolor{data141}{HTML}{c50629}
\definecolor{data142}{HTML}{bc0082}
\definecolor{data151}{HTML}{930f0d}
\definecolor{data152}{HTML}{a8f24a}
\definecolor{data153}{HTML}{f6fa40}
\definecolor{data161}{HTML}{cbfca7}
\definecolor{data162}{HTML}{930f0d}
\definecolor{data163}{HTML}{f6fa40}
\definecolor{data164}{HTML}{1804e0}
\definecolor{data165}{HTML}{5d0b6d}
\definecolor{data166}{HTML}{6e0d70}
\definecolor{data167}{HTML}{84eef4}
\definecolor{data168}{HTML}{3603c9}
\definecolor{data169}{HTML}{bc0082}

\definecolor{data171}{HTML}{1804e0}
\definecolor{data191}{HTML}{1804e0}
\definecolor{data192}{HTML}{a8f24a}
\definecolor{data193}{HTML}{6e0d70}
\definecolor{data194}{HTML}{930f0d}
\definecolor{data201}{HTML}{c50629}
\definecolor{data202}{HTML}{5d0b6d}
\definecolor{data203}{HTML}{3603c9}
\definecolor{data204}{HTML}{bc0082}
\definecolor{data211}{HTML}{84eef4}
\definecolor{data212}{HTML}{1804e0}
\definecolor{data213}{HTML}{f6fa40}
\definecolor{data214}{HTML}{3603c9}

\definecolor{data221}{HTML}{3603c9}
\definecolor{data222}{HTML}{a8f24a}
\definecolor{data223}{HTML}{bc0082}
\definecolor{data231}{HTML}{bc0082}
\definecolor{data232}{HTML}{1804e0}
\definecolor{data233}{HTML}{a8f24a}
\definecolor{data234}{HTML}{c50629}
\definecolor{data235}{HTML}{3603c9}
\definecolor{data251}{HTML}{bc0082}
\definecolor{data252}{HTML}{6e0d70}
\definecolor{data253}{HTML}{84eef4}
\definecolor{data254}{HTML}{1804e0}
\definecolor{data261}{HTML}{6e0d70}
\definecolor{data262}{HTML}{f6fa40}
\definecolor{data263}{HTML}{3603c9}
\definecolor{data264}{HTML}{bc0082}
\definecolor{data265}{HTML}{930f0d}

\definecolor{primaitw11}{HTML}{1804e0}
\definecolor{primaitw12}{HTML}{a8f24a}
\definecolor{glammitw11}{HTML}{84eef4}
\definecolor{glammitw12}{HTML}{bc0082}
\definecolor{lisaitw11}{HTML}{930f0d}
\definecolor{lisaitw12}{HTML}{1804e0}
\definecolor{lisaitw13}{HTML}{3603c9}
\definecolor{lisaitw14}{HTML}{a8f24a}

\definecolor{primaitw31}{HTML}{a8f24a}
\definecolor{primaitw32}{HTML}{6e0d70}
\definecolor{glammitw31}{HTML}{3603c9}
\definecolor{glammitw32}{HTML}{6e0d70}
\definecolor{lisaitw31}{HTML}{5d0b6d}
\definecolor{lisaitw32}{HTML}{cbfca7}

\definecolor{primaitw41}{HTML}{1804e0}
\definecolor{primaitw42}{HTML}{cbfca7}
\definecolor{glammitw41}{HTML}{3603c9}
\definecolor{glammitw42}{HTML}{bc0082}
\definecolor{lisaitw41}{HTML}{a8f24a}
\definecolor{lisaitw42}{HTML}{1804e0}

\definecolor{primaitw51}{HTML}{a8f24a}
\definecolor{glammitw51}{HTML}{84eef4}
\definecolor{glammitw52}{HTML}{1804e0}
\definecolor{lisaitw51}{HTML}{bc0082}
\definecolor{lisaitw52}{HTML}{84eef4}

\definecolor{primaitw61}{HTML}{f6fa40}
\definecolor{primaitw62}{HTML}{84eef4}
\definecolor{glammitw61}{HTML}{84eef4}
\definecolor{glammitw62}{HTML}{f6fa40}
\definecolor{glammitw63}{HTML}{c50629}
\definecolor{lisaitw61}{HTML}{a8f24a}
\definecolor{lisaitw62}{HTML}{84eef4}
\definecolor{lisaitw63}{HTML}{bc0082}

\newcommand{\modelname}{\textsc{Prima}}
\newcommand{\benchmarkname}{\textsc{M$^4$Seg}}
\newcommand{\benchmarknamebold}{\textbf{\textsc{M$\mathbf{^4}$Seg}}}
\newcommand{\modulename}{\textsc{SQuARE}}
\usepackage{xspace}
\usepackage{multirow}
\newcommand{\R}{\mathbb{R}}
\newcommand{\cc}{\cellcolor{SageGreen}}
\newcommand{\gc}{\cellcolor{LGray}}
\definecolor{mydarkgreen}{rgb}{0.0, 0.5, 0.0}

\usepackage{tikz}
\usepackage{booktabs}
\usepackage{tabularx}
\usepackage{xcolor}

\definecolor{stat1a}{HTML}{F5A261}
\definecolor{stat1b}{HTML}{FF8A3D}
\definecolor{stat1c}{HTML}{F06A00}

\definecolor{stat2a}{HTML}{FFB45F}
\definecolor{stat2b}{HTML}{FF7A1A}
\definecolor{stat2c}{HTML}{E95C00}
\definecolor{stat2d}{HTML}{CC4F00}

\definecolor{stat3a}{HTML}{EF3340}
\definecolor{stat3b}{HTML}{DD2432}
\definecolor{stat3c}{HTML}{CD1727}
\definecolor{stat3d}{HTML}{BC0F22}
\definecolor{stat3e}{HTML}{A9081D}
\definecolor{stat3f}{HTML}{930015}

\definecolor{stat4a}{HTML}{B18BD0}
\definecolor{stat4b}{HTML}{9A68C1}
\definecolor{stat4c}{HTML}{844DB2}
\definecolor{stat4d}{HTML}{7136A2}
\definecolor{stat4e}{HTML}{5E248F}

\usepackage{xcolor}
\usepackage{amssymb}
\usepackage{graphicx}
\usepackage{fdsymbol}

\definecolor{illinoisorange}{HTML}{FF5F05}
\definecolor{vtmaroon}{HTML}{861F41}

\newcommand{\uiuc}{\textsuperscript{
  \textcolor{illinoisorange}{$\blacklozenge$}}}
\newcommand{\vt}{\textsuperscript{
  \textcolor{vtmaroon}{$\spadesuit$}}}

\definecolor{wacvblue}{rgb}{0.21,0.49,0.74}
\usepackage[breaklinks,colorlinks,allcolors=wacvblue]{hyperref}
\usepackage{orcidlink}

\usepackage{booktabs}
\usepackage{tabularx}
\usepackage{siunitx}

\def\wacvPaperID{1729}
\def\confName{WACV}
\def\confYear{2027}

\begin{document}

\title{\includegraphics[height=1cm]{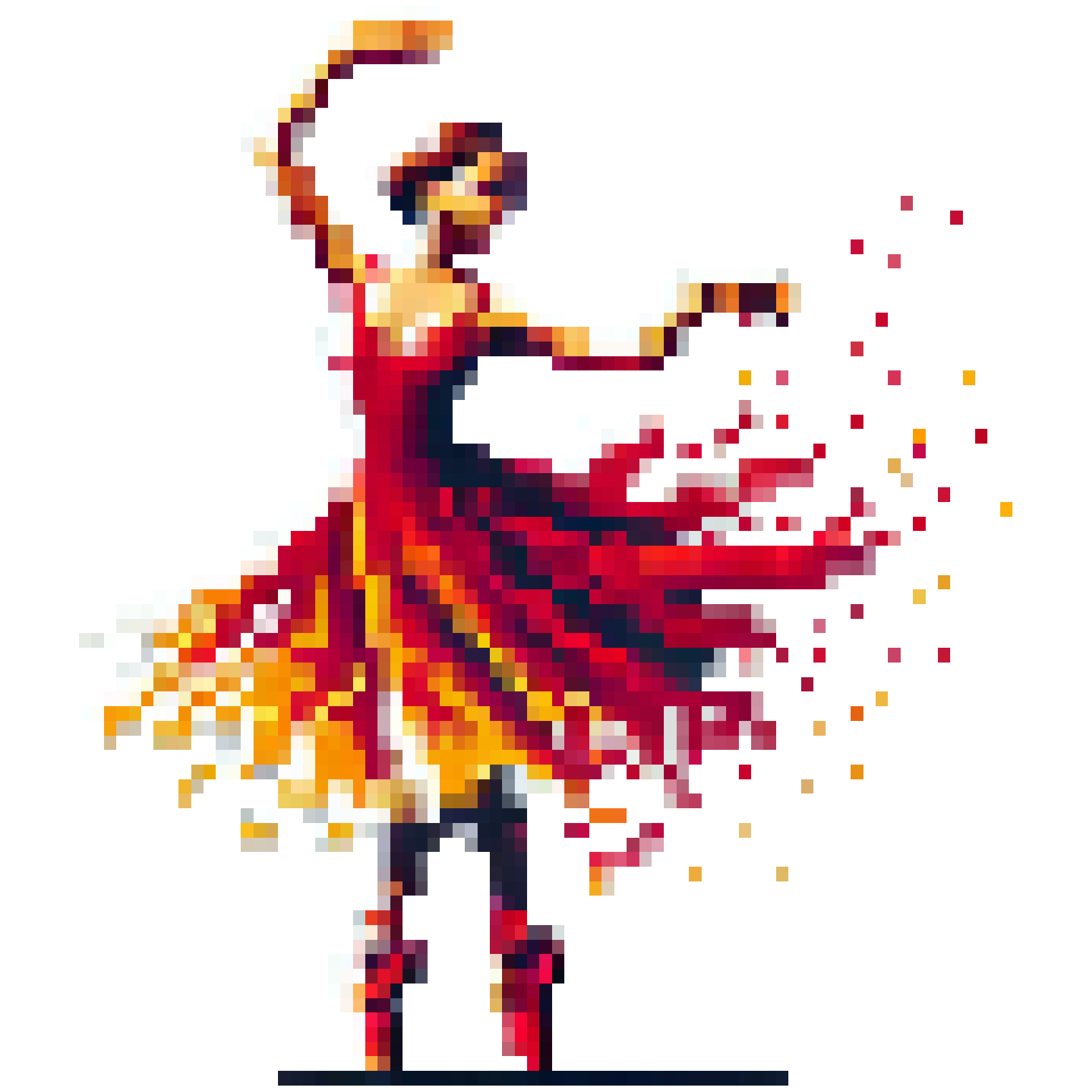}\modelname{}: Multi-Image Vision-Language Models for Reasoning Segmentation}
\author{
Kiet A. Nguyen\uiuc \quad
Muntasir Wahed\uiuc \quad
Adheesh Sunil Juvekar\uiuc \quad
Xinzhuo Li\uiuc \\
Xiaona Zhou\uiuc \quad
Vedant Shah\uiuc \quad
Tianjiao Yu\uiuc \quad
Pinar Yanardag\vt \quad
Ismini Lourentzou\uiuc \\[2pt]
\uiuc University of Illinois Urbana-Champaign \quad \vt Virginia Tech \\
{\tt\small
\{kietan2, lourent2\}@illinois.edu
}\\[2pt]
\href{https://plan-lab.github.io/prima}{\textcolor{illinoisorange}{plan-lab.github.io/prima}}
}

\twocolumn[{
\renewcommand\twocolumn[1][]{#1}
\maketitle
\begin{center}
\vspace{-0.8cm}
    \centering
    \includegraphics[width=0.95\linewidth]{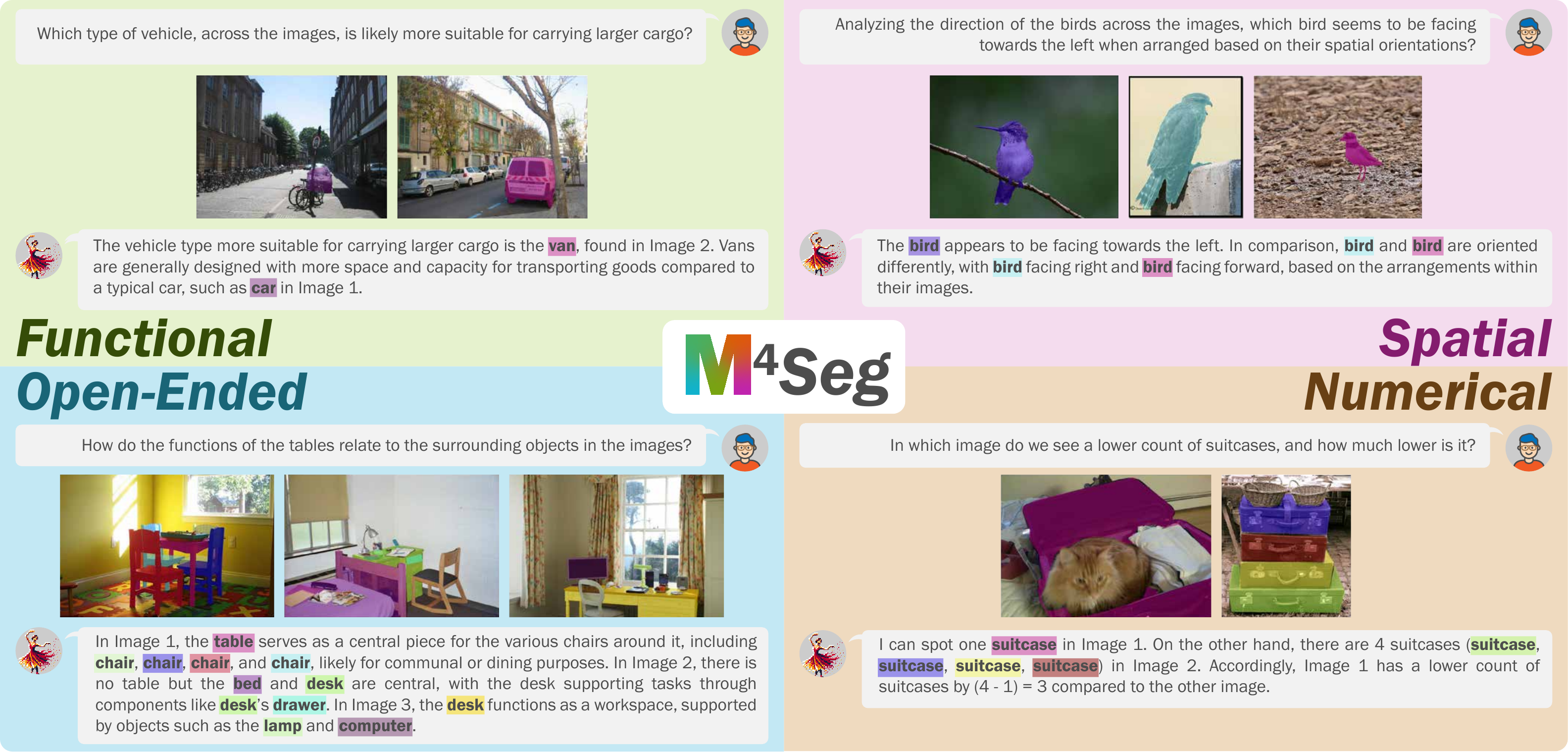}
    \vspace{-0.1cm}
    \captionof{figure}{\textbf{We introduce \emph{multi-image pixel-grounded reasoning}}, a new task that requires LVLMs to answer questions over sets of images with pixel-level grounding. To support this task, we curate \benchmarkname{}, a benchmark comprising $\sim$750K question-answer (QA) pairs over multi-image sets with corresponding segmentation masks. As shown, examples from \benchmarkname{} span functional, spatial, open-ended, and numerical reasoning; matching colors link answer mentions to their segmentation masks. Additionally, we propose \modelname{}, a model trained on \benchmarkname{} to efficiently localize and compare objects and their contextual relationships across scenes.\looseness-1}
    \label{fig1}
\end{center}
}]

\begin{abstract}
Despite significant advancements in Large Vision-Language Models (LVLMs)’ capabilities, existing pixel-grounding models primarily operate in single-image settings, limiting their ability to perform detailed, fine-grained comparisons across multiple images. Conversely, current multi-image understanding models lack pixel-level grounding. We introduce the task of \emph{multi-image pixel-grounded reasoning} alongside \textbf{\modelname{}}, an LVLM that integrates reasoning segmentation with robust multi-image reasoning to produce contextually rich, pixel-grounded explanations. Central to \modelname{} is \modulename{}, a vision module that injects cross-image relational context into compact query-based visual tokens before fusing them with the language backbone. To support training and evaluation, we curate \benchmarknamebold{}, a new multi-image reasoning segmentation benchmark consisting of $\sim$750K question–answer pairs that require fine-grained visual understanding across multiple images. \modelname{} outperforms state-of-the-art baselines, yielding absolute gains of 11.8 and 12.2 points in Recall and S-IoU, respectively. Ablations further demonstrate \modulename{}'s effectiveness in capturing cross-image relationships.
\end{abstract}

\section{Introduction}\label{sec:intro}

Large Vision-Language Models (LVLMs) have demonstrated strong performance across a wide range of perception and reasoning tasks, \eg question answering~\cite{liu2024visual,li2026toward, susladkar2026best}, image captioning~\cite{li2023blip, wang2022ofa}, object detection~\cite{cheng2401yolo}, spatial reasoning \cite{shen2026fine,ogunleye20263d} \etc, due to their ability to jointly understand and process visual and textual information. The synergy between vision and language has also paved the way for more sophisticated image understanding capabilities, including localized reasoning~\cite{zhang2023gpt4roi}, semantic segmentation~\cite{li2024semantic}, reasoning segmentation~\cite{lai2024lisa,li2026counterfactual}, \etc These advancements have enabled precise pixel-level grounding and reasoning referring to objects within an image, which improves both the quality and the explainability of the generated response.\looseness-1

Despite these advances in single-image visual perception, fine-grained comparative analysis across images remains an open challenge, especially in scenarios where understanding subtle differences or similarities between images is crucial~\cite{liu2025omnidiff,wu2026micon,qiao2025multiple}. Unlike the single-image setting, multi-image reasoning segmentation requires models to understand visual co-references across scenes while producing segmentation masks that align with natural language descriptions. 
For example, depending on the task, the model must identify varied fine-grained differences in visual scenes, such as the presence and functions of objects, compare objects in terms of spatial arrangements and ground fine-grained functionalities across diverse contexts (\Cref{fig1}).
This capability supports a wide range of applications, from detailed visual analysis of medical images for disease progression to e-commerce and remote sensing~\cite{karwande2022chexrelnet,mbakwe2023hierarchical,wu2021chest,pippi2025tofu,crall2013hotspotter,zhu2025change3d}. Furthermore, fine-grained cross-image comparisons with pixel grounding can enhance interpretability with localized visual evidence.\looseness-1

\begin{figure}[t!]
    \centering
        \includegraphics[width=0.99\linewidth]{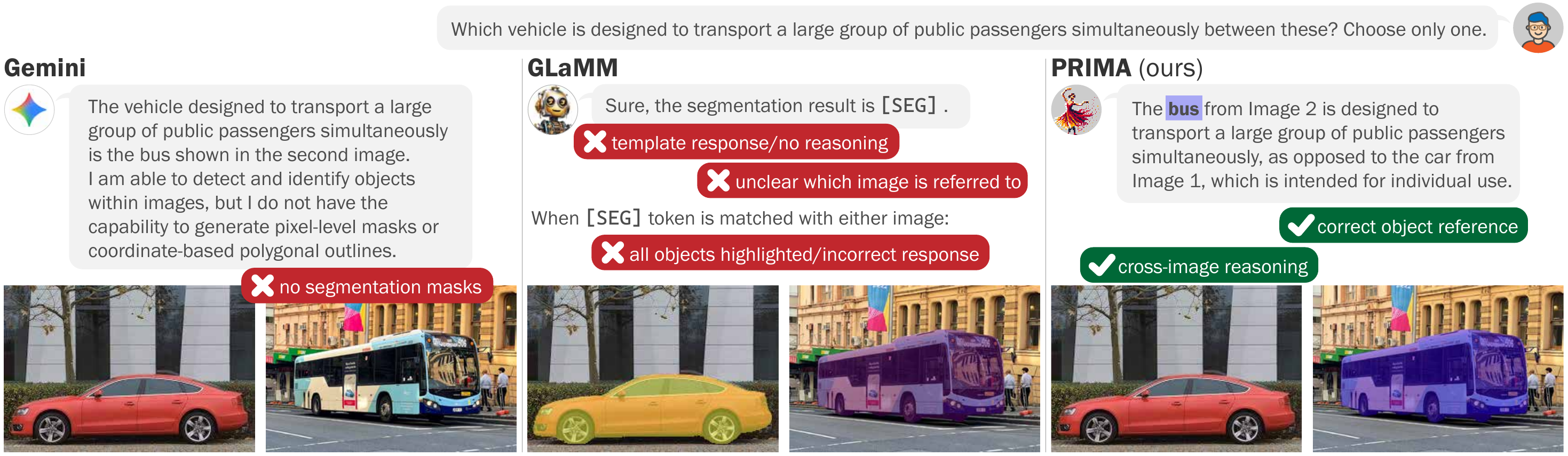} 
    
    \vspace{-0.2cm}
    \caption{\textbf{The challenges of multi-image pixel-grounded reasoning}. Traditional LVLMs (\eg Gemini) lack pixel-level grounding, while single-image reasoning segmentation models
(\eg GLaMM~\cite{rasheed2024glamm}) cannot reason across images. 
\modelname{} performs cross-image reasoning segmentation.
}
\label{fig:motivation}
\vspace{-0.2cm}
\end{figure}

Existing LVLM-based pixel-grounded reasoning~\cite{lai2024lisa, rasheed2024glamm} aligns text with regions or pixels but largely targets single images, limiting coherent multi-image reasoning. Conversely, multi-image understanding methods~\cite{huang2024sparkles,li2024fine} lack the pixel-level grounding needed for fine-grained analysis. Video LVLMs~\cite{lin2025glus, munasinghe2025videoglamm, zheng2025villa, yuan2025sa2va, wei2025hyperseg} relate frames through temporal compression~\cite{zhang2023video,li2024llama} or timestamp-aware encoders~\cite{fateh2024video}, but are designed for temporal continuity rather than pixel-level grounding across independent, unordered images without temporal or causal dependencies.\looseness-1

To bridge the gap between multi-image understanding and pixel-level grounding, we introduce \textbf{\textit{multi-image pixel-grounded reasoning}}, a new task in which models must reason across multiple images at a fine-grained level and produce a natural-language response accompanied by segmentation masks for the relevant instances. As shown in \Cref{fig:motivation}, this task presents challenges current methods struggle to address, including capturing semantic correspondences across disjoint visual contexts while preserving precise localization of potentially disparate object instances.

\begin{figure}[t]
    \centering
    \includegraphics[width=0.85\linewidth]{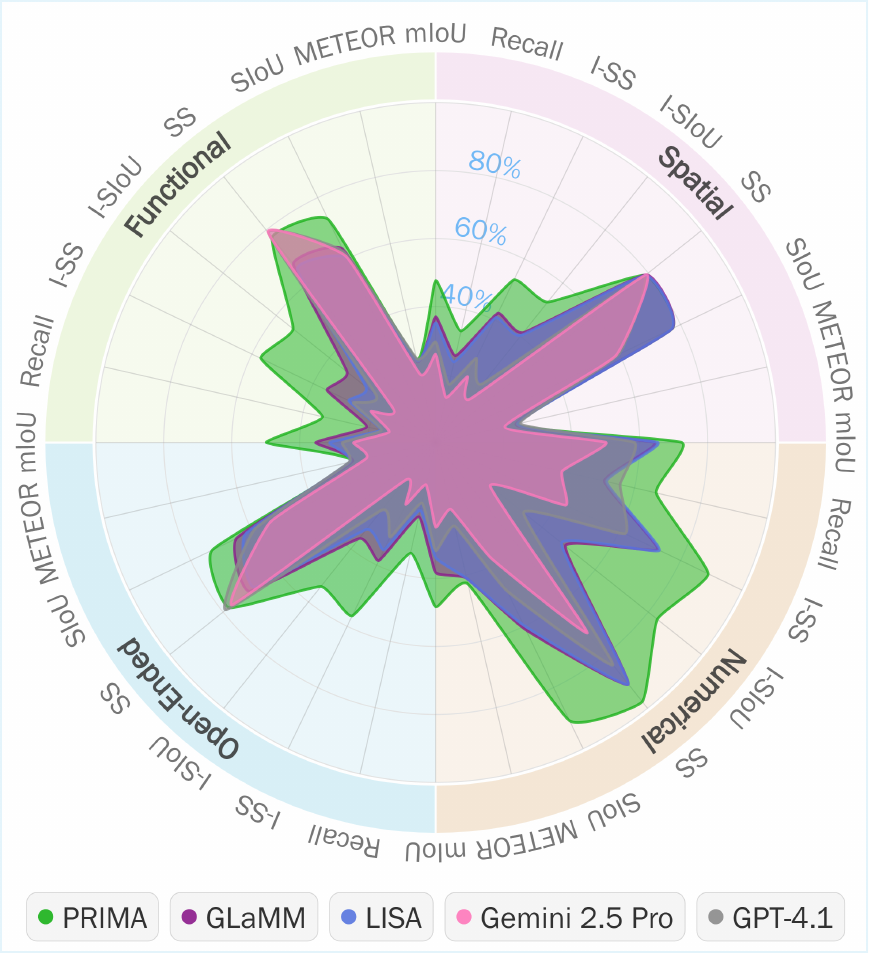}
        \vspace{-0.2cm}
    \caption{\textbf{Performance comparison across different types.} \modelname{} consistently outperforms strong baselines.}
    \label{fig:radar_results}
    \vspace{-0.3cm}
\end{figure}

To facilitate training and evaluation, we introduce \benchmarknamebold{}, a new dataset consisting of $\sim$750K questions derived from high-quality annotated datasets~\cite{zhou2017scene,wei2024ov,everingham2010pascal,chen2014detect,lin2014microsoft,ramanathan2023paco}.
\benchmarkname{} contains 207 unique objects and 248 unique parts, with QA pairs featuring multiple cross-image target segmentation masks, making it a challenging benchmark for multi-image pixel-grounded reasoning. 
Additionally, we propose \textbf{\modelname{}}, the first
\textbf{\underline{P}}ixel-g\textbf{\underline{R}}ounded Multi-\textbf{\underline{I}}mage Seg\textbf{\underline{M}}entation Re\textbf{\underline{A}}soning LVLM for this task. \modelname{} employs \modulename{}, a novel instruction-guided module to combine single- and multi-image understanding in a hierarchical manner.
In summary, the contributions of this work are:\looseness-1
\begin{itemize}
    \item \textbf{New Task:} We introduce the novel task of multi-image pixel-grounded reasoning, which requires fine-grained comparison and contextual understanding across multiple images at pixel level and natural language reasoning.
    \item \textbf{New Dataset:} We curate \benchmarkname{}, a new challenging benchmark with $\sim$750K multi-image QA pairs, annotated with multiple object and part segmentation masks to train and evaluate multi-image pixel-grounding models.
    \item \textbf{New Model:}
    We propose \modelname{}, an architecture designed for reasoning with contextually grounded segmentation masks across multiple images. \modelname{} enables instruction-guided cross-image alignment of relevant visual features via a novel \modulename{} module.
    Experiments (\Cref{fig:radar_results}) demonstrate \modelname{}’s impressive performance compared to strong baselines across segmentation (\textcolor{DarkSageGreen}{\textbf{+}$\uparrow$}$10.8$ points in mIoU, \textcolor{DarkSageGreen}{\textbf{+}$\uparrow$}$11.8$ Recall) and text-based metrics (\textcolor{DarkSageGreen}{\textbf{+}$\uparrow$}$6.1$ Semantic Similarity, \textcolor{DarkSageGreen}{\textbf{+}$\uparrow$}$12.2$ S-IoU).\looseness-1 
\end{itemize}

\section{Related Work}
LVLMs have advanced multimodal reasoning by integrating powerful visual encoders with LLMs~\cite{alayrac2022flamingo, li2023blip, liu2024visual, bai2025qwen2vl}. Prior work follows two largely separate directions: ungrounded multi-image reasoning models and pixel-grounded single-image LVLMs. Bridging these directions is non-trivial: reasoning across disjoint images while producing mask-level grounding requires models to establish cross-image semantic correspondences, compare instances across scenes, and localize the relevant entities with pixel-level accuracy. We address this challenge by introducing the task of \textit{multi-image pixel-grounded reasoning}.

\noindent \textbf{Multi-Image Understanding in LVLMs.}
Recent LVLM research extends single-image models to multi-image understanding settings through improved data curation, input formatting, and scaling strategies
\cite{huang2024sparkles, lin2025comparison, jia2025leopard,li2024fine}.
Scalable variants
\cite{li2024llava, jiang2024mantis, ye2025mplug, wang2024longllava}
expand input capacity to longer or heterogeneous image sequences. Recent benchmarks (Table \ref{tab:dataset_summary})
formalize evaluation for multi-image reasoning. However, these works focus on text-level outputs without explicit localized supervision; \modelname{} instead addresses multi-image reasoning with mask-level grounding.

\noindent \textbf{LVLM-Based Single-Image Segmentation Reasoning.}
Early LVLM-based localization approaches focused on coarse grounding via bounding boxes or pointing mechanisms
\cite{wang2022ofa, peng2024kosmos, chen2023shikra, you2024ferret}.
Recent methods move toward pixel-level grounding by integrating segmentation decoders with language supervision
\cite{zhang2025psalm, zhang2024omg, zhang2024next, zhang2024groundhog}.
These models,
along with referring segmentation frameworks like LISA~\cite{lai2024lisa} and GLaMM~\cite{rasheed2024glamm}, enable mask-level grounding within conversational settings. Despite advancing fine-grained spatial reasoning, these approaches assume single-image scenarios; in contrast, \modelname{} extends mask-level grounding to multi-image scenarios, requiring reasoning across disjoint scenes while producing coherent, grounded explanations.

\noindent \textbf{LVLM-Based Multi-Image Localized Reasoning.}
Extending pixel-level grounding to multiple images has largely been explored in the context of video segmentation \cite{lin2025glus, munasinghe2025videoglamm, zheng2025villa, gong2025devil, yuan2025sa2va, wei2025hyperseg}. These methods
combine language models with segmentation decoders and temporal modules to reason across frames. However, temporal alignment simplifies object association, making these settings fundamentally different from reasoning over disjoint image collections without shared spatial or temporal cues. In the disjoint setting, works such as MC-Bench~\cite{xu2025mc}, Migician~\cite{li2025migician}, and MIRG-RL~\cite{zheng2025mirg} explore cross-image grounding but focus primarily on bounding-box localization, while CALICO~\cite{nguyen2025calico} studies semantic co-segmentation without open-ended mask-level reasoning. 
Our work addresses these gaps by introducing \modelname{}, which performs grounded reasoning across disjoint images, and \textbf{\benchmarkname{}}, a benchmark explicitly designed to evaluate fine-grained, cross-image pixel grounding.

\noindent \textbf{Part-Level Segmentation.}
Part segmentation decomposes objects into semantic components, with benchmarks including Pascal-Part~\cite{chen2014detect}, PartImageNet~\cite{he2022partimagenet}, ADE20K-Part234~\cite{wei2024ov}, and PACO~\cite{ramanathan2023paco}. Recent work studies open-vocabulary, hierarchical, and language-guided object--part segmentation~\cite{sun2023going,li2024partglee,wang2024unveiling,choi2025fine}. Pixel-grounded LVLMs further incorporate object and part information into language-conditioned mask generation~\cite{rasheed2024glamm,lai2024lisa,wu2024see}. \modelname{} extends this line to open-ended reasoning and object- and part-level grounding across multiple disjoint images.

\begin{table}[t]
\centering
\caption{\textbf{Comparison of benchmarks and models for Vision-Language Reasoning and Segmentation}, each assessed by the availability of localization annotations, \emph{multi-image} inputs, and \emph{general reasoning}. \benchmarkname{} and \modelname{} provide multi-image, pixel-grounded reasoning.}
\vspace{-0.2cm}

\resizebox{\linewidth}{!}{
    \begin{tabular}{>{\centering\arraybackslash}m{0.3cm}|p{8.6cm}>{\centering\arraybackslash}p{1.5cm}>{\centering\arraybackslash}p{1.5cm}>{\centering\arraybackslash}p{1.5cm}>{\centering\arraybackslash}p{1.9cm}}
        \toprule
        \textbf{} & \textbf{Dataset (Model, Venue)} & \textbf{Bounding Boxes} & \textbf{Seg. Masks} & \textbf{Multi-Image} & \textbf{General Reasoning} \\
        \midrule 
        \midrule
         \multirow{9}{*}{\rotatebox{90}{\textbf{Image-level}}}
         & \gc \textbf{CompBench} (---, NeurIPS-24)~\cite{kil2024compbench}  & \gc \xmark & \gc \xmark & \gc \cmark & \gc \xmark \\
         & \textbf{M4-Instruct} (LLaVA-NeXT-Interleave, arXiv-24)~\cite{li2024llava} & \xmark & \xmark & \cmark & \cmark \\
         & \gc \textbf{MANTIS-INSTRUCT} (MANTIS, TMLR-24)~\cite{jiang2024mantis} & \gc \xmark & \gc \xmark & \gc \cmark & \gc \cmark \\
         & \textbf{MMRA} (---, arXiv-24)~\cite{wu2024mmra} & \xmark & \xmark & \cmark & \xmark \\
         & \gc \textbf{ReMI} (---, NeurIPS-24)~\cite{kazemi2024remi} & \gc \xmark & \gc \xmark & \gc \cmark & \gc \xmark \\
         & \textbf{MMDU} (---, NeurIPS-24)~\cite{liu2024mmdu} & \xmark & \xmark & \cmark & \cmark \\
         & \gc \textbf{MMIU} (---, ICLR-25)~\cite{meng2025mmiu} & \gc \xmark & \gc \xmark & \gc \cmark & \gc \xmark \\ 
         & \textbf{OmniDiff} (M$^3$Diff, ICCV-25)~\cite{liu2025omnidiff} & \xmark & \xmark & \cmark & \xmark \\
         \midrule
        \multirow{14}{*}{\rotatebox{90}{\textbf{Localization}}}  
         & \gc \textbf{AS-V2} (ASMv2, ECCV-24)~\cite{wang2024all} & \gc \cmark & \gc \xmark & \gc \xmark & \gc \cmark \\
        & \textbf{MGSC} (MGLMM, AAAI-25)~\cite{yuan2025instruction} & \cmark & \cmark & \xmark & \cmark \\
                 & \gc \textbf{\textsc{LVIS-Ground}} (Groma, ECCV-24)~\cite{ma2024groma} & \gc \cmark & \gc \xmark & \gc \xmark & \gc \cmark \\
         & \textbf{FP-RefCOCO} (SESAME, CVPR-24)~\cite{wu2024see} & \cmark & \cmark & \xmark & \xmark \\
         & \gc \textbf{LLM-Seg40K} (LLM-Seg, CVPRW-24)~\cite{wang2024llm} & \gc \cmark & \gc \cmark & \gc \xmark & \gc \xmark \\
         & \textbf{MRES-32M} (UniRES, CVPR-24)~\cite{wang2024unveiling} & \cmark & \cmark & \xmark & \xmark \\
         & \gc \textbf{RecapD} (RGPT, CVPR-24)~\cite{guo2024regiongpt} & \gc \cmark & \gc \cmark & \gc \xmark & \gc \cmark \\
         & \textbf{MUSE} (PixelLM, CVPR-24)~\cite{ren2024pixellm} & \cmark & \cmark & \xmark & \cmark \\
         & \gc \textbf{ReasonSeg} (LISA, CVPR-24)~\cite{lai2024lisa} & \gc \cmark & \gc \cmark & \gc \xmark & \gc \cmark \\
         & \textbf{GranD} (GLaMM, CVPR-24)~\cite{rasheed2024glamm} & \cmark & \cmark & \xmark & \cmark \\
         & \gc \textbf{MMR} (M$^2$SA, ICLR-25)~\cite{jang2025mmr} & \gc \cmark & \gc \cmark & \gc \xmark & \gc \cmark \\
         & \textbf{MIG-Bench} (Migician, ACL-25)~\cite{li2025migician} & \cmark & \xmark & \cmark & \cmark \\
         & \gc \textbf{MC-Bench} (---, ICCV-25)~\cite{xu2025mc} & \gc \cmark & \gc \xmark & \gc \cmark & \gc \cmark \\
         & \textbf{MixedParts} (CALICO, CVPR-25)~\cite{nguyen2025calico} & \cmark & \cmark & \cmark & \xmark \\
         \cmidrule{2-6}
         & \cc \textbf{\benchmarkname} ~(ours) & \cc \cmark & \cc \cmark & \cc \cmark & \cc \cmark \\
        \bottomrule
    \end{tabular}
}
\label{tab:dataset_summary}
\end{table}

\section{Method}
\begin{figure*}[t!]
    \centering
        \includegraphics[width=0.99\linewidth]{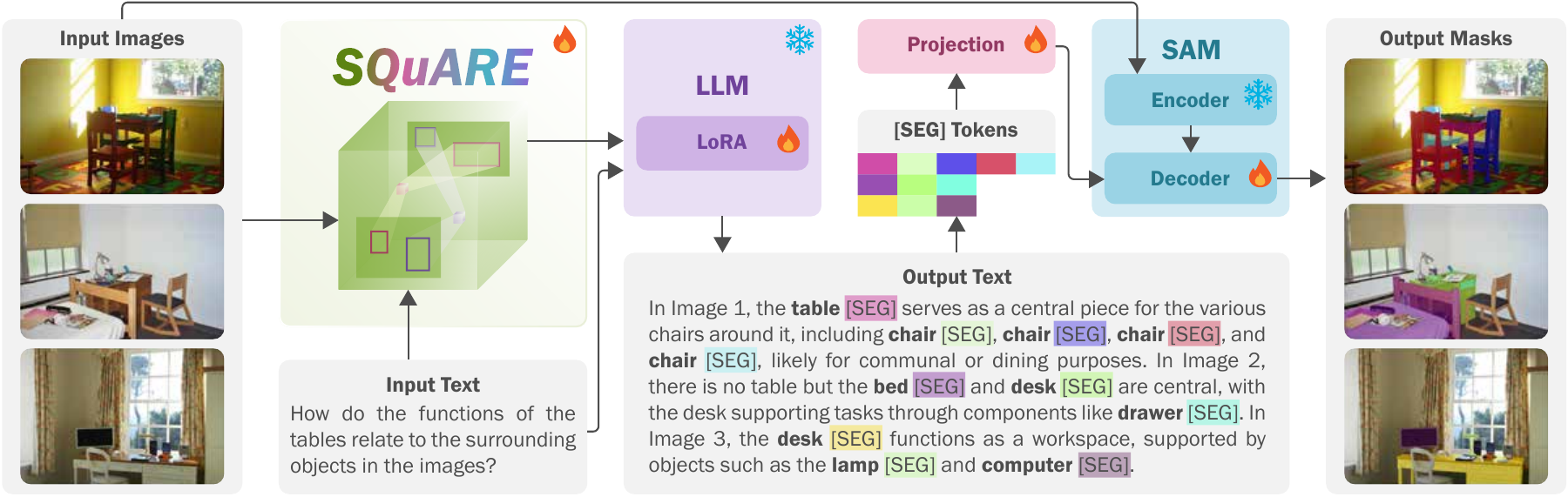} 
    
    \vspace{-0.2cm}
    \caption{\textbf{\modelname{} Overview}. Leveraging a LoRA-finetuned language model, a novel \modulename{} vision encoder (details in Fig. \ref{fig:square}), and a SAM-based decoder, \modelname{} dynamically generates segmentation masks corresponding to objects referenced in the generated responses, supporting pixel-grounded reasoning in complex multi-image tasks.
}
\label{fig:architecture}
\end{figure*}

\subsection{Problem Definition}
In multi-image pixel-grounded reasoning, inputs consist of a text prompt $\mathbf{T}_P$ and a set of $N_I$ images 
$\mathbf{I} = \{\mathbf{I}_1, \ldots, \mathbf{I}_{N_I}\}$, with each $\mathbf{I}_j \in \mathbb{R}^{3 \times H_j \times W_j}$. 
The goal is to generate a textual response that addresses the prompt while \emph{grounding} relevant noun phrases to segmentation masks across the images.
The response marks grounded noun phrases using special tags, each associated with a \emph{segmentation prompt} indicating the target image (\eg via an image identifier). 
From the response, we extract grounded phrases
$\mathbf{G} = \{\mathbf{G}_1, \ldots, \mathbf{G}_{N_I}\}$,
where $\mathbf{G}_j = \{\mathbf{g}_{j1}, \ldots, \mathbf{g}_{jG_j}\}$ denotes the phrases grounded in image $j$ and $G_j \ge 0$ is the number of grounded phrases for that image.
Each phrase $\mathbf{g}_{ji}$ is mapped to a binary segmentation mask
$\mathbf{m}_{ji} \in \{0,1\}^{H_j \times W_j}$,
forming the mask set $\mathbf{M}_j = \{\mathbf{m}_{j1}, \ldots, \mathbf{m}_{jG_j}\}$.
Here $\mathbf{m}_{ji}[u,v] = 1$ iff pixel $(u,v)$ belongs to the entity referenced by $\mathbf{g}_{ji}$ in image $j$.
We allow $G_j = 0$ for some images, provided that at least one grounded phrase exists overall, \ie $\sum_{j=1}^{N_I} G_j \ge 1$.

To tackle this task, we introduce \modelname{}, comprising an LLM backbone $\mathcal{B}$ that produces the text response via $\mathbf{T} \!=\! \mathcal{B}(\mathbf{T}_P, \mathbf{I})$,
and a segmentation module $\mathcal{S}$ that converts segmentation prompts into masks conditioned on the images, with $\mathbf{M} \!=\! \mathcal{S}(\mathbf{I}, \mathbf{S})$.
Here, $\mathbf{S} \subset \mathbf{T}$ denotes the segmentation prompts extracted from the response and grouped per image according to the embedded image identifiers.
For image $j$, let $\mathbf{S}_j \!=\! \{ \mathbf{s}_{j1}, \ldots, \mathbf{s}_{jG_j} \}$ be the prompts aligned to $\mathbf{G}_j$ in order of appearance, yielding $\mathbf{M}_j \!=\! \{ \mathbf{m}_{j1}, \ldots, \mathbf{m}_{jG_j} \}$.
This alignment induces a one-to-one correspondence between grounded phrases and predicted masks within each image. We thus want to fit $\mathcal{B}$ to produce accurate, well-structured responses (reasoning + grounding markup) and $\mathcal{S}$ to decode the segmentation prompts into high-quality masks across varying image resolutions and counts. 

\subsection{\textbf{\modelname{}} Architecture}

Our \modelname{} architecture, as illustrated in \Cref{fig:architecture}, consists of an LVLM backbone with a novel vision module
\modulename{} for intra- and inter-image understanding,
alongside a segmentation module for pixel grounding.

\noindent \textbf{Large Vision-Language Model $\mathcal{B}$.}
We integrate an LVLM backbone for multimodal reasoning, initialized from a strong single-image reasoning segmentation model \cite{rasheed2024glamm}. The images are encoded by $\mathcal{R}:\R^{3 \times H \times W} \mapsto \R^{L_I \times D}$ to get $\mathbf{I}_\mathcal{R}\!=\!\mathcal{R}(\mathbf{I}) \in \R^{N_I \times L_I \times D}$, aligning with the hidden dimension of the LLM. These embeddings are then interleaved with the embedded text prompt $\mathbf{T}_\text{emb}\! \in \R^{L_T \times D}$, forming multimodal embeddings $\mathbf{T}_\text{mm} \in \R^{(L_T + N_I \cdot L_I) \times D}$ to be fed into the LLM. 
In practice, we prepend a given text instruction with a sentence that triggers the model's visual processing, \eg ``The \texttt{<image>} (IMAGE1), \texttt{<image>} (IMAGE2), and \texttt{<image>} (IMAGE3) provide an overview of the pictures.'' Here, the $i^\text{th}$ \texttt{<image>} token is replaced with the corresponding projected image embeddings $\mathbf{I}^{(i)}_\mathcal{R} \in \R^{L_I \times D}$, and the remaining prompt is embedded via a standard text embedding layer $f_{\mathcal{B} \text{ emb}}$. The LVLM is trained to generate text that combines reasoning about the question with information necessary for segmentation.\looseness-1

\begin{figure}[t]
    \centering
    \includegraphics[width=0.9\linewidth]{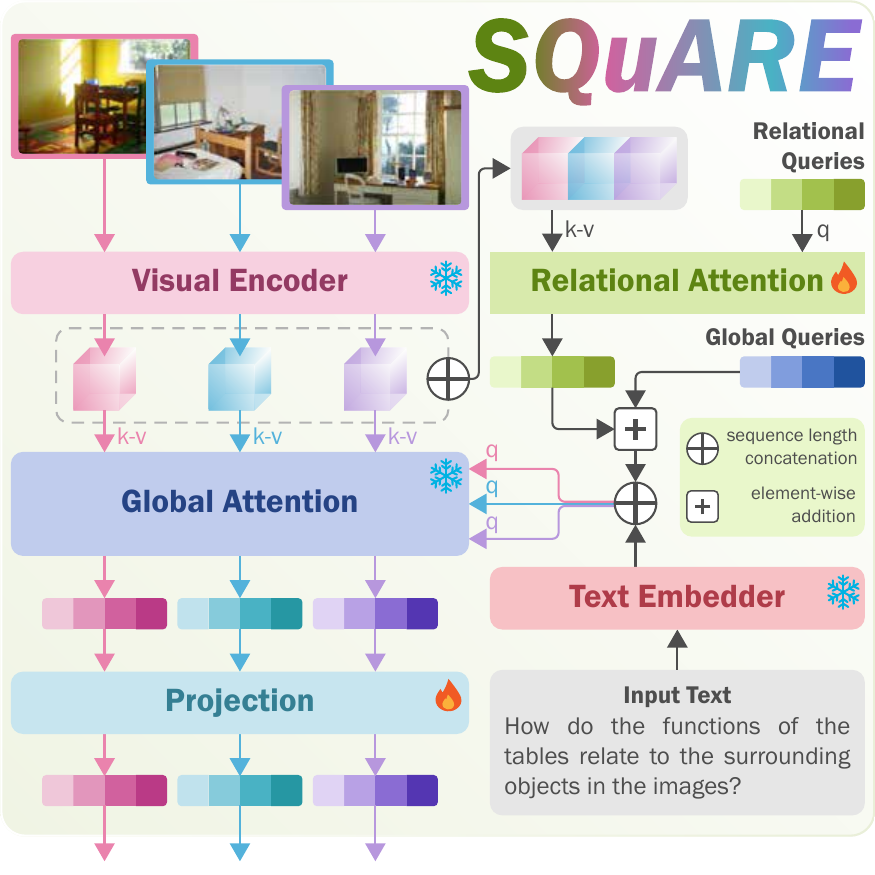}
    \vspace{-0.2cm}
    \caption{\textbf{Our proposed \modulename{} module}. Learnable relational queries attend over the concatenated multi-image features to form a shared relational representation. This representation is injected into the query pathway for global feature extraction, producing enriched visual representations that capture cross-image interactions.}
    \label{fig:square}
    \vspace{-0.2cm}
\end{figure}

\noindent \textbf{\underline{S}hared \underline{Qu}ery-\underline{A}ttention \underline{R}elational \underline{E}ncoder (\modulename{}).}
To enable efficient multi-image understanding, we introduce \modulename{}, a novel vision module that hierarchically extends a query-based vision encoder~\cite{li2023blip} to extract rich text-guided, cross-image features. Standard LLaVA-style LVLMs~\cite{li2024llava, lai2024lisa, rasheed2024glamm} typically employ a CLIP image encoder~\cite{radford2021learning} followed by a projection layer to align visual features with the language space. However, this approach preserves the full sequence of image embeddings, which is often long (\eg 256~\cite{lai2024lisa} or 576~\cite{rasheed2024glamm}) and quickly becomes computationally expensive as the number of images increases.
In contrast, query-driven encoders~\cite{li2023blip, dai2023instructblip} have shown strong performance, particularly for multi-image understanding~\cite{li2024fine}, while significantly reducing sequence length. Prior work also suggests that, although cross-modal interactions already occur within the LLM, enriching the encoder’s query representations prior to LLM integration further improves performance~\cite{dai2023instructblip}. Building on these insights, our novel \modulename{} module (\Cref{fig:square}) injects both text and cross-image signals into the query pathway before passing features to the LLM.\looseness-1

\modelname{} employs a query-based vision encoder composed of an image encoder $\mathcal{V}$, learnable global queries $\mathbf{q}_0$, and a global attention module $\mathcal{Q}$. To explicitly model cross-image relationships, we introduce a complementary set of \textit{relational queries} $\mathbf{c}_0$ together with a \textit{relational attention} module $\mathcal{C}$. Both query sets share the same dimensionality, $\mathbf{q}_0, \mathbf{c}_0 \in \mathbb{R}^{L_I \times D_I}$, allowing relational interactions to be integrated directly within the visual query space. The resulting visual representations are projected into the LLM space via $f_{i\to t}: \mathbb{R}^{D_I} \mapsto \mathbb{R}^{D}$.
Formally, given input images $\mathbf{I}$, we first pass them through the vision encoder to obtain image features $\mathbf{I}_\mathcal{V}\!=\!\mathcal{V}(\mathbf{I}) \in \R^{N_I \times L_V \times D_V}$. We then concatenate these features across the sequence dimension, forming a multi-context representation $\mathbf{I}_{\mathcal{V}\;\text{fused}}\!=\!\bigoplus_{j=1}^{N_I} \left\{\mathbf{I}_{\mathcal{V}j}\right\} \in \R^{\left(N_I \times L_V\right) \times D_V}$, which is then provided as key-value to the relational attention module $\mathcal{C}$ to enable cross-context fusion:
\begin{equation*}
\setlength{\abovedisplayskip}{3pt}
\setlength{\belowdisplayskip}{3pt}
\setlength{\abovedisplayshortskip}{2pt}
\setlength{\belowdisplayshortskip}{2pt}
    \mathbf{c}\!=\!\mathcal{C}\!\left(\mathbf{c}_0, \mathbf{I}_{\mathcal{V}\;\text{fused}}, \mathbf{I}_{\mathcal{V}\;\text{fused}}\right) \in \R^{L_I \times D_I}.
\end{equation*}
The resulting relational features are added to the base query tokens to obtain enriched queries $\mathbf{q}'_0=\mathbf{c} + \mathbf{q}_0$. Concatenating these with the embedded instruction prompt $\mathbf{T}_Q$ yields the final query sequence, which we use to independently extract single-image global representations from $\mathbf{I}_\mathcal{V}$:
\begin{equation*}
\setlength{\abovedisplayskip}{3pt}
\setlength{\belowdisplayskip}{3pt}
\setlength{\abovedisplayshortskip}{2pt}
\setlength{\belowdisplayshortskip}{2pt}
    \mathbf{I}_{\mathcal{Q}}\!=\!\left[\mathcal{Q}\!\left(\mathbf{q}'_0 \oplus \mathbf{T}_Q, \mathbf{I}_\mathcal{V}, \mathbf{I}_\mathcal{V} \right)\right]_{1:L_I} \in \R^{N_I \times L_I \times D_I},
\end{equation*}
where we retain the first $L_I$ output tokens corresponding to the enriched queries for each image. Lastly, to obtain an aligned representation for the LLM, we project $\mathbf{I}_\mathcal{Q}$ and get the final output $\mathbf{I}_\mathcal{R} = f_{i\to t}\left(\mathbf{I}_\mathcal{Q}\right) \in \R^{N_I \times L_I \times D}$.

\noindent \textbf{Pixel Grounding.} We equip \modelname{} with pixel grounding capabilities by incorporating SAM~\cite{kirillov2023segment}, a robust pretrained open-world segmentation model capable of generating masks from diverse types of input prompts. Following past works \cite{rasheed2024glamm, lai2024lisa}, we utilize the \texttt{[SEG]} token's final hidden state to prompt the segmentation module, alongside the encoded images. Consequently, each grounded noun phrase $\mathbf{g}_{ji}$ in the LVLM’s output is paired with its unique \texttt{[SEG]} token. To distinguish tokens associated with different images, we append an image identifier to each segmentation token, \eg \texttt{[SEG] (IMAGE2)}. Finally, to clearly delineate noun phrases within the text output, we train a pair of grounding tokens, \texttt{<p>} and \texttt{</p>}, to enclose each noun phrase. After text generation, we pass the \texttt{[SEG]} tokens, alongside the corresponding SAM-encoded image embeddings, into the SAM decoder for segmentation.

Formally, the generated output gives \texttt{[SEG]} token embeddings $\mathbf{S}\!=\!\{\mathbf{S}_1, \cdots, \mathbf{S}_{N_I}\}$, each $\mathbf{S}_j\!=\!\left\{\mathbf{s}_{j1}, \cdots, \mathbf{s}_{jG_j}\right\} \in \R^{G_j \times D}$ containing the $G_j$ segmentation tokens corresponding to image $j$. We project these tokens via $f_\text{t--s}: \R^D \mapsto \R^{D_S}$ to $\mathcal{S}$'s prompt encoding space: $\mathbf{S}_\text{pro}\!=\!f_\text{t--s} \left(\mathbf{S}\right)$. We encode the images with $\mathcal{S}$'s frozen encoder $\mathcal{S}_\text{enc}: \R^{3 \times H \times W} \mapsto \R^{L_S \times D_S}$ to get $\mathbf{I}_\text{seg}\!=\!\mathcal{S}_\text{enc}(\mathbf{I})$, and finally obtain segmentation masks $\mathbf{M}$ via the decoder $\mathcal{S}_\text{dec}$ by $\mathbf{M}\!=\!\mathcal{S}_\text{dec}(\mathbf{I}_\text{seg}, \mathbf{S}_\text{pro})$.

\begin{figure}[t]
    \centering
    \small
    \setlength{\tabcolsep}{5pt}
    \captionof{table}{\textbf{\benchmarkname{} sample distribution} with respect to annotation source and question type alongside unique object and part counts.}
    \vspace{-0.2cm}
    \label{fig:data_stats}
    \begin{tabularx}{\linewidth}{@{}lX@{}}
        \toprule
        \textbf{Statistic} & \textbf{Distribution (\%)} \\
        \midrule

        Source &
        \begin{tikzpicture}[x=\linewidth,y=1cm]
            \node[font=\scriptsize] at (0.18649953,0.42) {\textbf{ADE20K}};
            \node[font=\scriptsize] at (0.58010808,0.42) {\textbf{COCO}};
            \node[font=\scriptsize] at (0.89360855,0.42) {\textbf{PASCAL}};

            \fill[stat1a] (0.00000000,0) rectangle (0.37299906,0.28);
            \fill[stat1b] (0.37299906,0) rectangle (0.78721709,0.28);
            \fill[stat1c] (0.78721709,0) rectangle (1.00000000,0.28);

            \node[font=\tiny] at (0.18649953,0.14) {37.3};
            \node[font=\tiny] at (0.58010808,0.14) {41.4};
            \node[font=\tiny,text=white] at (0.89360855,0.14) {21.3};
        \end{tikzpicture}
        \\[-1pt]

        Question type &
        \begin{tikzpicture}[x=\linewidth,y=1cm]
            \node[font=\scriptsize] at (0.07275647,0.42) {\textbf{Num.}};
            \node[font=\scriptsize] at (0.23336150,0.42) {\textbf{Spatial}};
            \node[font=\scriptsize] at (0.45935057,0.42) {\textbf{Functional}};
            \node[font=\scriptsize] at (0.79874554,0.42) {\textbf{Open-Ended}};

            \fill[stat2a] (0.00000000,0) rectangle (0.14551294,0.28);
            \fill[stat2b] (0.14551294,0) rectangle (0.32121005,0.28);
            \fill[stat2c] (0.32121005,0) rectangle (0.59749108,0.28);
            \fill[stat2d] (0.59749108,0) rectangle (1.00000000,0.28);

            \node[font=\tiny] at (0.07275647,0.14) {14.6};
            \node[font=\tiny] at (0.23336150,0.14) {17.6};
            \node[font=\tiny,text=white] at (0.45935057,0.14) {27.6};
            \node[font=\tiny,text=white] at (0.79874554,0.14) {40.3};
        \end{tikzpicture}
        \\[-1pt]

        \# unique objects &
        \begin{tikzpicture}[x=\linewidth,y=1cm]
            \node[font=\scriptsize] at (0.17387096,0.42) {\textbf{1--3}};
            \node[font=\scriptsize] at (0.44912239,0.42) {\textbf{4--5}};
            \node[font=\scriptsize] at (0.61523877,0.42) {\textbf{6--7}};
            \node[font=\scriptsize] at (0.72615026,0.42) {\textbf{8--9}~~~};
            \node[font=\scriptsize] at (0.81555564,0.42) {\textbf{10--12}};
            \node[font=\scriptsize] at (0.92939273,0.42) {\textbf{13+}};

            \fill[stat3a] (0.00000000,0) rectangle (0.34774193,0.28);
            \fill[stat3b] (0.34774193,0) rectangle (0.55050285,0.28);
            \fill[stat3c] (0.55050285,0) rectangle (0.67997469,0.28);
            \fill[stat3d] (0.67997469,0) rectangle (0.77232582,0.28);
            \fill[stat3e] (0.77232582,0) rectangle (0.85878546,0.28);
            \fill[stat3f] (0.85878546,0) rectangle (1.00000000,0.28);

            \node[font=\tiny,text=white] at (0.17387096,0.14) {34.8};
            \node[font=\tiny,text=white] at (0.44912239,0.14) {20.3};
            \node[font=\tiny,text=white] at (0.61523877,0.14) {12.9};
            \node[font=\tiny,text=white] at (0.72615026,0.14) {9.2};
            \node[font=\tiny,text=white] at (0.81555564,0.14) {8.6};
            \node[font=\tiny,text=white] at (0.92939273,0.14) {14.1};
        \end{tikzpicture}
        \\[-1pt]

        \# unique parts &
        \begin{tikzpicture}[x=\linewidth,y=1cm]
            \node[font=\scriptsize] at (0.05734314,0.42) {\textbf{1--5}};
            \node[font=\scriptsize] at (0.27421221,0.42) {\textbf{6--10}};
            \node[font=\scriptsize] at (0.59557434,0.42) {\textbf{11--15}};
            \node[font=\scriptsize] at (0.85253512,0.42) {\textbf{16--20}};
            \node[font=\scriptsize] at (0.97382986,0.42) {\textbf{21+}};

            \fill[stat4a] (0.00000000,0) rectangle (0.11468628,0.28);
            \fill[stat4b] (0.11468628,0) rectangle (0.43373815,0.28);
            \fill[stat4c] (0.43373815,0) rectangle (0.75741053,0.28);
            \fill[stat4d] (0.75741053,0) rectangle (0.94765972,0.28);
            \fill[stat4e] (0.94765972,0) rectangle (1.00000000,0.28);

            \node[font=\tiny] at (0.05734314,0.14) {11.5};
            \node[font=\tiny,text=white] at (0.27421221,0.14) {31.9};
            \node[font=\tiny,text=white] at (0.59557434,0.14) {32.4};
            \node[font=\tiny,text=white] at (0.85253512,0.14) {19.0};
            \node[font=\tiny,text=white] at (0.97382986,0.14) {5.2};
        \end{tikzpicture}
        \\

        \bottomrule
    \end{tabularx}
\end{figure}

\noindent \textbf{Training Loss.}
\modelname{} is trained with next-token prediction loss $\mathcal{L}_\text{text}$. The segmentation decoder is optimized with Dice ($\mathcal{L}_\text{Dice}$) \cite{milletari2016v} and focal loss ($\mathcal{L}_\text{focal}$) \cite{lin2017focal}, following prior works \cite{kirillov2023segment, lai2024lisa, rasheed2024glamm}. The overall training objective is $\mathcal{L}\!=\!\alpha \mathcal{L}_\text{text} + \beta \mathcal{L}_\text{Dice} + \gamma \mathcal{L}_\text{focal}$ with hyperparameters $\alpha$, $\beta$, and $\gamma$.\looseness-1
\section{\benchmarkname{} Dataset}
To address the lack of multi-image benchmarks for pixel-grounded reasoning, we introduce \benchmarkname{}, a dataset for evaluating LVLMs on this task. Unlike prior benchmarks, \benchmarkname{} contains open-set question-answer pairs that require object- and part-level reasoning across multiple images, with corresponding segmentation masks. As shown in \Cref{fig1}, \benchmarkname{} covers diverse comparisons at different reasoning levels, such as comparing object sizes across images or identifying the distinct functions of objects in different contexts. The open-world nature makes \benchmarkname{} a challenging pixel grounding benchmark, requiring complex cross-image visual co-reference, relational reasoning, and grounding at the fine-grained level.

\noindent \textbf{Dataset Construction.} We construct \benchmarkname{} from six public datasets with human-annotated masks: three part-level datasets—ADE20K-Part234~\cite{wei2024ov}, PASCAL-Part~\cite{chen2014detect}, and PACO-LVIS~\cite{ramanathan2023paco}—and their respective object-level source datasets, ADE20K~\cite{zhou2017scene}, PASCAL VOC 2010~\cite{everingham2010pascal}, and COCO~\cite{lin2014microsoft} (\Cref{fig:data_stats}). We preserve the original dataset splits when constructing \benchmarkname. Training and test examples are drawn from their respective source-image pools, ensuring that no source image is shared between training and evaluation. For datasets without a dedicated test split, we use the original validation split as the source of test images. For each source image, we sample related images depicting similar scenes or activities and containing comparable, but non-identical, objects. Using the sampled images, masks, and object bounding boxes, we generate multi-image question-answer pairs following prior work~\cite{jang2025mmr, rasheed2024glamm} and ensuring every example requires reasoning across images and grounds relevant objects or parts. 
\benchmarkname{} questions span (a) functional reasoning about object use and interactions, (b) spatial reasoning across images, (c) numerical reasoning involving counts or quantities, and (d) open-ended reasoning that integrates information from multiple images. 
Apart from meticulously designed prompts, we apply a series of rigorous filtering steps to further ensure data quality. Specifically, we exclude responses without grounding and responses referencing absent or hallucinated objects. For the test set, we additionally exclude examples involving more than 16 segmentation masks. To ensure a reliable test set, we retain high-quality examples based on an ensemble score comprising question and answer naturalness, correctness, clarity, grounding ambiguity, completeness, and overall quality; details in Appendix \ref{supp:dataset}.\looseness-1
\section{Experiments}

\begin{table}[t!]
\centering
\caption{\textbf{\benchmarkname{} Results}. $^\dagger$Non-segmentation models; masks generated post-hoc with SAM~\cite{kirillov2023segment}. Best performance in \textbf{bold}.}
\setlength\tabcolsep{2pt}
\vspace{-0.2cm}
\resizebox{0.99\linewidth}{!}{
\begin{tabular}{lcccccccc}
\toprule
\textbf{Model} & \textbf{mIoU$\uparrow$} & \textbf{Recall$\uparrow$} & \textbf{I-SS$\uparrow$} & \textbf{I-SIoU$\uparrow$}  & \textbf{SS$\uparrow$} & \textbf{S-IoU$\uparrow$} & \textbf{METEOR$\uparrow$} \\ 
\midrule
\multicolumn{8}{c}{\cellcolor[gray]{0.9}\textbf{Open-Source LVLMs (post-hoc grounding)$^\dagger$}} \\
Qwen3-VL-8B & 26.73 & 17.57 & 25.13 & 19.17 & 75.22 & 59.45 & 22.14  \\
NVIDIA-Nemotron-Nano-12B-v2-VL &  15.78 & 10.00 & 14.04 & 11.03 & 66.31 & 51.33 & 21.39  \\
Qwen3-VL-235B-A22B & 30.69 & 21.05 & 27.96 & 19.25 & 76.89 & 56.32 & 23.72   \\
InternVL3-78B & 24.32 & 15.87  &  26.81 & 22.43 & 68.03 & 52.18 & 20.59 \\
Pixtral-Large-124B & 34.22 & 20.80 & 33.42 & 26.28 & 77.68 & 60.39 & 23.61  \\
\midrule
\multicolumn{8}{c}{\cellcolor[gray]{0.9}\textbf{Closed-Source LVLMs (post-hoc grounding)$^\dagger$}} \\
GPT-4.1 & 35.12 & 23.65 & 34.81 & 25.22 & 79.62 & 58.15 & 25.67   \\
Gemini 2.5 Pro & 32.10 & 19.64 & 27.04 & 16.94 & 76.42 & 52.28 & 20.68  \\
\midrule
\multicolumn{8}{c}{\cellcolor[gray]{0.9}\textbf{Pixel-Grounding LVLMs}} \\
LISA & 41.88 & 27.68 & 44.85 & 36.8 & 76.07 & 64.67 & 28.63  \\
GLaMM & 44.15 & 29.53 & 47.64 & 40.05 & 77.09 & 66.53 & 28.16  \\
\midrule 
\cc \textbf{\modelname{}} (ours) & \cc \textbf{54.95} & \cc \textbf{41.33} & \cc \textbf{64.49} & \cc \textbf{61.48} & \cc \textbf{83.19} & \cc \textbf{78.76} & \cc \textbf{28.81} \\ 
\midrule
\textcolor{DarkSageGreen}{$\Delta$\textbf{Ours - GLaMM}} & 
\textcolor{DarkSageGreen}{\textbf{+}$\uparrow$} 10.8  & 
\textcolor{DarkSageGreen}{\textbf{+}$\uparrow$} 11.8 & 
\textcolor{DarkSageGreen}{\textbf{+}$\uparrow$} 16.85 & 
\textcolor{DarkSageGreen}{\textbf{+}$\uparrow$} 21.43 &
\textcolor{DarkSageGreen}{\textbf{+}$\uparrow$} 6.10 &
\textcolor{DarkSageGreen}{\textbf{+}$\uparrow$} 12.23 &
\textcolor{DarkSageGreen}{\textbf{+}$\uparrow$} 0.65 \\
\bottomrule
\end{tabular}
}
\label{tab:main_exp}
\end{table}

\textbf{Experimental Setup.}
To the best of our knowledge, \modelname{} is the first model capable of addressing fine-grained pixel reasoning tasks involving multiple targets across multiple images. We initialize \modelname{} from GLaMM \cite{rasheed2024glamm}, a strong LVLM for single-image pixel-grounded reasoning segmentation, providing a robust segmentation-capable foundation on which we extend multi-image reasoning. To establish baselines, we first adapt various general-purpose VLMs, \eg GPT-4o and 4.1~\cite{hurst2024gpt, openai2025gpt41}, Gemini-2.5 Pro~\cite{comanici2025gemini}, InternVL3-78B~\cite{zhu2025internvl3}, Qwen3-VL~\cite{bai2025qwen3} (8B and 235B), Nemotron-Nano-12B-v2-VL~\cite{deshmukh2025nvidia}, and Pixtral Large~\cite{mistral2024pixtralLarge}. However, since these VLMs do not natively support pixel grounding, we convert the text-based grounding information following previous work~\cite{ren2024grounded}. Additionally, we adapt two state-of-the-art pixel-grounding LVLMs, LISA~\cite{lai2024lisa} and GLaMM~\cite{rasheed2024glamm}, both lacking explicit multi-image reasoning as they process only one image. Following previous work~\cite{nguyen2025calico,xu2025mc}, we modify these models to be able to reason over multiple images.
We finetune GLaMM and LISA on \benchmarkname{} using their reported hyperparameters. To ensure comparable exposure to training data, all trained models observe the same number of effective samples. 
Following prior work~\cite{rasheed2024glamm, yuan2024osprey}, we employ multiple metrics to evaluate the generated segmentation masks and natural language responses, including mIoU, Recall, IoU-Matched Semantic Similarity (I-SS), IoU-Matched Semantic IoU (I-SIoU), Semantic Similarity (SS), S-IoU, and METEOR. Additional details can be found in Appendix \ref{supp:metrics}.\looseness-1
\begin{table}[t] 
    \centering
    \caption{\textbf{Multi-Image Understanding Results.}}
    \vspace{-0.2cm}
    \label{tab:merged_benchmarks}

    {\small (a) BLINK} \\
    \resizebox{1.0\linewidth}{!}{
        \begin{tabular}{l|cccc}
\toprule
\textbf{Model}                 & {\textbf{Similarity}} & {\textbf{Jigsaw}} & {\textbf{Multi-view}} & {\textbf{Avg.}} \\
\midrule
Qwen-VL-Max             & 51.1                          & 4.7                       & 51.9                          & 35.9                               \\
LLaVA-v1.5-13b        & 47.4                          & 58.0                     & 44.4                          & 49.9                               \\
InstructBLIP-7B       & 47.4                          & 52.7                      & 55.6                          & 51.9                               \\
InstructBLIP-13B      & 49.6                          & 52.7                      & 57.1                          & 53.2                               \\
LLaVA-v1.6 34B        & 48.9                          & 54.7                      & 62.4                          & 55.3                               \\
 LLaVA-v1.6-7B & 45.9 & 47.3 & 44.4 & 45.9 \\
            Qwen2.5-VL-7B & 88.9 & 63.3 & 42.1 & 64.8 \\
            Qwen3-VL-8B & 84.4 & 56.7 & 55.6 & 65.6 \\
            Qwen3-VL-235B-A22B & 88.9 & 78.7 & 42.1 & 69.9 \\
GPT-4o                & 72.6                          & 55.3                      & 59.4                          & 62.4                               \\
\midrule
\cc \textbf{\modelname{}-7B}             & \cc  52.6                          & \cc  52.0                      & \cc  57.2                          & \cc  53.9                               \\
\bottomrule     
\end{tabular}
\label{tab:blink}

    }
    \vfill

    {\small (b) CompBench} \\
    \resizebox{1.0\linewidth}{!}{
        
\begin{tabular}{l|ccccc|cc|c|c}
\toprule
\multirow{2}{*}{\textbf{Model}} & \multicolumn{5}{c|}{\textbf{Attribute}} & \multicolumn{2}{c|}{\textbf{Emot.}} & \textbf{Spat.} & \textbf{Quan.} \\
 & \textbf{ST} & \textbf{FA} & \textbf{VA} & \textbf{CU} & \textbf{WF} & \textbf{CE} & \textbf{FE} & \textbf{ND} & \textbf{VQ} \\ \midrule
Random Choice & 50.0 & 50.0 & 50.0 & 50.0 & 50.0 & 50.0 & 50.0 & 50.0 & 33.3 \\ 
GPT-4V & {91.8} & {89.0} & 76.9 & 71.4 & {72.1} & 91.8 & 83.4 & 56.1 & {63.8} \\
Gemini1.0-Pro & 71.9 & 76.3 & 69.3 & 59.9 & 54.9 & 60.6 & 71.2 & 56.6 & 54.6 \\
LLaVA-1.6-34B & 84.9 & 72.1 & {77.7} & {72.6} & 68.7 & {96.2} & {83.5} & {67.2} & 50.1 \\
VILA-1.5-40B & 69.9 & 66.2 & 70.9 & 55.9 & 52.0 & 57.1 & 55.6 & 51.8 & 47.7 \\
LLaVA-v1.6-7B & 52.6 & 47.4 & 63.8 & 50.0 & 50.0 & 50.0 & 50.0 & 50.0 & 40.9 \\
            Qwen2.5-VL-7B & 88.0 & 89.6 & 78.7 & 71.5 & 73.7 & 70.4 & 74.8 & 64.9 & 61.8 \\
            Qwen3-VL-8B & 90.6 & 87.8 & 84.0 & 74.3 & 78.0 & 97.3 & 76.8 & 73.6 & 59.3 \\
            Qwen3-VL-235B-A22B & 93.6 & 96.0 & 88.6 & 82.6 & 88.8 & 98.2 & 86.7 & 74.7 & 67.6 \\
\midrule
\cc \textbf{\modelname{}-7B} & \cc 77.2 & \cc 69.2 & \cc 68.1 & \cc 62.4 & \cc 67.7 & \cc 78.3 & \cc 70.1 & \cc 62.2 & \cc 42.2 \\ \bottomrule
\end{tabular}
    }
    \vspace{-0.2cm}
\end{table}

\subsection{Experimental Results}
\textbf{Multi-Image Pixel-grounded Reasoning.} 
As shown in \Cref{tab:main_exp}, \modelname{} significantly outperforms both general-purpose and pixel-grounding LVLM baselines. The general-purpose LVLMs struggle with this task, a limitation we attribute to their lack of task-specific training and cascading errors from using SAM~\cite{kirillov2023segment, ren2024grounded} on the textual grounding information. While GPT-4.1 is the top performer in this category, its performance remains limited with $35.12\%$ mIoU and $79.62\%$ SS. The reasoning segmentation baselines perform better, as they leverage pixel grounding capabilities and are finetuned on \benchmarkname{}. GLaMM, for instance, outperforms the general-purpose LVLMs with $44.2\%$ mIoU. Compared to all baselines, \modelname{} sets a new benchmark, surpassing the strongest baseline by $10.8$ and $11.8$ points in mIoU and Recall, respectively. \modelname{} also achieves absolute gains of up to $6.1$ and $12.2$ points on text metrics SS and S-IoU.\looseness-1
\begin{table}[t]
\centering

\caption{\textbf{Cosegmentation Results} on MSRC and iCoseg.}
\vspace{-0.2cm}
\resizebox{0.85\linewidth}{!}{
\begin{tabular}{lcc|cc}
\toprule
\multirow{2}{*}{\textbf{Model}} 
  & \multicolumn{2}{c|}{\textbf{MSRC}} 
  & \multicolumn{2}{c}{\textbf{iCoseg}} \\
& $\mathcal{P}$ & $\mathcal{J}$ & $\mathcal{P}$ & $\mathcal{J}$ \\ 
\midrule
Rubinstein \etal{}~\cite{rubinstein2013unsupervised}
& 92.2 & 74.7 & 89.6 & 70.2 \\
Jerripothula \etal{}~\cite{jerripothula2016image}
& 88.7 & 71.0 & 91.9 & 74.0 \\
Wang \etal{}~\cite{wang2017multiple}
& 93.6 & 76.0 & 93.8 & 77.0 \\
Faktor and Irani~\cite{faktor2013co}
& 92.0 & 77.0 & 94.4 & 78.2 \\
Li \etal{}~\cite{li2019deep}
& 95.2 & 82.9 & 95.1 & 84.2 \\
Zhang \etal{}~\cite{zhang2020deep}
& 95.2 & 81.9 & -- & 89.2 \\
CycleSegNet ~\cite{zhang2021cyclesegnet}
& \textbf{97.9} & \textbf{87.2} & -- & 92.1 \\
\midrule
\cc \textbf{\modelname{}} (ours) 
& \cc \underline{96.0} & \cc \underline{83.4} 
& \cc \textbf{98.4} & \cc \textbf{92.7} \\
\bottomrule
\end{tabular}
}
\label{tab:suppl:cosegm}
\vspace{-0.3cm}
\end{table}

\noindent\textbf{Multi-Image Understanding.} 
We evaluate \modelname{}'s broader multi-image understanding capabilities on BLINK~\cite{fu2024blink} and CompBench~\cite{kil2024compbench}. Although \modelname{} is primarily optimized for multi-image reasoning segmentation, it demonstrates strong zero-shot generalization to perception tasks. As shown in \Cref{tab:merged_benchmarks}, \modelname{} is competitive with and often surpasses substantially larger LVLMs. Although it trails the newer Qwen models~\cite{bai2025qwen2vl}, \modelname{} remains competitive without benchmark-specific training. These results suggest that fine-grained spatial supervision supports high-level comparative reasoning despite \modelname{}'s smaller scale and pixel-grounding specialization.

\noindent\textbf{Other Benchmarks.}
We evaluate object cosegmentation on MSRC~\cite{shotton2006textonboost} and iCoseg~\cite{batra2010icoseg}. As shown in \Cref{tab:suppl:cosegm}, \modelname{} achieves state-of-the-art performance on iCoseg and performs competitively on MSRC, trailing only CycleSegNet, which is tailored specifically to this task and does not support reasoning or text generation. Appendix~\ref{sec:other_benchmarks} provides detailed qualitative and additional analyses on \modelname{}'s cosegmentation performance across a wide variety of categories, alongside MIG-Bench multi-image bounding-box grounding \cite{li2025migician} and MMR single-image reasoning \cite{jang2025mmr} evaluations.

\subsection{Computational Efficiency}
\begin{figure}[t]
    \centering
    \includegraphics[width=0.83\linewidth]{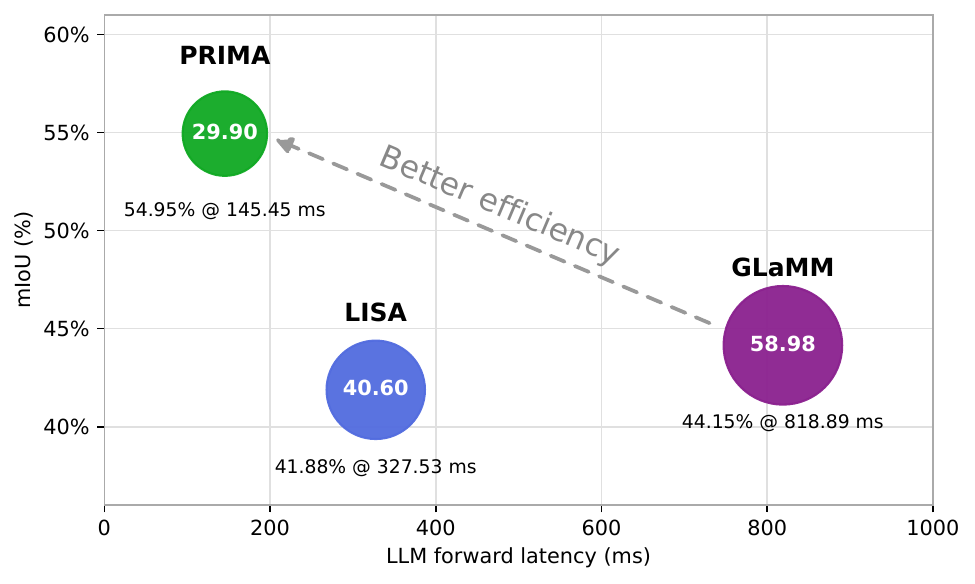}
        \vspace{-0.2cm}
    \caption{\textbf{mIoU vs LLM forward latency (ms).} Bubble area $\propto$ whole-model forward TFLOPs.}
    \label{fig:tokens}
    \vspace{-0.3cm}
\end{figure}

In \Cref{fig:tokens}, we profile \modelname{} against GLaMM and LISA using DeepSpeed's FLOPs profiler, with four data-parallel GPUs and a reported batch size of four per GPU. \modelname{} reduces the estimated whole-model forward computation per GPU to $29.90$ TFLOPs (vs. GLaMM's $58.98$ TFLOPs and LISA's $40.60$ TFLOPs), requiring $49.3\%$ and $26.4\%$ fewer floating-point operations, respectively. Its corresponding forward MAC count is $14.91$ TMACs, compared with GLaMM's $29.46$ TMACs and LISA's $20.27$ TMACs.
Our proposed \modulename{} contributes to these computational savings by compressing visual contexts into fewer, information-rich tokens, reducing the sequence length processed by the language backbone. The profiled language-backbone forward latency is reduced by \textbf{$5.63\times$} compared with GLaMM ($145.45$ ms vs. $818.89$ ms). These results highlight \modelname{}'s lower forward computational cost and faster language-backbone processing for multi-image pixel-grounded reasoning.\looseness-1

\begin{table*}[t!]
    \centering
    \resizebox{\linewidth}{!}{
        \begin{tabulary}{\linewidth}{
            >{\centering\arraybackslash}p{.55\linewidth}
            >{\centering\arraybackslash}p{.55\linewidth}
            >{\centering\arraybackslash}p{.55\linewidth}
            >{\centering\arraybackslash}p{.55\linewidth}
            >{\centering\arraybackslash}p{.55\linewidth}
            >{\centering\arraybackslash}p{.55\linewidth}
        }
        \toprule
        \multicolumn{2}{c}{\Large \textbf{Ground Truth}} &
            \multicolumn{2}{c}{\Large \textbf{\modelname{}}} &
            \multicolumn{2}{c}{\Large \textbf{GLaMM}} \\
            \midrule
            \multicolumn{6}{c}{
            \large\textbf{Q: Which object in the living room is best for holding drinks or snacks during a gathering, and what is the alternative in the laundry room for placing small items?}
  }\\
            \midrule
            \includegraphics[width=\linewidth]{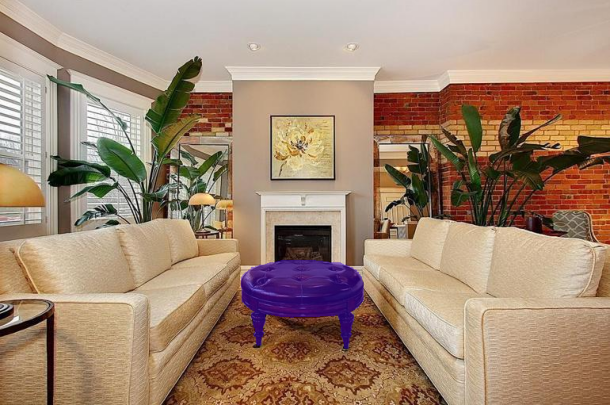} & 
            \includegraphics[height=.66\linewidth]{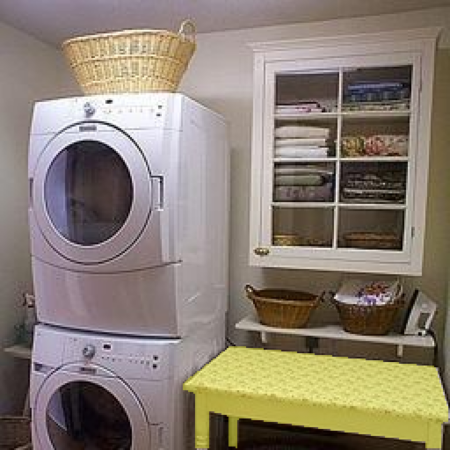} & 
            \includegraphics[width=\linewidth]{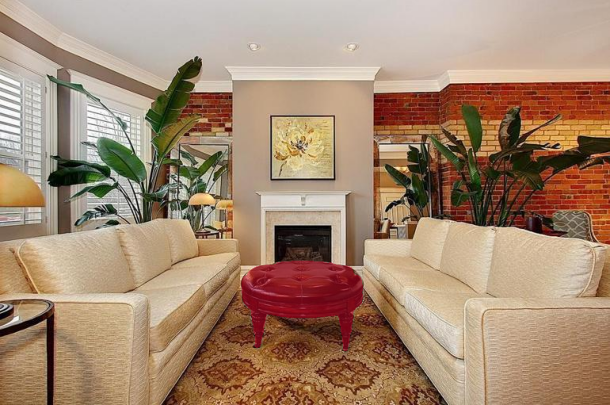} & 
            \includegraphics[height=.66\linewidth]{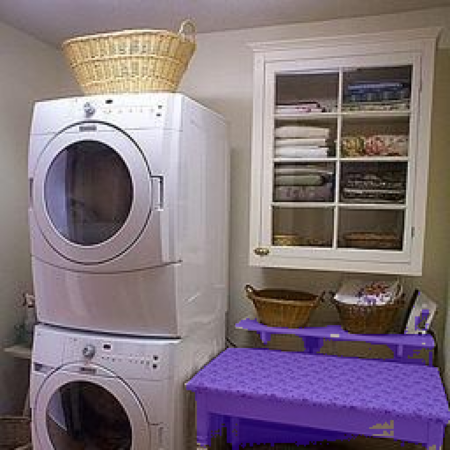} &
            \includegraphics[width=\linewidth]{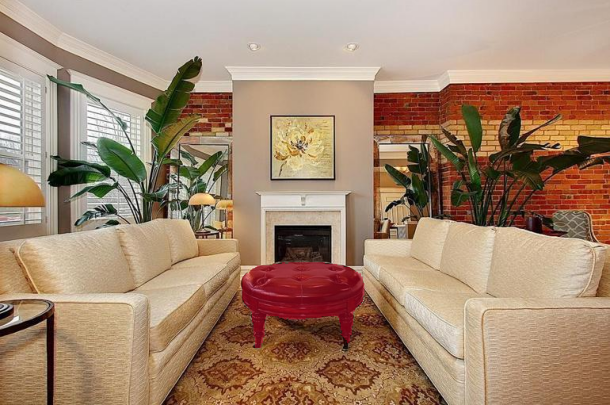} & 
            \includegraphics[height=.66\linewidth]{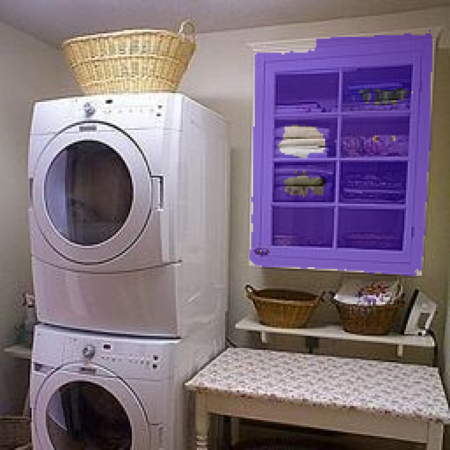} \\
            \multicolumn{2}{p{1.1\textwidth}}{\large
            In the living room setup, the \colorbox{gt51!50}{coffee table} is most suitable for holding
drinks or snacks. In a laundry room setting, the \colorbox{gt52!50}{table} offers an alternate
option for placing smaller items temporarily.
            } 
            &
            \multicolumn{2}{p{1.1\textwidth}}{ \large
            In a living room setup, the most suitable object for holding drinks or snacks is the \colorbox{prima51!50}{coffee table}. In a laundry room setting, the most suitable option for placing smaller items temporarily is the \colorbox{prima52!50}{table}.
            } 
            &
            \multicolumn{2}{p{1.1\textwidth}}{\large
            In a living room setup, the \colorbox{glamm51!50}{coffee table} is most suitable for holding drinks or snacks during social gatherings. In a laundry room setting, the \colorbox{glamm52!50}{cabinet} can serve as an alternative for placing smaller items temporarily.
            } \\
            \midrule
                    \multicolumn{6}{c}{\large \textbf{Q: In which image can we find the highest number of bicycles?}}\\
        \midrule
            \multicolumn{2}{c}{
              \begin{tabular}{@{}ccc@{}}
                \includegraphics[width=.22\linewidth]{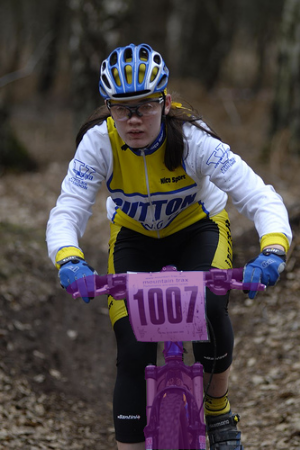}
                \includegraphics[width=.44\linewidth]{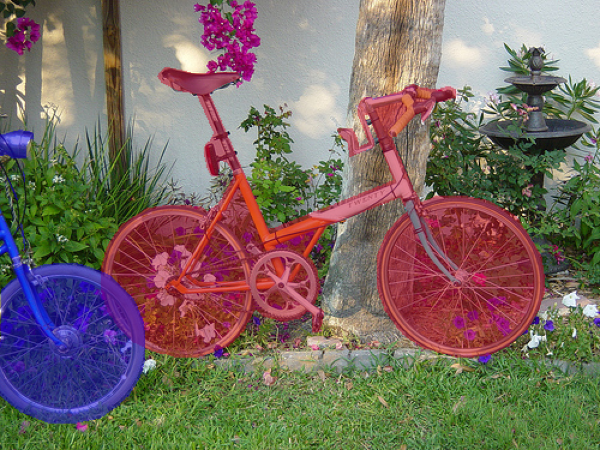}
                \includegraphics[width=.44\linewidth]{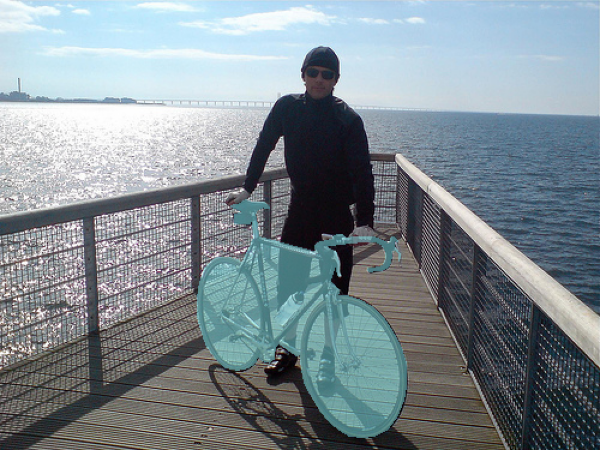}
              \end{tabular}
            } &
            \multicolumn{2}{c}{
              \begin{tabular}{@{}ccc@{}}
                \includegraphics[width=.22\linewidth]{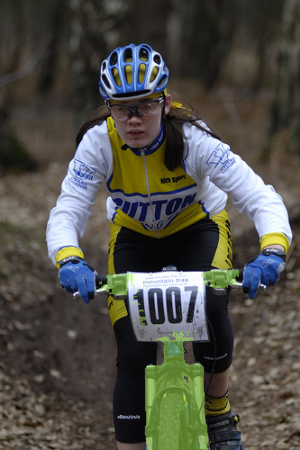}
                \includegraphics[width=.44\linewidth]{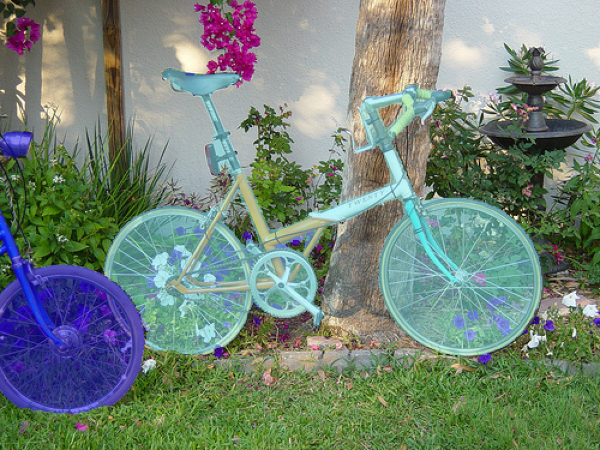}
                \includegraphics[width=.44\linewidth]{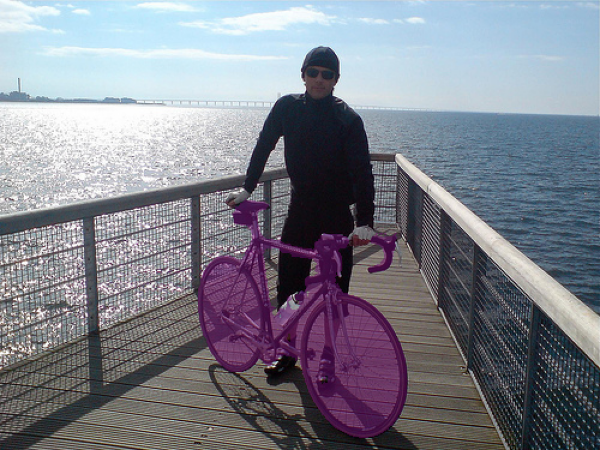}
              \end{tabular}
            } &
            \multicolumn{2}{c}{
              \begin{tabular}{@{}ccc@{}}
                \includegraphics[width=.22\linewidth]{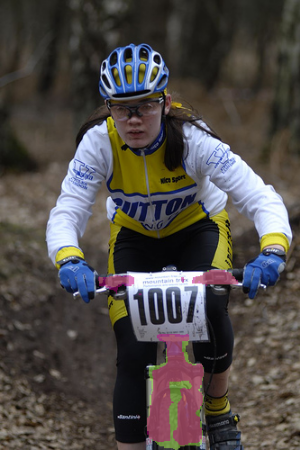}
                \includegraphics[width=.44\linewidth]{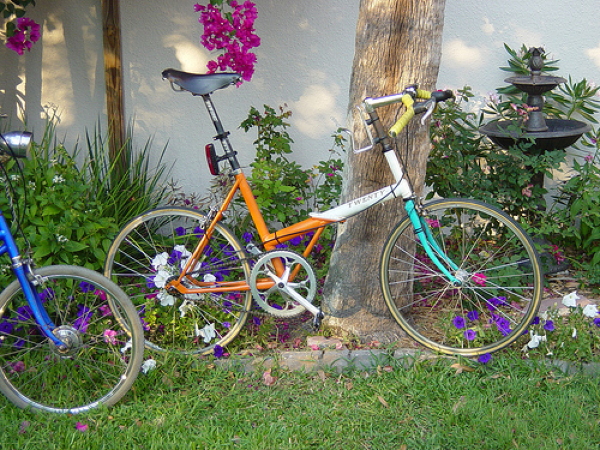}
                \includegraphics[width=.44\linewidth]{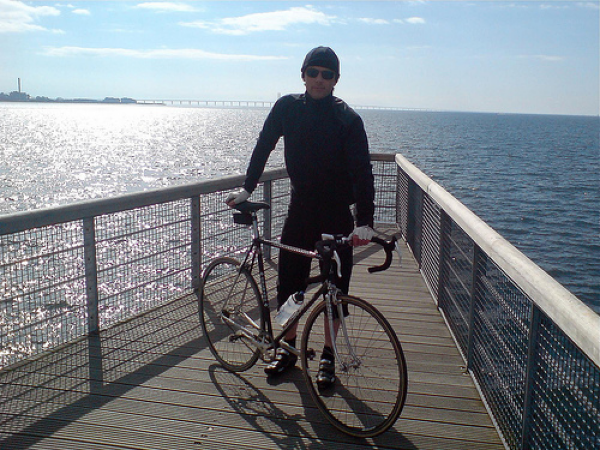}
              \end{tabular}
            } \\
            \multicolumn{2}{p{1.1\textwidth}}{\large
            In Image 1, there is one \colorbox{gt91!50}{bicycle}. On the other hand, Image 2 has 2 bicycles (\colorbox{gt92!50}{bicycle}, \colorbox{gt93!50}{bicycle}) in it. And, finally, Image 3 contains a single \colorbox{gt94!50}{bicycle}. So, the image with the highest number of bicycles is Image 2.
            } 
            &
            \multicolumn{2}{p{1.1\textwidth}}{\large
            Image 1 has one \colorbox{prima91!50}{bicycle} in it. And, Image 2 contains 2 bicycles (\colorbox{prima92!50}{bicycle}, \colorbox{prima93!50}{bicycle}). And, finally, Image 3 features one \colorbox{prima94!50}{bicycle}. That means the image with the highest number of bicycles is Image 2.
            } 
            &
            \multicolumn{2}{p{1.1\textwidth}}{\large
            Image 1 contains 2 bicycles (\colorbox{glamm91!50}{bicycle}, \colorbox{glamm92!50}{bicycle}). And, Image 2 contains a single \colorbox{glamm93!50}{bicycle}. That means the image with the highest number of bicycles is Image 2.
            } \\
        \midrule
            \multicolumn{6}{c}{\Large \textbf{\colorbox{DarkRed}{\textcolor{white}{Failure case}} Q: Which of these animals might rely more on human interaction for training and
care?}}\\
        \midrule
            \includegraphics[width=\linewidth]{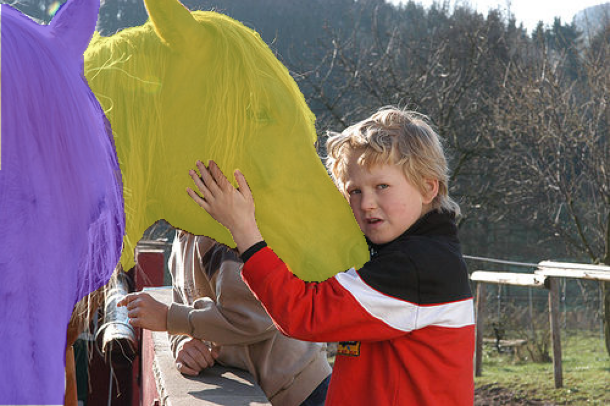} & 
            \includegraphics[height=0.66\linewidth]{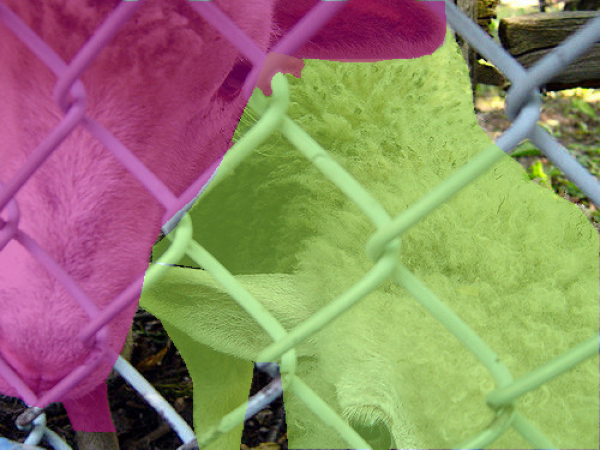} & 
            \includegraphics[width=\linewidth]{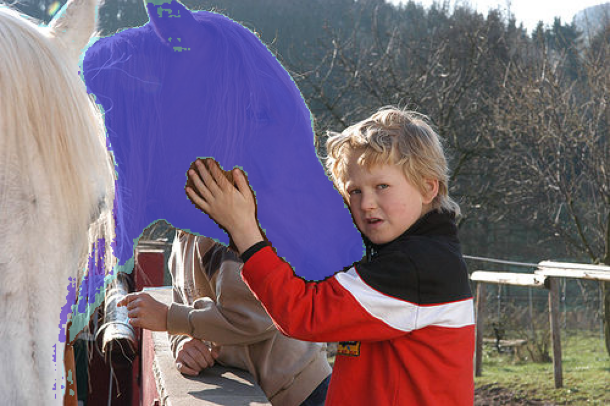} & 
            \includegraphics[height=0.66\linewidth]{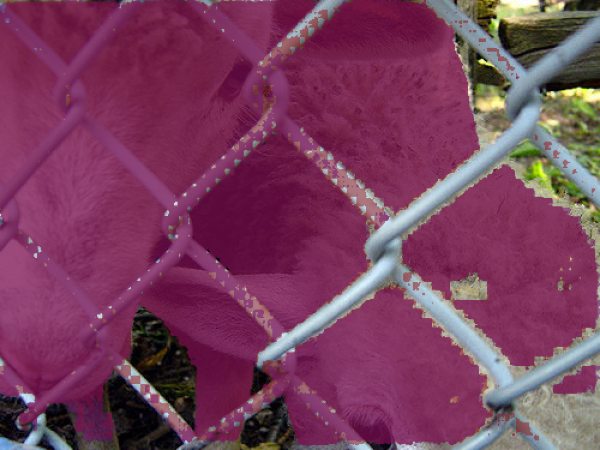} &
            \includegraphics[width=\linewidth]{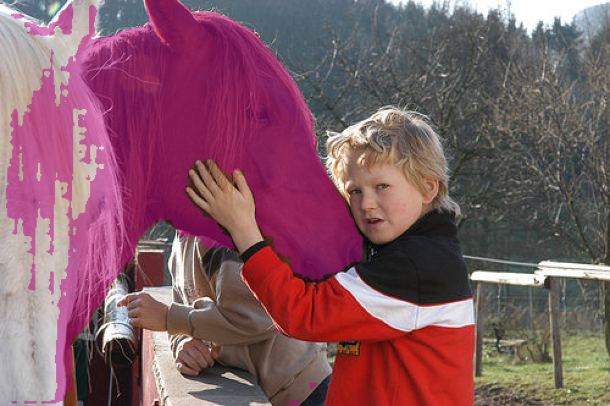} & 
            \includegraphics[height=0.66\linewidth]{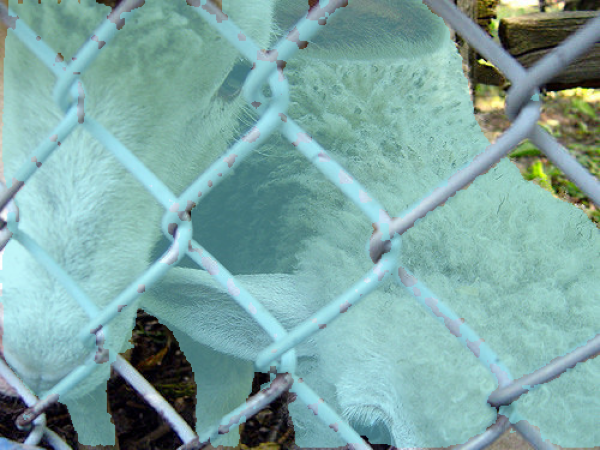} \\
            \multicolumn{2}{p{1.1\textwidth}}{\Large
            Horses, such as \colorbox{gt31!50}{horse} and \colorbox{gt32!50}{horse} from Image 1, often rely more on human
interaction for training and care compared to sheep like \colorbox{gt33!50}{sheep} and \colorbox{gt34!50}{sheep}
from Image 2, as they are commonly trained for riding and specific tasks.}
            &
            \multicolumn{2}{p{1.1\textwidth}}{\Large
            The horses, \colorbox{prima31!50}{horse} and \colorbox{prima32!50}{horse}, might rely more on human interaction for training and care, as they are typically kept and cared for by humans, unlike the sheep, \colorbox{prima33!50}{sheep} and \colorbox{prima34!50}{sheep}, which are more likely to be managed in a natural environment.
            } 
            &
            \multicolumn{2}{p{1.1\textwidth}}{\Large
            The \colorbox{glamm31!50}{dog} in Image 1 might rely more on human interaction for training and care compared to the \colorbox{glamm32!50}{sheep} in Image 2, as dogs are often trained by humans for various tasks and activities.
            } \\
        \bottomrule
        \end{tabulary}
    }
    \vspace{-0.1cm}
    \captionof{figure}{\textbf{Qualitative Results.} \modelname{} exhibits strong qualitative performance in both segmentation and reasoning.}
    \label{tab:config_comparison}
    \vspace{-0.3cm}
\end{table*}

\begin{figure*}[t!]
    \centering
    \resizebox{.996\textwidth}{!}{
    \begin{subfigure}[b]{0.32\textwidth}
        \centering
        \includegraphics[width=\textwidth]{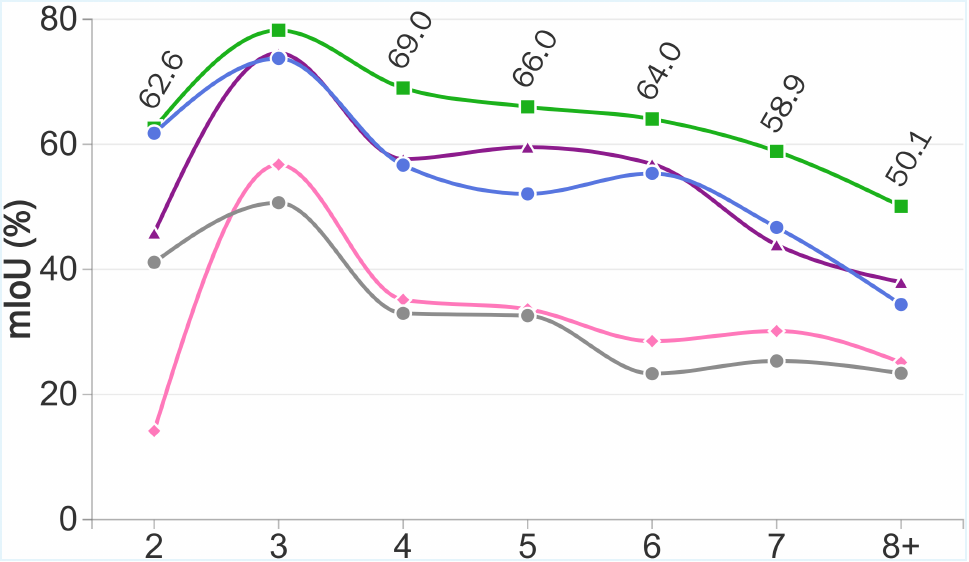}
        \vspace{-.5cm}
        \caption{\# total objects}
        \label{fig:perf_objects}
    \end{subfigure}
    \hfill
    \begin{subfigure}[b]{0.32\textwidth}
        \centering
        \includegraphics[width=\textwidth]{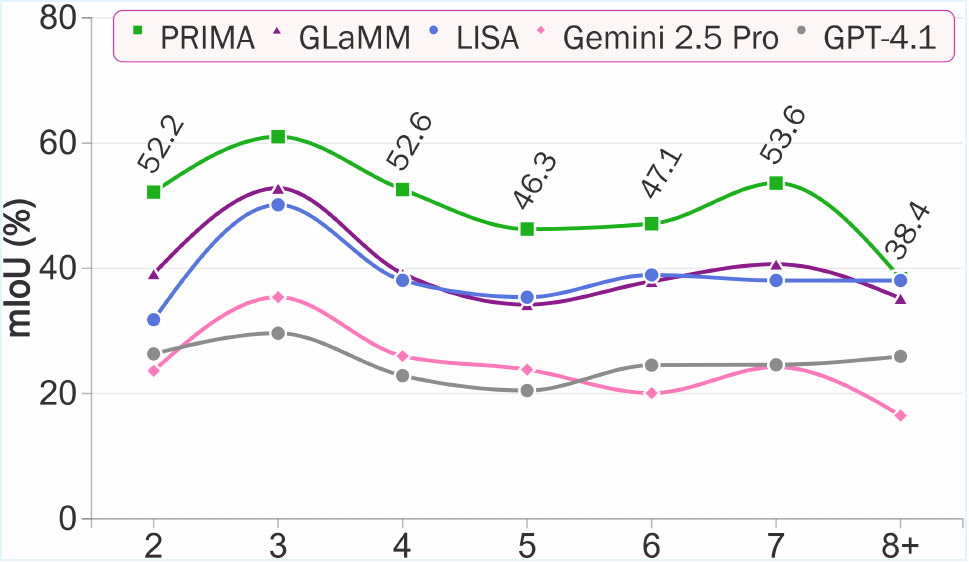}
        \vspace{-.5cm}
        \caption{\# target masks}
        \label{fig:perf_masks}
    \end{subfigure}
    \hfill
    \begin{subfigure}[b]{0.32\textwidth}
        \centering
        \includegraphics[width=\textwidth]{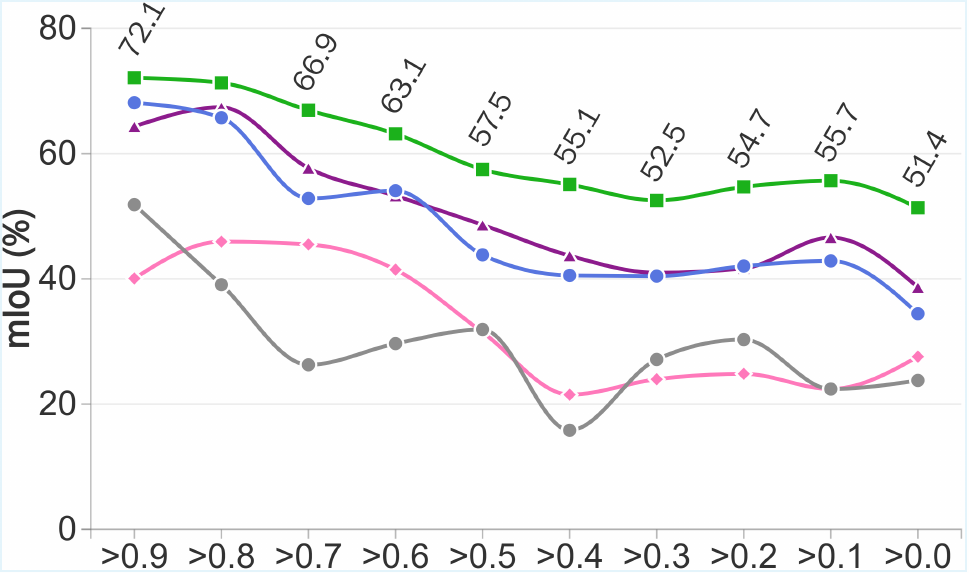}
        \vspace{-.5cm}
        \caption{object visibility}
        \label{fig:perf_visibility}
    \end{subfigure}
    }
    \vspace{-0.6cm}
    \caption{\textbf{Performance comparison for (a) \# objects and (b) \# target masks in images, and (c) visibility of objects.}}
    \label{fig:perf_ablation}
    \vspace{-0.1cm}
\end{figure*}

\subsection{Ablations}
\noindent \textbf{Number of objects in the images}:
\Cref{fig:perf_objects} analyzes model robustness as scene complexity increases.
Performance generally declines as the object count grows,
indicating that crowded scenes remain challenging for all models.
However, \modelname{} consistently outperforms prior methods across all settings.
Notably, while the improvement over
the next best baseline is within $\sim$5 points in mIoU for images containing 2-3 objects, the performance gap widens in denser scenes with significant absolute gains of $11.35$ points in mIoU at 4 objects and $12.50$ at 8+ objects. \modelname{} exhibits a slower degradation trend, maintaining
$\sim50\%$ mIoU even in the most crowded settings, whereas competing models
drop below $40\%$ mIoU.
Results confirm that explicitly modeling cross-image interactions enables
more robust reasoning in complex visual environments with many candidate objects.

\noindent \textbf{Number of target masks}:
\Cref{fig:perf_masks} analyzes performance as the target mask count increases. Although performance generally decreases with more target masks, \modelname{} consistently outperforms baselines across all settings. In particular, margins over GLaMM remain substantial for moderate mask counts (2–7 targets), reaching absolute improvements of $13.42$ mIoU points at 4 masks. This suggests that \modelname{} effectively handles scenarios requiring multiple instances, a key
requirement for complex multi-image reasoning.

\noindent \textbf{Object visibility}:
\Cref{fig:perf_visibility} analyzes performance as a function of object
visibility. Object visibility is computed as the ratio of
the object mask area to the total image area, weighted by the relative scale
disparity among objects within a scene. Performance generally decreases as object
visibility becomes lower, since smaller or partially visible objects provide
weaker visual cues for grounding.  Despite this increasing difficulty,
\modelname{} consistently outperforms baselines across all visibility
ranges. The improvement is particularly pronounced for low-visibility objects,
where \modelname{} maintains a margin between $9.06$ and $13.01\%$ mIoU over prior
methods on instances with visibility below $0.4$, suggesting that explicitly modeling cross-image context helps resolve
ambiguous or weak visual signals.

\begin{table}[t!]
\centering
\caption{\textbf{Ablation on alternatives to \modulename{}}.}
\setlength\tabcolsep{2pt}
\vspace{-0.2cm}
\resizebox{\linewidth}{!}{
\begin{tabular}{lcccccccc}
\toprule
\textbf{Model} & \textbf{mIoU$\uparrow$} & \textbf{Recall$\uparrow$} & \textbf{I-SS$\uparrow$} & \textbf{I-SIoU$\uparrow$}  & \textbf{SS$\uparrow$} & \textbf{S-IoU$\uparrow$} & \textbf{METEOR$\uparrow$} \\ 
\midrule
\xmark{} \modulename{} \cmark{} Pooling & 48.56 & 32.13 & 49.67 & 39.6 & 71.73 & 58.17 & 28.79\\
\xmark{} \modulename{} \cmark{} Projection & 48.98 & 34.86 & 55.67 & 51.87 & 78.81 & 72.61 & 28.73\\
\midrule 
\cc \textbf{\modelname{}} (ours) & \cc 54.95 & \cc 41.33 & \cc 64.49 & \cc 61.48 & \cc 83.19 & \cc 78.76 & \cc 28.81 \\

\midrule
\textcolor{DarkSageGreen}{$\Delta$\textbf{Ours - 2nd best}} & 
\textcolor{DarkSageGreen}{\textbf{+}$\uparrow$}5.97  & 
\textcolor{DarkSageGreen}{\textbf{+}$\uparrow$}6.47  & 
\textcolor{DarkSageGreen}{\textbf{+}$\uparrow$}8.82 & 
\textcolor{DarkSageGreen}{\textbf{+}$\uparrow$}9.61 &
\textcolor{DarkSageGreen}{\textbf{+}$\uparrow$}4.38 &
\textcolor{DarkSageGreen}{\textbf{+}$\uparrow$}6.15  &
\textcolor{DarkSageGreen}{\textbf{+}$\uparrow$}0.02  \\
\bottomrule
 
\end{tabular}
}
\label{tab:ablation}
\vspace{-0.2cm}
\end{table}

\noindent \textbf{\modulename{} module}: \Cref{tab:ablation} compares \modulename{} with two simpler cross-image baselines:
(1) Pooling, which replaces each image's embeddings with a pooled representation, and (2) Projection, which concatenates embeddings along the hidden dimension, projects them back to the original dimension, and splits them to their original sizes. Across all metrics, our attention-based \modulename{} consistently performs best, with absolute gains up to \textcolor{DarkSageGreen}{\textbf{+}$\uparrow$} $5.97$ points in mIoU, \textcolor{DarkSageGreen}{\textbf{+}$\uparrow$} $6.47$ Recall, \textcolor{DarkSageGreen}{\textbf{+}$\uparrow$} $8.82$ I-SS and \textcolor{DarkSageGreen}{\textbf{+}$\uparrow$} $9.61$ I-SIoU. These results indicate that explicitly modeling cross-image interactions benefits both segmentation quality and the generated text.

\noindent Appendix \ref{app:additional_exps} provides further analyses of image assignment and mask cardinality, along with controlled tests of cross-image reasoning.

\subsection{Qualitative Analysis}
\Cref{tab:config_comparison} compares \modelname{} with the strongest baseline, GLaMM. Each row shows the inputs/question, ground truth, and model responses/masks, with matching colors linking object mentions to masks.
In the first example, \modelname{} correctly identifies both tables, whereas GLaMM selects a cabinet.
The second demonstrates \modelname{}’s robust counting and segmentation ability, even when target objects are only partially visible (Images 1 and 2).
Overall, \modelname{} produces cleaner masks and better grounding alignment. The final row shows a failure case: in a crowded scene, \modelname{} gets the categories/count correct but does not cleanly separate instances, while GLaMM misidentifies the horse as a dog; additional qualitative examples in Appendix \ref{suppl:more-qualitatives}.
\section{Conclusion}
In this work, we introduce the task of \emph{multi-image pixel-grounded reasoning}, which requires models to jointly reason across multiple disjoint images and ground their explanations in segmentation masks. 
We curate \benchmarkname{}, a large-scale dataset of multi-image reasoning segmentation QA pairs, and propose \modelname{}, an LVLM that captures cross-image interactions via our \modulename{} module, enabling efficient relational reasoning across disjoint scenes. Experiments show \modelname{} outperforms strong LVLM and reasoning-segmentation baselines on grounding and language metrics while remaining computationally efficient.

\section*{Acknowledgments}
This research is based upon work supported by the NSF CAREER Award \#2542328 and the U.S. Defense Advanced Research Projects Agency (DARPA) under awards HR001125C0303 and HR00112390062. The views and conclusions contained herein are those of the authors and should not be interpreted as necessarily representing the official policies, either expressed or implied, of NSF, DARPA, or the U.S. Government. The U.S. Government is authorized to reproduce and distribute reprints for governmental purposes notwithstanding any copyright annotation therein. This work used the Delta GPUs provided by the National Center for Supercomputing Applications through allocations CIS240752, CIS240817, and CIS250130 from the Advanced Cyberinfrastructure Coordination Ecosystem: Services \& Support (ACCESS, \citet{boerner2023access}) program, supported by National Science Foundation grants \#2138259, \#2138286, \#2138307, \#2137603, and \#2138296.

{
    \small
    \bibliographystyle{ieeenat_fullname}
    \bibliography{main}

@String(IJCV  = {International Journal of Computer Vision (IJCV)})

@String(CVPR  = {IEEE Conference on Computer Vision and Pattern Recognition (CVPR)})

@String(ICCV  = {International Conference on Computer Vision (ICCV)})

@String(ECCV  = {European Conference on Computer Vision (ECCV)})

@String(NeurIPS = {Advances in Neural Information Processing Systems (NeurIPS)})

@String(ICML  = {International Conference on Machine Learning (ICML)})

@String(ICLR  = {International Conference on Learning Representations (ICLR)})

@String(ACCV  = {Asian Conference on Computer Vision (ACCV)})

@String(CVPRW = {IEEE Conf. on Computer Vision and Pattern Recognition Workshops (CVPRW)})

@String(AAAI  = {AAAI Conference on Artificial Intelligence (AAAI)})

@String(ICIP  = {IEEE International Conference on Image Processing (ICIP)})

@String(TMLR  = {Transactions on Machine Learning Research (TMLR)})

@String(TIP   = {IEEE Transactions on Image Processing (TIP)})

@String(TMM   = {IEEE Transactions on Multimedia (TMM)})

@String(FINDINGSACL = {Findings of the Association for Computational Linguistics (Findings of ACL)})

@String(THREEDV = {International Conference on 3D Vision (3DV)})

@inproceedings{huang2024sparkles,
  title={{Sparkles: Unlocking Chats Across Multiple Images for Multimodal Instruction-Following Models}},
  author={Huang, Yupan and Meng, Zaiqiao and Liu, Fangyu and Su, Yixuan and Collier, Nigel and Lu, Yutong},
  booktitle={{ICLR Workshop on Navigating and Addressing Data Problems for Foundation Models}},
year = 2024
}

@inproceedings{li2024fine,
  title={{Fine-tuning Multimodal LLMs to Follow Zero-shot Demonstrative Instructions}},
  author={Li, Juncheng and Pan, Kaihang and Ge, Zhiqi and Gao, Minghe and Ji, Wei and Zhang, Wenqiao and Chua, Tat-Seng and Tang, Siliang and Zhang, Hanwang and Zhuang, Yueting},
  booktitle=ICLR,
  year={2024}
}

@inproceedings{lin2025comparison,
  title={{Comparison Visual Instruction Tuning}},
  author={Lin, Wei and Mirza, Muhammad Jehanzeb and Doveh, Sivan and Feris, Rogerio and Giryes, Raja and Hochreiter, Sepp and Karlinsky, Leonid},
  booktitle=CVPRW,
  pages={2998--3008},
  year={2025}
}

@article{yuan2025instruction,
  title={{Instruction-Guided Multi-Granularity Segmentation and Captioning with Large Multimodal Model}},
  author={Yuan, Xu and Zhou, Li and Sun, Zenghui and Zhou, Zikun and Lan, Jinsong},
  journal=AAAI,
  volume={39},
  number={9},
  pages={9725--9733},
  year={2025}
}

@inproceedings{wang2024llm,
  title={{LLM-Seg: Bridging Image Segmentation and Large Language Model Reasoning}},
  author={Wang, Junchi and Ke, Lei},
  booktitle=CVPRW,
  pages={{1765--1774}},
  year={2024}
}

@inproceedings{zhang2024groundhog,
  title={{GROUNDHOG: Grounding Large Language Models to Holistic Segmentation}},
  author={Zhang, Yichi and Ma, Ziqiao and Gao, Xiaofeng and Shakiah, Suhaila and Gao, Qiaozi and Chai, Joyce},
  booktitle=CVPR,
  pages={{14227--14238}},
  year={2024}
}

@inproceedings{guo2024regiongpt,
  title={{RegionGPT: Towards Region Understanding Vision Language Model}},
  author={Guo, Qiushan and De Mello, Shalini and Yin, Hongxu and Byeon, Wonmin and Cheung, Ka Chun and Yu, Yizhou and Luo, Ping and Liu, Sifei},
  booktitle={{Proceedings of the IEEE/CVF Conference on Computer Vision and Pattern Recognition}},
  pages={{13796--13806}},
  year={2024}
}

@inproceedings{ma2024groma,
  title={{Groma: Localized Visual Tokenization for Grounding Multimodal Large Language Models}},
  author={Ma, Chuofan and Jiang, Yi and Wu, Jiannan and Yuan, Zehuan and Qi, Xiaojuan},
  booktitle=ECCV,
  pages={{417--435}},
  year={2024}}

@inproceedings{wu2024see,
  title={{See Say and Segment: Teaching LMMs to Overcome False Premises}},
  author={Wu, Tsung-Han and Biamby, Giscard and Chan, David and Dunlap, Lisa and Gupta, Ritwik and Wang, Xudong and Gonzalez, Joseph E and Darrell, Trevor},
  booktitle=CVPR,
  pages={{13459--13469}},
  year={2024}
}

@inproceedings{wang2024unveiling,
  title={{Unveiling Parts Beyond Objects: Towards Finer-Granularity Referring Expression Segmentation}},
  author={Wang, Wenxuan and Yue, Tongtian and Zhang, Yisi and Guo, Longteng and He, Xingjian and Wang, Xinlong and Liu, Jing},
  booktitle=CVPR,
  pages={{12998--13008}},
  year={{2024}}
}

@inproceedings{ye2025mplug,
  title={{mPLUG-Owl3: Towards Long Image-Sequence Understanding in Multi-Modal Large Language Models}},
  author={Ye, Jiabo and Xu, Haiyang and Liu, Haowei and Hu, Anwen and Yan, Ming and Qian, Qi and Zhang, Ji and Huang, Fei and Zhou, Jingren},
  booktitle=ICLR,
  year={2025}
}

@inproceedings{li2024semantic,
  title={{Segment and Recognize Anything at Any Granularity}},
  author={Li, Feng and Zhang, Hao and Sun, Peize and Zou, Xueyan and Liu, Shilong and Li, Chunyuan and Yang, Jianwei and Zhang, Lei and Gao, Jianfeng},
  booktitle=ECCV,
  pages={467--484},
  year={2024},
  organization={Springer}
}

@inproceedings{lai2024lisa,
  title={{LISA: Reasoning Segmentation via Large Language Model}},
  author={Lai, Xin and Tian, Zhuotao and Chen, Yukang and Li, Yanwei and Yuan, Yuhui and Liu, Shu and Jia, Jiaya},
  booktitle=CVPR,
  pages={{9579--9589}},
  year={2024}
}

@article{liu2024visual,
  title={{Visual Instruction Tuning}},
  author={Liu, Haotian and Li, Chunyuan and Wu, Qingyang and Lee, Yong Jae},
  journal=NeurIPS,
  volume={{36}},
  year={2023}
}

@inproceedings{li2023blip,
  title={{BLIP-2: Bootstrapping Language-Image Pre-training with Frozen Image Encoders and Large Language Models}},
  author={Li, Junnan and Li, Dongxu and Savarese, Silvio and Hoi, Steven},
  booktitle=ICML,
  pages={{19730--19742}},
  year={2023}}

@article{alayrac2022flamingo,
  title={{Flamingo: a Visual Language Model for Few-Shot Learning}},
  author={Alayrac, Jean-Baptiste and Donahue, Jeff and Luc, Pauline and Miech, Antoine and Barr, Iain and Hasson, Yana and Lenc, Karel and Mensch, Arthur and Millican, Katherine and Reynolds, Malcolm and others},
  journal=NeurIPS,
  volume={{35}},
  pages={{23716--23736}},
  year={2022}
}

@inproceedings{susladkar2026best,
  title={Best of Both Worlds: Multimodal Reasoning and Generation via Unified Discrete Flow Matching},
  author={Susladkar, Onkar Kishor and Prakash, Tushar and Deshmukh, Gayatri and Nguyen, Kiet A. and Zhang, Jiaxun and Juvekar, Adheesh Sunil and Bao, Tianshu and Chai, Lin and Mittal, Sparsh and Dhillon, Inderjit S and Lourentzou, Ismini},
  booktitle={International Conference on Machine Learning},
  year={2026}
}

@inproceedings{li2026counterfactual,
  title={Counterfactual segmentation reasoning: Diagnosing and mitigating pixel-grounding hallucination},
  author={Li, Xinzhuo and Juvekar, Adheesh and Zhang, Jiaxun and Liu, Xingyou and Wahed, Muntasir and Nguyen, Kiet A and Shen, Yifan and Yu, Tianjiao and Lourentzou, Ismini},
booktitle = {Proceedings of the IEEE/CVF Conference on Computer Vision and Pattern Recognition (CVPR) Findings},
  pages={7450--7460},
  year={2026}
}

@inproceedings{karwande2022chexrelnet,
  title={Chexrelnet: An anatomy-aware model for tracking longitudinal relationships between chest x-rays},
  author={Karwande, Gaurang and Mbakwe, Amarachi B and Wu, Joy T and Celi, Leo A and Moradi, Mehdi and Lourentzou, Ismini},
  booktitle={International Conference on Medical Image Computing and Computer-Assisted Intervention},
  pages={581--591},
  year={2022},
  organization={Springer}
}

@article{wu2021chest,
  title={Chest imagenome dataset for clinical reasoning},
  author={Wu, Joy T and Agu, Nkechinyere N and Lourentzou, Ismini and Sharma, Arjun and Paguio, Joseph A and Yao, Jasper S and Dee, Edward C and Mitchell, William and Kashyap, Satyananda and Giovannini, Andrea and others},
  journal={arXiv preprint arXiv:2108.00316},
  year={2021}
}

@inproceedings{crall2013hotspotter,
  title={Hotspotter—patterned species instance recognition},
  author={Crall, Jonathan P and Stewart, Charles V and Berger-Wolf, Tanya Y and Rubenstein, Daniel I and Sundaresan, Siva R},
  booktitle={2013 IEEE workshop on applications of computer vision (WACV)},
  pages={230--237},
  year={2013},
  organization={IEEE}
}

@article{pippi2025tofu,
  title={ToFu: Visual Tokens Reduction via Fusion for Multi-modal, Multi-patch, Multi-image Task},
  author={Pippi, Vittorio and Guillaumin, Matthieu and Cascianelli, Silvia and Cucchiara, Rita and Jaritz, Maximilian and Bazzani, Loris},
  journal={arXiv preprint arXiv:2503.04444},
  year={2025}
}

@inproceedings{mbakwe2023hierarchical,
  title={Hierarchical vision transformers for disease progression detection in chest x-ray images},
  author={Mbakwe, Amarachi B and Wang, Lyuyang and Moradi, Mehdi and Lourentzou, Ismini},
  booktitle={International Conference on Medical Image Computing and Computer-Assisted Intervention},
  pages={685--695},
  year={2023},
  organization={Springer}
}

@inproceedings{zhu2025change3d,
  title={Change3D: Revisiting change detection and captioning from a video modeling perspective},
  author={Zhu, Duowang and Huang, Xiaohu and Huang, Haiyan and Zhou, Hao and Shao, Zhenfeng},
  booktitle={2025 IEEE/CVF Conference on Computer Vision and Pattern Recognition (CVPR)},
  pages={24011--24022},
  year={2025},
  organization={IEEE}
}

@article{li2026toward,
  title={Toward cognitive supersensing in multimodal large language model},
  author={Li, Boyi and Shen, Yifan and Liu, Yuanzhe and Xu, Yifan and Liu, Jiateng and Li, Xinzhuo and Li, Zhengyuan and Zhu, Jingyuan and Zhong, Yunhan and Lan, Fangzhou and Cao, Jianguo and Rehg, James M. and Ji, Heng and Lourentzou, Ismini and Cao, Xu},
  journal={arXiv preprint arXiv:2602.01541},
  year={2026}
}

@inproceedings{shen2026fine,
  title={Fine-grained preference optimization improves spatial reasoning in vlms},
  author={Shen, Yifan and Liu, Yuanzhe and Zhu, Jingyuan and Cao, Xu and Zhang, Xiaofeng and He, Yixiao and Ye, Wenming and Rehg, James and Lourentzou, Ismini},
  booktitle=NeurIPS,
  year={2025}
}

@inproceedings{ogunleye20263d,
  title={3d-vcd: Hallucination mitigation in 3d-llm embodied agents through visual contrastive decoding},
  author={Ogunleye, Makanjuola Adekunmi and Abdelrahman, Eman and Lourentzou, Ismini},
  booktitle=CVPR,
  year={2026}
}

@inproceedings{peng2024kosmos,
  title={{Grounding Multimodal Large Language Models to the World}},
  author={Peng, Zhiliang and Wang, Wenhui and Dong, Li and Hao, Yaru and Huang, Shaohan and Ma, Shuming and Ye, Qixiang and Wei, Furu},
  booktitle=ICLR,
  year={2024}
}

@inproceedings{you2024ferret,
  title={{Ferret: Refer and ground anything anywhere at any granularity}},
  author={You, Haoxuan and Zhang, Haotian and Gan, Zhe and Du, Xianzhi and Zhang, Bowen and Wang, Zirui and Cao, Liangliang and Chang, Shih-Fu and Yang, Yinfei},
  booktitle=ICLR,
  year={2024}
}

@article{zhang2023gpt4roi,
  title={{GPT4RoI: Instruction Tuning Large Language Model on Region-of-Interest}},
  author={Zhang, Shilong and Sun, Peize and Chen, Shoufa and Xiao, Min and Shao, Wenqi and Zhang, Wenwei and Liu, Yu and Chen, Kai and Luo, Ping},
  journal={{arXiv preprint arXiv:2307.03601}},
  year={2023}
}

@inproceedings{kirillov2023segment,
  title={{Segment Anything}},
  author={Kirillov, Alexander and Mintun, Eric and Ravi, Nikhila and Mao, Hanzi and Rolland, Chloe and Gustafson, Laura and Xiao, Tete and Whitehead, Spencer and Berg, Alexander C and Lo, Wan-Yen and others},
  booktitle=ICCV,
  pages={{4015--4026}},
  year={2023}
}

@inproceedings{zhang2024next,
  title={{NExT-Chat: An LMM for Chat, Detection and Segmentation}},
  author={Zhang, Ao and Yao, Yuan and Ji, Wei and Liu, Zhiyuan and Chua, Tat-Seng},
  booktitle=ICML,
  pages={60116--60133},
  year={2024}
}

@inproceedings{zhang2024omg,
  title={{OMG-LLaVA: Bridging Image-level, Object-level, Pixel-level Reasoning and Understanding}},
  author={Zhang, Tao and Li, Xiangtai and Fei, Hao and Yuan, Haobo and Wu, Shengqiong and Ji, Shunping and Loy, Chen Change and Yan, Shuicheng},
  booktitle=NeurIPS,
  pages={71737--71767},
  year={2024}
}

@article{chen2023shikra,
  title={{Shikra: Unleashing Multimodal LLM’s Referential Dialogue Magic}},
  author={Chen, Keqin and Zhang, Zhao and Zeng, Weili and Zhang, Richong and Zhu, Feng and Zhao, Rui},
  journal={{arXiv preprint arXiv:2306.15195}},
  year={2023}
}

@inproceedings{wang2022ofa,
  title={{OFA: Unifying Architectures, Tasks, and Modalities Through a Simple Sequence-to-Sequence Learning Framework}},
  author={Wang, Peng and Yang, An and Men, Rui and Lin, Junyang and Bai, Shuai and Li, Zhikang and Ma, Jianxin and Zhou, Chang and Zhou, Jingren and Yang, Hongxia},
  booktitle=ICML,
  pages={{23318--23340}},
  year={2022}}

@inproceedings{cheng2401yolo,
  title={{YOLO-World: Real-Time Open-Vocabulary Object Detection}},
  author={Cheng, Tianheng and Song, Lin and Ge, Yixiao and Liu, Wenyu and Wang, Xinggang and Shan, Ying},
  booktitle=CVPR,
  year={2024}
}

@inproceedings{ren2024pixellm,
  title={{PixelLM: Pixel Reasoning with Large Multimodal Model}},
  author={Ren, Zhongwei and Huang, Zhicheng and Wei, Yunchao and Zhao, Yao and Fu, Dongmei and Feng, Jiashi and Jin, Xiaojie},
  booktitle=CVPR,
  pages={{26374--26383}},
  year={2024}
}

@inproceedings{rasheed2024glamm,
  title={{GLaMM: Pixel Grounding Large Multimodal Model}},
  author={Rasheed, Hanoona and Maaz, Muhammad and Shaji, Sahal and Shaker, Abdelrahman and Khan, Salman and Cholakkal, Hisham and Anwer, Rao M and Xing, Eric and Yang, Ming-Hsuan and Khan, Fahad S},
  booktitle=CVPR,
  pages={{13009--13018}},
  year={2024}
}

@inproceedings{zhang2025psalm,
  title={{PSALM: Pixelwise SegmentAtion with Large Multi-Modal Model}},
  author={Zhang, Zheng and Ma, Yeyao and Zhang, Enming and Bai, Xiang},
  booktitle=ECCV,
  pages={74--91},
  year={2025}
}

@article{oquab2023dinov2,
  title={{DINOv2: Learning Robust Visual Features without Supervision}},
  author={Oquab, Maxime and Darcet, Timoth{\'e}e and Moutakanni, Th{\'e}o and Vo, Huy and Szafraniec, Marc and Khalidov, Vasil and Fernandez, Pierre and Haziza, Daniel and Massa, Francisco and El-Nouby, Alaaeldin and others},
  journal=TMLR,
  year={2024}
}

@inproceedings{yuan2024osprey,
  title={{Osprey: Pixel Understanding with Visual Instruction Tuning}},
  author={Yuan, Yuqian and Li, Wentong and Liu, Jian and Tang, Dongqi and Luo, Xinjie and Qin, Chi and Zhang, Lei and Zhu, Jianke},
  booktitle=CVPR,
  pages={{28202--28211}},
  year={2024}
}

@inproceedings{chen2014detect,
  title={{Detect What You Can: Detecting and Representing Objects using Holistic Models and Body Parts}},
  author={Chen, Xianjie and Mottaghi, Roozbeh and Liu, Xiaobai and Fidler, Sanja and Urtasun, Raquel and Yuille, Alan},
  booktitle=CVPR,
  pages={{1971--1978}},
  year={2014}
}

@inproceedings{zhou2017scene,
  title={{Scene Parsing through ADE20K Dataset}},
  author={Zhou, Bolei and Zhao, Hang and Puig, Xavier and Fidler, Sanja and Barriuso, Adela and Torralba, Antonio},
  booktitle=CVPR,
  pages={{633--641}},
  year={2017}
}

@inproceedings{ramanathan2023paco,
  title={{PACO: Parts and Attributes of Common Objects}},
  author={Ramanathan, Vignesh and Kalia, Anmol and Petrovic, Vladan and Wen, Yi and Zheng, Baixue and Guo, Baishan and Wang, Rui and Marquez, Aaron and Kovvuri, Rama and Kadian, Abhishek and others},
  booktitle=CVPR,
  pages={{7141--7151}},
  year={2023}
}

@article{li2024llava,
  title={{LLaVA-NeXT-Interleave: Tackling Multi-image, Video, and 3D in Large Multimodal Models}},
  author={Li, Feng and Zhang, Renrui and Zhang, Hao and Zhang, Yuanhan and Li, Bo and Li, Wei and Ma, Zejun and Li, Chunyuan},
  journal={{arXiv preprint arXiv:2407.07895}},
  year={2024}
}

@article{jia2025leopard,
  title={{LEOPARD: A Vision Language Model For Text-Rich Multi-Image Tasks}},
  author={Jia, Mengzhao and Yu, Wenhao and Ma, Kaixin and Fang, Tianqing and Zhang, Zhihan and Ouyang, Siru and Zhang, Hongming and Yu, Dong and Jiang, Meng},
  journal=TMLR,
  year={2025}
}

@article{wang2024longllava,
  title={{LongLLaVA: Scaling Multi-modal LLMs to 1000 Images Efficiently via Hybrid Architecture}},
  author={Wang, Xidong and Song, Dingjie and Chen, Shunian and Zhang, Chen and Wang, Benyou},
  journal={{arXiv preprint arXiv:2409.02889}},
  year={2024}
}

@article{jiang2024mantis,
  title={{MANTIS: Interleaved Multi-Image Instruction Tuning}},
  author={Dongfu Jiang and Xuan He and Huaye Zeng and Cong Wei and Max W.F. Ku and Qian Liu and Wenhu Chen},
  journal=TMLR,
  year={2024}
}

@inproceedings{radford2021learning,
  title={{Learning Transferable Visual Models From Natural Language Supervision}},
  author={Radford, Alec and Kim, Jong Wook and Hallacy, Chris and Ramesh, Aditya and Goh, Gabriel and Agarwal, Sandhini and Sastry, Girish and Askell, Amanda and Mishkin, Pamela and Clark, Jack and others},
  booktitle=ICML,
  pages={{8748--8763}},
  year={2021}}

@article{dai2023instructblip,
  title={{InstructBLIP: Towards General-purpose Vision-Language Models with Instruction Tuning}},
  author={Dai, Wenliang and Li, Junnan and Li, Dongxu and Tiong, Anthony and Zhao, Junqi and Wang, Weisheng and Li, Boyang and Fung, Pascale N and Hoi, Steven},
  journal=NeurIPS,
  volume={36},
  pages={49250--49267},
  year={2023}
}

@article{sun2023eva,
  title={{EVA-CLIP: Improved Training Techniques for CLIP at Scale}},
  author={Sun, Quan and Fang, Yuxin and Wu, Ledell and Wang, Xinlong and Cao, Yue},
  journal={{arXiv preprint arXiv:2303.15389}},
  year={2023}
}

@inproceedings{milletari2016v,
  title={{V-Net: Fully Convolutional Neural Networks for Volumetric Medical Image Segmentation}},
  author={Milletari, Fausto and Navab, Nassir and Ahmadi, Seyed-Ahmad},
  booktitle=THREEDV,
  pages={{565--571}},
  year={2016}}

@inproceedings{lin2017focal,
  title={{Focal Loss for Dense Object Detection}},
  author={Lin, Tsung-Yi and Goyal, Priya and Girshick, Ross and He, Kaiming and Doll{\'a}r, Piotr},
  booktitle=ICCV,
  pages={2980--2988},
  year={2017}
}

@article{wei2024ov,
  title={{OV-PARTS: Towards Open-Vocabulary Part Segmentation}},
  author={Wei, Meng and Yue, Xiaoyu and Zhang, Wenwei and Kong, Shu and Liu, Xihui and Pang, Jiangmiao},
  journal=NeurIPS,
  volume={{36}},
  year={2023}
}

@inproceedings{jang2025mmr,
  title={{MMR: A Large-scale Benchmark Dataset for Multi-target and Multi-granularity Reasoning Segmentation}},
  author={Jang, Donggon and Cho, Yucheol and Lee, Suin and Kim, Taehyeon and Kim, Daeshik},
  booktitle=ICLR,
  year={{2025}}
}

@article{kil2024compbench,
    title={{MLLM-CompBench: A Comparative Reasoning Benchmark for Multimodal LLMs}},
    author={Kil, Jihyung and Mai, Zheda and Lee, Justin and Chowdhury, Arpita and Wang, Zihe and Cheng, Kerrie and 
      Wang, Lemeng and Liu, Ye and Chao, Wei-Lun},
    journal=NeurIPS,
    year={2024}
  }

@article{wu2024mmra,
  title={{MMRA: A Benchmark for Multi-granularity Multi-image Relational Association}},
  author={Wu, Siwei and Zhu, Kang and Bai, Yu and Liang, Yiming and Li, Yizhi and Wu, Haoning and Liu, Jiaheng and Liu, Ruibo and Qu, Xingwei and Cheng, Xuxin and others},
  journal={{arXiv preprint arXiv:2407.17379}},
  year={2024}
}

@article{conti2023vocabulary,
  title={{Vocabulary-free Image Classification}},
  author={Conti, Alessandro and Fini, Enrico and Mancini, Massimiliano and Rota, Paolo and Wang, Yiming and Ricci, Elisa},
  journal=NeurIPS,
  volume={{36}},
  pages={{30662--30680}},
  year={2023}
}

@inproceedings{reimers2019sentence,
  title={{Sentence-BERT: Sentence Embeddings using Siamese BERT-Networks}},
  author={Reimers, Nils and Gurevych, Iryna},
  booktitle = "Proceedings of the Conference on Empirical Methods in Natural Language Processing and the International Joint Conference on Natural Language Processing (EMNLP-IJCNLP)",
  year={2019}
}

@article{kazemi2024remi,
  title={{ReMI: A Dataset for Reasoning with Multiple Images}},
  author={Kazemi, Mehran and Dikkala, Nishanth and Anand, Ankit and Devic, Petar and Dasgupta, Ishita and Liu, Fangyu and Fatemi, Bahare and Awasthi, Pranjal and Guo, Dee and Gollapudi, Sreenivas and others},
  journal=NeurIPS,
  volume={37},
  pages={60088--60109},
  year={2024}
}

@inproceedings{nguyen2025calico,
  title={{CALICO: Part-Focused Semantic Co-Segmentation with Large Vision-Language Models}},
  author={Nguyen, Kiet A and Juvekar, Adheesh and Yu, Tianjiao and Wahed, Muntasir and Lourentzou, Ismini},
  booktitle=CVPR,
  pages={4550--4561},
  year={2025}
}

@article{liu2024mmdu,
  title={{MMDU: A Multi-Turn Multi-Image Dialog Understanding Benchmark and Instruction-Tuning Dataset for LVLMs}},
  author={Liu, Ziyu and Chu, Tao and Zang, Yuhang and Wei, Xilin and Dong, Xiaoyi and Zhang, Pan and Liang, Zijian and Xiong, Yuanjun and Qiao, Yu and Lin, Dahua and others},
  journal=NeurIPS,
  volume={37},
  pages={8698--8733},
  year={2024}
}

@inproceedings{meng2025mmiu,
  title={{MMIU: Multimodal Multi-image Understanding for Evaluating Large Vision-Language Models}},
  author={Meng, Fanqing and Wang, Jin and Li, Chuanhao and Lu, Quanfeng and Tian, Hao and Yang, Tianshuo and Liao, Jiaqi and Zhu, Xizhou and Dai, Jifeng and Qiao, Yu and others},
  booktitle=ICLR,
  year={2025}
}

@inproceedings{li2025migician,
  title={{Migician: Revealing the Magic of Free-Form Multi-Image Grounding in Multimodal Large Language Models}},
  author={Li, You and Huang, Heyu and Chen, Chi and Huang, Kaiyu and Huang, Chao and Guo, Zonghao and Liu, Zhiyuan and Xu, Jinan and Li, Yuhua and Li, Ruixuan and Sun, Maosong},
  booktitle=FINDINGSACL,
  year={2025}
}

@inproceedings{xu2025mc,
  title={{MC-Bench: A Benchmark for Multi-Context Visual Grounding in the Era of MLLMs}},
  author={Xu, Yunqiu and Zhu, Linchao and Yang, Yi},
  booktitle=ICCV,
  pages={17675--17687},
  year={2025}
}

@inproceedings{liu2025omnidiff,
  title={{OmniDiff: A Comprehensive Benchmark for Fine-grained Image Difference Captioning}},
  author={Liu, Yuan and Hou, Saihui and Hou, Saijie and Du, Jiabao and Meng, Shibei and Huang, Yongzhen},
  booktitle=ICCV,
  pages={21440--21449},
  year={2025}
}

@article{ren2024grounded,
  title={{Grounded SAM: Assembling Open-World Models for Diverse Visual Tasks}},
  author={Ren, Tianhe and Liu, Shilong and Zeng, Ailing and Lin, Jing and Li, Kunchang and Cao, He and Chen, Jiayu and Huang, Xinyu and Chen, Yukang and Yan, Feng and others},
  journal={arXiv preprint arXiv:2401.14159},
  year={2024}
}

@article{zhu2025internvl3,
  title={{InternVL3: Exploring Advanced Training and Test-Time Recipes for Open-Source Multimodal Models}},
  author={Zhu, Jinguo and Wang, Weiyun and Chen, Zhe and Liu, Zhaoyang and Ye, Shenglong and Gu, Lixin and Tian, Hao and Duan, Yuchen and Su, Weijie and Shao, Jie and others},
  journal={arXiv preprint arXiv:2504.10479},
  year={2025}
}

@article{bai2025qwen2vl,
  title={{Qwen2.5-VL Technical Report}},
  author={Bai, Shuai and Chen, Keqin and Liu, Xuejing and Wang, Jialin and Ge, Wenbin and Song, Sibo and Dang, Kai and Wang, Peng and Wang, Shijie and Tang, Jun and others},
  journal={arXiv preprint arXiv:2502.13923},
  year={2025}
}

@misc{mistral2024pixtralLarge,
  title        = {{Pixtral Large}},
  author       = {{Mistral AI Team}},
  howpublished = {Mistral AI blog},
  year         = {2024},
  url          = {https://mistral.ai/news/pixtral-large}
}

@article{comanici2025gemini,
  title={{Gemini 2.5: Pushing the Frontier with Advanced Reasoning, Multimodality, Long Context, and Next Generation Agentic Capabilities}},
  author={Comanici, Gheorghe and Bieber, Eric and Schaekermann, Mike and Pasupat, Ice and Sachdeva, Noveen and Dhillon, Inderjit and Blistein, Marcel and Ram, Ori and Zhang, Dan and Rosen, Evan and others},
  journal={arXiv preprint arXiv:2507.06261},
  year={2025}
}

@inproceedings{zheng2025villa,
  title={{ViLLa: Video Reasoning Segmentation with Large Language Model}},
  author={Zheng, Rongkun and Qi, Lu and Chen, Xi and Wang, Yi and Wang, Kun and Zhao, Hengshuang},
  booktitle=ICCV,
  pages={23667--23677},
  year={2025}
}

@inproceedings{wei2025hyperseg,
  title={{HyperSeg: Hybrid Segmentation Assistant with Fine-grained Visual Perceiver}},
  author={Wei, Cong and Zhong, Yujie and Tan, Haoxian and Liu, Yong and Hu, Jie and Li, Dengjie and Zhao, Zheng and Yang, Yujiu},
  booktitle=CVPR,
  pages={8931--8941},
  year={2025}
}

@article{yuan2025sa2va,
  title={{Sa2VA: Marrying SAM2 with LLaVA for Dense Grounded Understanding of Images and Videos}},
  author={Yuan, Haobo and Li, Xiangtai and Zhang, Tao and Huang, Zilong and Xu, Shilin and Ji, Shunping and Tong, Yunhai and Qi, Lu and Feng, Jiashi and Yang, Ming-Hsuan},
  journal={arXiv preprint arXiv:2501.04001},
  year={2025}
}

@inproceedings{munasinghe2025videoglamm,
  title={{VideoGLaMM: A Large Multimodal Model for Pixel-Level Visual Grounding in Videos}},
  author={Munasinghe, Shehan and Gani, Hanan and Zhu, Wenqi and Cao, Jiale and Xing, Eric and Khan, Fahad Shahbaz and Khan, Salman},
  booktitle=CVPR,
  pages={19036--19046},
  year={2025}
}

@inproceedings{zhang2023video,
  title={{Video-LLaMA: An Instruction-tuned Audio-Visual Language Model for Video Understanding}},
  author={Zhang, Hang and Li, Xin and Bing, Lidong},
  booktitle={Conference on Empirical Methods in Natural Language Processing: System Demonstrations},
  pages={543--553},
  year={2023}
}

@article{hurst2024gpt,
  title={{GPT-4o System Card}},
  author={Hurst, Aaron and Lerer, Adam and Goucher, Adam P and Perelman, Adam and Ramesh, Aditya and Clark, Aidan and Ostrow, AJ and Welihinda, Akila and Hayes, Alan and Radford, Alec and others},
  journal={arXiv preprint arXiv:2410.21276},
  year={2024}
}

@article{zheng2025mirg,
  title={{MIRG-RL: Multi-Image Reasoning and Grounding with Reinforcement Learning}},
  author={Zheng, Lihao and Chen, Jiawei and Shen, Xintian and Ma, Hao and Wei, Tao},
  journal={arXiv preprint arXiv:2509.21788},
  year={2025}
}

@inproceedings{lin2025glus,
  title={{GLUS: Global-Local Reasoning Unified into A Single Large Language Model for Video Segmentation}},
  author={Lin, Lang and Yu, Xueyang and Pang, Ziqi and Wang, Yu-Xiong},
  booktitle=CVPR,
  pages={8658--8667},
  year={2025}
}

@inproceedings{gong2025devil,
  title={{The Devil is in Temporal Token: High Quality Video Reasoning Segmentation}},
  author={Gong, Sitong and Zhuge, Yunzhi and Zhang, Lu and Yang, Zongxin and Zhang, Pingping and Lu, Huchuan},
  booktitle=CVPR,
  pages={29183--29192},
  year={2025}
}

@article{zhang2020deep,
  title={{Deep Object Co-Segmentation via Spatial-Semantic Network Modulation}},
  author={Zhang, Kaihua and Chen, Jin and Liu, Bo and Liu, Qingshan},
  journal=AAAI,
  volume={34},
  number={07},
  pages={12813--12820},
  year={2020}
}

@inproceedings{li2019deep,
  title={{Deep Object Co-Segmentation}},
  author={Li, Weihao and Hosseini Jafari, Omid and Rother, Carsten},
  booktitle=ACCV,
  pages={638--653},
  year={2019},
  organization={Springer}
}

@article{zhang2021cyclesegnet,
  title={{CycleSegNet: Object Co-Segmentation With Cycle Refinement and Region Correspondence}},
  author={Zhang, Chi and Li, Guankai and Lin, Guosheng and Wu, Qingyao and Yao, Rui},
  journal=TIP,
  volume={30},
  pages={5652--5664},
  year={2021},
  publisher={IEEE}
}

@article{jerripothula2016image,
  title={{Image Co-segmentation via Saliency Co-fusion}},
  author={Jerripothula, Koteswar Rao and Cai, Jianfei and Yuan, Junsong},
  journal=TMM,
  volume={18},
  number={9},
  pages={1896--1909},
  year={2016},
  publisher={IEEE}
}

@article{wang2017multiple,
  title={{Multiple Semantic Matching on Augmented $N$-Partite Graph for Object Co-Segmentation}},
  author={Wang, Chuan and Zhang, Hua and Yang, Liang and Cao, Xiaochun and Xiong, Hongkai},
  journal=TIP,
  volume={26},
  number={12},
  pages={5825--5839},
  year={2017},
  publisher={IEEE}
}

@inproceedings{rubinstein2013unsupervised,
  title={{Unsupervised Joint Object Discovery and Segmentation in Internet Images}},
  author={Rubinstein, Michael and Joulin, Armand and Kopf, Johannes and Liu, Ce},
  booktitle=CVPR,
  pages={1939--1946},
  year={2013}
}

@inproceedings{faktor2013co,
  title={{Co-Segmentation by Composition}},
  author={Faktor, Alon and Irani, Michal},
  booktitle=ICCV,
  pages={1297--1304},
  year={2013}
}

@inproceedings{rother2006cosegmentation,
  title={{Cosegmentation of Image Pairs by Histogram Matching --- Incorporating a Global Constraint into MRFs}},
  author={Rother, Carsten and Kolmogorov, Vladimir and Minka, Tom and Blake, Andrew},
  booktitle=CVPR,
  volume={1},
  pages={993--1000},
  year={2006},
  organization={IEEE}
}

@inproceedings{shotton2006textonboost,
  title={{TextonBoost: Joint Appearance, Shape and Context Modeling for Multi-Class Object Recognition and Segmentation}},
  author={Shotton, Jamie and Winn, John and Rother, Carsten and Criminisi, Antonio},
  booktitle=ECCV,
  pages={1--15},
  year={2006},
  organization={Springer}
}

@inproceedings{batra2010icoseg,
  title={{iCoseg: Interactive Co-segmentation with Intelligent Scribble Guidance}},
  author={Batra, Dhruv and Kowdle, Adarsh and Parikh, Devi and Luo, Jiebo and Chen, Tsuhan},
  booktitle=CVPR,
  pages={3169--3176},
  year={2010},
  organization={IEEE}
}

@inproceedings{jerripothula2014automatic,
  title={{Automatic Image Co-Segmentation Using Geometric Mean Saliency}},
  author={Jerripothula, Koteswar Rao and Cai, Jianfei and Meng, Fanman and Yuan, Junsong},
  booktitle=ICIP,
  pages={3277--3281},
  year={2014},
  organization={IEEE}
}

@article{everingham2010pascal,
  title={{The PASCAL Visual Object Classes (VOC) Challenge}},
  author={Everingham, Mark and Van Gool, Luc and Williams, Christopher KI and Winn, John and Zisserman, Andrew},
  journal=IJCV,
  volume={88},
  number={2},
  pages={303--338},
  year={2010},
  publisher={Springer}
}

@inproceedings{lin2014microsoft,
  title={{Microsoft COCO: Common Objects in Context}},
  author={Lin, Tsung-Yi and Maire, Michael and Belongie, Serge and Hays, James and Perona, Pietro and Ramanan, Deva and Doll{\'a}r, Piotr and Zitnick, C Lawrence},
  booktitle=ECCV,
  pages={740--755},
  year={2014},
  organization={Springer}
}

@article{wu2026micon,
  title={{MICON-Bench: Benchmarking and Enhancing Multi-Image Context Image Generation in Unified Multimodal Models}},
  author={Wu, Mingrui and Liu, Hang and Ji, Jiayi and Sun, Xiaoshuai and Ji, Rongrong},
  journal={arXiv preprint arXiv:2602.19497},
  year={2026}
}

@inproceedings{qiao2025multiple,
  title={{Multiple Images Distract Large Multimodal Models via Attention Fragmentation}},
  author={Qiao, Tingrui and Zhao, Di and Li, Yuzhuo and Pang, Bo and Walker, Caroline and Cunningham, Chris and Koh, Yun Sing},
  booktitle=ECCV,
  year={2026}
}

@article{fateh2024video,
  title={{Video LLMs for Temporal Reasoning in Long Videos}},
  author={Fateh, Fawad Javed and Ahmed, Umer and Khan, Hamza and Zia, M Zeeshan and Tran, Quoc-Huy},
  journal={arXiv preprint arXiv:2412.02930},
  year={2024}
}

@inproceedings{li2024llama,
  title={{LLaMA-VID: An Image is Worth 2 Tokens in Large Language Models}},
  author={Li, Yanwei and Wang, Chengyao and Jia, Jiaya},
  booktitle=ECCV,
  pages={323--340},
  year={2024}
}

@inproceedings{fu2024blink,
  title={{BLINK: Multimodal Large Language Models Can See but Not Perceive}},
  author={Fu, Xingyu and Hu, Yushi and Li, Bangzheng and Feng, Yu and Wang, Haoyu and Lin, Xudong and Roth, Dan and Smith, Noah A and Ma, Wei-Chiu and Krishna, Ranjay},
  booktitle=ECCV,
  pages={148--166},
  year={2024}
}

@inproceedings{li2024partglee,
  title={{PartGLEE: A Foundation Model for Recognizing and Parsing Any Objects}},
  author={Li, Junyi and Wu, Junfeng and Zhao, Weizhi and Bai, Song and Bai, Xiang},
  booktitle=ECCV,
  pages={475--494},
  year={2024},
  organization={Springer}
}

@inproceedings{sun2023going,
  title={{Going Denser with Open-Vocabulary Part Segmentation}},
  author={Sun, Peize and Chen, Shoufa and Zhu, Chenchen and Xiao, Fanyi and Luo, Ping and Xie, Saining and Yan, Zhicheng},
  booktitle=ICCV,
  pages={15407--15419},
  year={2023},
  organization={IEEE}
}

@inproceedings{he2022partimagenet,
  title={{PartImageNet: A Large, High-Quality Dataset of Parts}},
  author={He, Ju and Yang, Shuo and Yang, Shaokang and Kortylewski, Adam and Yuan, Xiaoding and Chen, Jie-Neng and Liu, Shuai and Yang, Cheng and Yu, Qihang and Yuille, Alan},
  booktitle=ECCV,
  pages={128--145},
  year={2022},
  organization={Springer}
}

@inproceedings{choi2025fine,
  title={{Fine-Grained Image-Text Correspondence with Cost Aggregation for Open-Vocabulary Part Segmentation}},
  author={Choi, Jiho and Lee, Seonho and Lee, Minhyun and Lee, Seungho and Shim, Hyunjung},
  booktitle=CVPR,
  pages={9782--9793},
  year={2025},
  organization={IEEE}
}

@article{bai2025qwen3,
  title={{Qwen3-VL Technical Report}},
  author={Bai, Shuai and Cai, Yuxuan and Chen, Ruizhe and Chen, Keqin and Chen, Xionghui and Cheng, Zesen and Deng, Lianghao and Ding, Wei and Gao, Chang and Ge, Chunjiang and others},
  journal={arXiv preprint arXiv:2511.21631},
  year={2025}
}

@inproceedings{boerner2023access,
  title={{ACCESS: Advancing Innovation: NSF's Advanced Cyberinfrastructure Coordination Ecosystem: Services \& Support}},
  author={Boerner, Timothy J and Deems, Stephen and Furlani, Thomas R and Knuth, Shelley L and Towns, John},
  booktitle={2023 Practice and Experience in Advanced Research Computing, PEARC 2023},
  pages={173--176},
  year={2023},
  organization={Association for Computing Machinery}
}

@misc{openai2025gpt41,
  author       = {{OpenAI}},
  title        = {{GPT-4.1}},
  year         = {2025},
  howpublished = {\url{https://openai.com/index/gpt-4-1/}}
}

@inproceedings{wang2024all,
  title={{The All-Seeing Project V2: Towards General Relation Comprehension of the Open World}},
  author={Wang, Weiyun and Ren, Yiming and Luo, Haowen and Li, Tiantong and Yan, Chenxiang and Chen, Zhe and Wang, Wenhai and Li, Qingyun and Lu, Lewei and Zhu, Xizhou and others},
  booktitle=ECCV,
  pages={471--490},
  year={2024}
}

@article{deshmukh2025nvidia,
  title={{NVIDIA Nemotron Nano v2 VL}},
  author={Deshmukh, Amala Sanjay and Chumachenko, Kateryna and Rintamaki, Tuomas and Le, Matthieu and Poon, Tyler and Taheri, Danial Mohseni and Karmanov, Ilia and Liu, Guilin and Seppanen, Jarno and Chen, Guo and others},
  journal={arXiv preprint arXiv:2511.03929},
  year={2025}
}
}

\clearpage
\appendix
\clearpage
\setcounter{page}{1}

\section{\benchmarkname{} Dataset Construction}
\label{supp:dataset}
\subsection{Seed Datasets}
\begin{itemize}
    \item {\textbf{\textsc{ADE20K}}~\cite{zhou2017scene}: \textsc{ADE20K} is a large-scale scene parsing benchmark of diverse indoor and outdoor images with dense pixel-level annotations, covering $\sim$150 object/stuff categories and is commonly employed to train and evaluate fine-grained semantic segmentation models. For instance-level supervision, we use the instance segmentation subset from the MIT Scene Parsing dataset.\footnote{\url{http://sceneparsing.csail.mit.edu/}.}}\looseness-1
    \item {\textbf{\textsc{ADE20K-Part-234}}~\cite{wei2024ov}: \textsc{ADE20K-Part-234} is a cleaned \textsc{ADE20K} subset providing part-level masks for 44 object classes over 234 part categories. Data is derived from SceneParse150 by keeping objects with multiple frequently annotated parts, merging duplicated labels, and filtering noisy annotations.}
    \item {\textbf{PASCAL VOC 2010}~\cite{everingham2010pascal}: PASCAL VOC 2010 is a standard benchmark for object recognition and segmentation, containing natural images annotated for classification, detection, and segmentation across 20 categories. We use VOC 2010 for object-level annotations.}
    \item {\textbf{PASCAL-Part}~\cite{chen2014detect}: PASCAL-Part extends PASCAL VOC 2010 with pixel-wise masks for object parts (\eg head, torso, legs) across the same 20 categories, enabling semantic part segmentation and finer-grained evaluation. We adopt PASCAL-Part for part-level annotations, following a procedure similar to \cite{wei2024ov} to merge duplicate parts and reduce label noise.}
    \item {\textbf{\textsc{COCO}}~\cite{lin2014microsoft}: \textsc{COCO} (Common Objects in Context) is a large-scale dataset of everyday scenes with dense annotations for object detection, instance segmentation, keypoints, and captioning across 80 object categories. We use the \textsc{COCO} 2017 split for object-level annotations.}
    \item {\textbf{\textsc{PACO-LVIS}}~\cite{ramanathan2023paco}: \textsc{PACO-LVIS} is the LVIS-image split of \textsc{PACO} built on \textsc{COCO} images, augmenting \textsc{LVIS} with fine-grained part segmentation and object/part attributes (75 object classes and 456 object--part classes). We use \textsc{PACO-LVIS} for part-level annotations.}
\end{itemize}

\subsection{Dataset Sampling}
\noindent \textbf{Feature Extraction.}
To find similar images, we first extract  DINOv2~\cite{oquab2023dinov2} features $\text{DINO}(\mathbf{I}_{i})$ corresponding to each image sample $\mathbf{I}_{i}$. We then define the distance between two image samples $\mathbf{I}_i$ and $\mathbf{I}_j$ using cosine distance
\begin{equation}
    d(\mathbf{I}_i, \mathbf{I}_j) = 1 - \frac{\text{DINO}(\mathbf{I}_{i}) \cdot \text{DINO}(\mathbf{I}_{j})}{\Vert \text{DINO}(\mathbf{I}_{i}) \Vert \cdot \Vert \text{DINO}(\mathbf{I}_{j}) \Vert } 
\end{equation}
To ensure the diversity and relevance of the image sets, for each image, we sample a set containing 2 or 3 related images according to the following two criteria:

\noindent \textbf{Nearest Neighbor Sampling}\label{ss:nns}:
For each image in the dataset, we sample $n\!\in\!\{1, 2\}$ images from its $k$-nearest neighbors, where $k$ is empirically set to $20$. This ensures that the \benchmarkname{} dataset contains image sets that compare objects and parts in similar scenes.

\noindent \textbf{Object-Category-based Sampling}\label{ss:oc}:
We also manually validate all possible object pairs across the three datasets by evaluating whether comparisons between objects from given categories are meaningful. For each object, we then sample all images containing at least one occurrence of that object. Next, we filter images containing objects from relevant categories and randomly sample $n\!\in\!\{2, 3\}$ images from the remaining $k_2$-nearest neighbors, where $k_2$ is empirically set to $5$. This criterion ensures that the \benchmarkname{} dataset consists of image sets that compare objects and parts that are closely related.\looseness-1
\subsection{Question-Answer Pair Generation}
We synthesize the numerical QA pairs using rule-based templates that utilize the instance segmentation annotation available in the source datasets. For the functional, spatial, and open-ended QA pairs, we utilize the GPT-4o API to generate question-answer pairs for the sampled image sets. We employ a temperature of $1.0$ to maintain a balance between correctness and diversity. We carefully design a prompt to generate 3-5 question-answer pairs per image set. We modify the prompt for each specific type of reasoning questions, \ie functional, spatial, and open-ended. An example user prompt for open-ended reasoning QA generation is provided in \Cref{fig:prompt_full}. The prompt is composed of several carefully crafted components to guide the generation using precise image description and annotation, question and answer guidelines, and format requirements. We provide each of these components in Figures \ref{fig:prompt_image_description}--\ref{fig:prompt_format_requirements}.

To ensure precise identification and referencing, in each prompt, we provide the language model with a complete list of all the objects and parts, along with their associated bounding-box information. To distinguish multiple instances of the same object or part, we append unique identifiers to their names in the format
$\text{NAME}\_\text{XYY}$ for objects and $\text{NAME}\_\text{XYYZZ}$ for parts. Here, the first digit $\text{X}$ denotes the image the annotation belongs to. The next two digits $\text{YY}$ provide a unique identifier for the object within that image. Similarly, the last two digits $\text{ZZ}$ denote a unique identifier for the part in that image. These identifiers are later used to extract the corresponding grounding information during data preprocessing. 

\begin{figure*}[t!]
    \centering
    \resizebox{.9\textwidth}{!}{
    \begin{subfigure}[b]{0.33\textwidth}
        \centering
        \includegraphics[width=\textwidth]{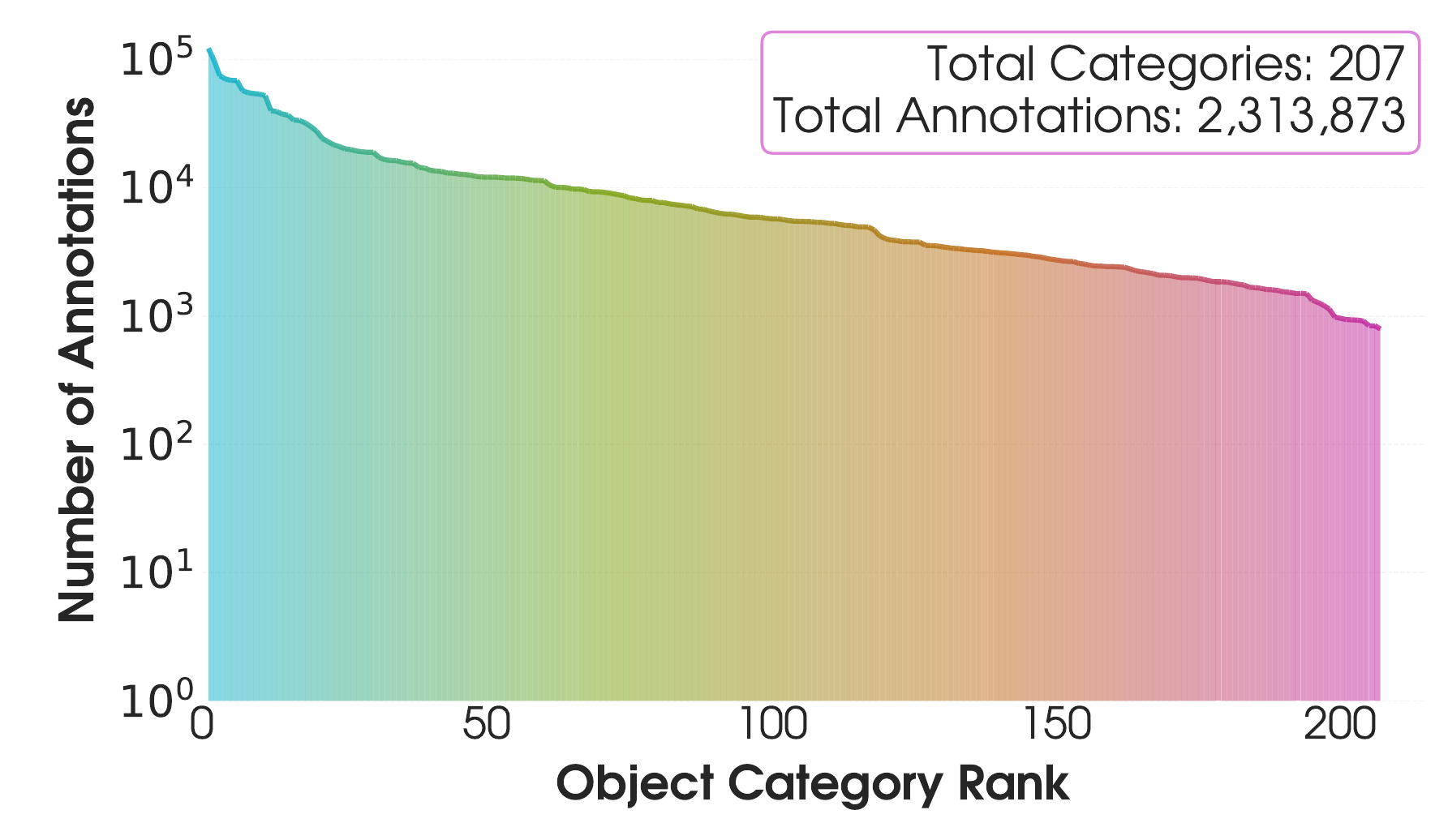}
        \caption{Objects Distribution}
        \label{fig:objects_distribution}
    \end{subfigure}
    \hfill
    \begin{subfigure}[b]{0.33\textwidth}
        \centering
        \includegraphics[width=\textwidth]{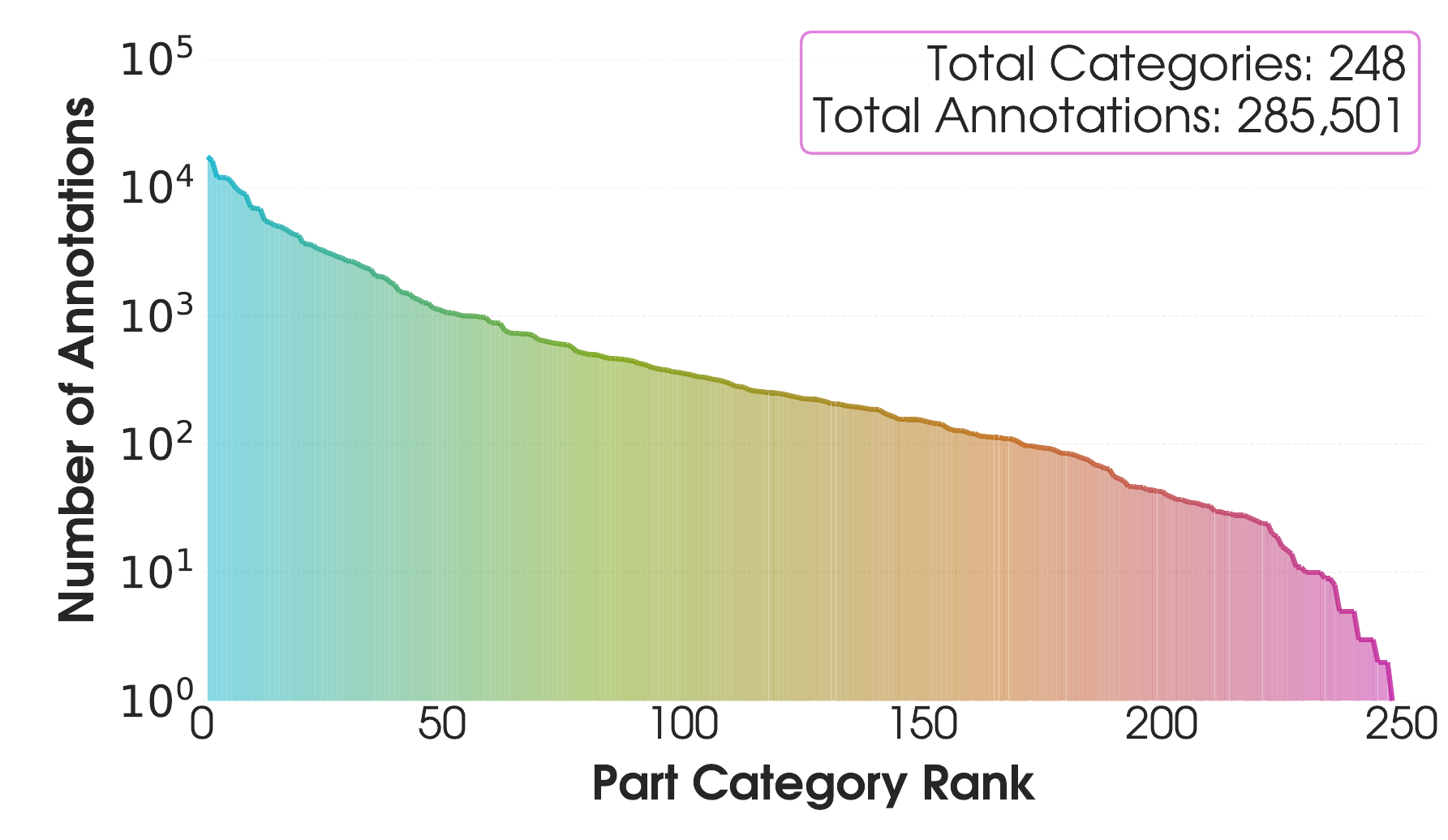}
        \caption{Parts Distribution}
        \label{fig:parts_distribution}
    \end{subfigure}
    \hfill
    \begin{subfigure}[b]{0.33\textwidth}
        \centering
        \includegraphics[width=\textwidth]{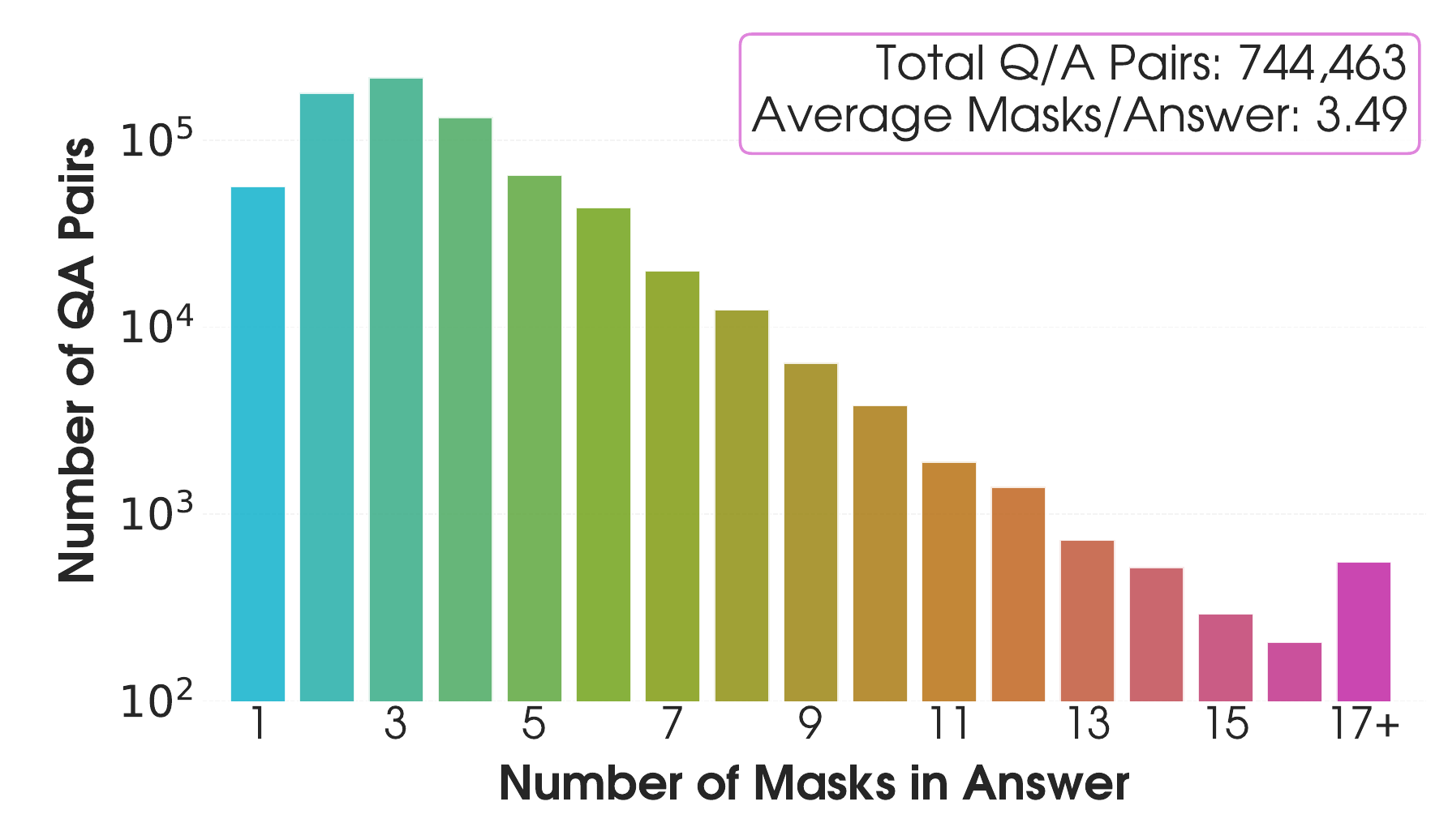}
        \caption{Targets Distribution}
        \label{fig:targets_distribution}
    \end{subfigure}
    }
    \vspace{-0.2cm}
    \caption{\textbf{\benchmarkname{} distribution plots for (a)  objects, (b) parts, and (c)  segmentation targets}. Here, (a) and (b) illustrate the frequency of the i-th object and i-th part, respectively, sorted by frequency, while (c) shows the distribution of answers with $m$ masks.}
    \label{fig:distribution_plots}
    \vspace{-0.1cm}
\end{figure*}

\subsection{Dataset Filtering}
Beyond careful curation of image sets and prompts, to further ensure the quality of the \benchmarkname{} dataset, we implement a robust filtering strategy to eliminate faulty question-answer pairs. Specifically: (a) we discard questions that merely list objects or parts in the images without grounding; (b) we filter out questions referencing locations as bounding boxes; (c) we filter pairs where answers do not refer to all provided images via either text or masks; (d) we retain only examples where answers include at least one segmentation mask from any image; and (e) we exclude pairs mentioning objects or parts not present in the images. This process discards 134,941 question-answer pairs ($\sim$15.4\% of the dataset).
The final \benchmarkname{} data consists of $\sim$743K and 1000 training and test samples, respectively. \Cref{fig:distribution_plots} presents detailed statistics and distributions of objects, parts, and targets in \benchmarkname{}. Across both splits, QA pairs contain an average of 3.5 target masks (objects or parts); the test set is additionally capped at 16 targets in a QA pair. These targets are identified at both object and part levels, making \benchmarkname{} a valuable resource for advancing models in fine-grained multi-image visual understanding. Figures \ref{tab:suppl:dataset_examples1}--\ref{tab:suppl:dataset_examples4} show \benchmarkname{} examples of all categories.

\subsection{Human Validation}
We manually validate a stratified sample of 100 test examples
(25 per question type), covering all six source datasets and both object-
and part-level grounding. In this sample, 99\% of grounded entity labels
agree with the reference answer. All audited masks localize the intended
entity, while 93\% accurately delineate its spatial extent under a stricter
mask-quality criterion. The remaining 7\% exhibit inherited boundary noise,
primarily moderate over-segmentation into surrounding regions or
under-segmentation, rather than incorrect target identity.
Crucially, 94\% of examples require evidence from at least two images,
including 100\% of counting and spatial examples and 96\% of open-ended
examples. Under the stricter criterion that the answer cannot be obtained
by considering the images separately, 83\% satisfy both requirements.
These results support the semantic integrity of the grounding annotations, confirming that the benchmark evaluates
genuinely multi-image questions.

\section{Experimental Setup}\label{supp:metrics}
\subsection{Baselines}
\textbf{General-purpose baselines}. We adapt several open-source and closed-source general-purpose VLMs, \eg GPT-4o~\cite{hurst2024gpt}, GPT-4.1~\cite{openai2025gpt41}, Gemini-2.5 Pro~\cite{comanici2025gemini}, InternVL3-78B~\cite{zhu2025internvl3}, Qwen3-VL~\cite{bai2025qwen3} (8B and 235B), Nemotron-Nano-12B-v2-VL~\cite{deshmukh2025nvidia} and Pixtral Large~\cite{mistral2024pixtralLarge}. Since none of these natively support pixel grounding, we follow previous works~\cite{ren2024grounded} to adapt these models for the pixel-grounded reasoning task. More concretely, we prompt these models to generate reasoning along with structured grounding information that we later utilize to generate pixel grounding using SAM~\cite{kirillov2023segment}. We experiment with three different types of grounding information, \eg bounding box (xyxy), bounding box (xywh), and coordinates. We also experiment with both normalized and non-normalized coordinate systems and observe that all models perform better with normalized coordinate systems. Initial experiments demonstrate that smaller models perform well with the bounding box (xyxy) system, whereas larger ($>$12B) models perform better with the normalized coordinate system. Additionally, we observe that smaller models struggle with this task, so we report one 8B and one 12B model and focus the rest of the baselines on larger models. For a fair evaluation, we report the best score across the different settings for each model.

\noindent \textbf{Pixel grounding baselines}. We adapt two SOTA single-image pixel-grounding LVLMs, LISA~\cite{lai2024lisa} and GLaMM~\cite{rasheed2024glamm} for our new task. Following previous work~\cite{nguyen2025calico,xu2025mc}, we modify these models to be able to reason over multiple images. Specifically, we generate the image encodings for each of the images and project them to the language model space sequentially. We observe that the baselines sometimes fail to generate responses with pixel grounding. For a fair evaluation, we do not penalize the failure cases for the baselines and only calculate the metrics on the valid samples for each baseline.  \modelname{} is trained for 1 epoch on \benchmarkname{} with an effective batch size of 160 using four A100-40GB GPUs, and the training schedules of GLaMM and LISA are adjusted accordingly.\looseness-1

\begin{table*}[t!]
\centering
\caption{\textbf{iCoseg Per-Category Results.} We report per-category and average Jaccard indexes ($\mathcal{J}$). \modelname{} achieves the best performance on five categories and obtains the highest average Jaccard index among the compared methods.}
\vspace{-0.1cm}
\resizebox{\textwidth}{!}{
\begin{tabular}{lccccccccc}
\toprule
\textbf{Model} & bear2 & brownbear & cheetah & elephant & helicopter & hotballoon & panda1 & panda2 & Average \\ 
\midrule
\citet{rubinstein2013unsupervised}
& 65.3 & 73.6 & 69.7 & 68.8 & 80.3 & 65.7 & 75.9 & 62.5 & 70.2 \\
\citet{jerripothula2014automatic}
& 70.1 & 66.2 & 75.4 & 73.5 & 76.6 & 76.3 & 80.6 & 71.8 & 73.8 \\
\citet{jerripothula2016image}
& 67.5 & 72.5 & 78.0 & 79.9 & 80.3 & 80.2 & 72.2 & 61.4 & 74.0 \\
\citet{faktor2013co}
& 72.0 & 92.0 & 67.0 & 67.0 & 82.0 & 88.0 & 70.0 & 55.0 & 78.2 \\
\citet{li2019deep}
& 88.7 & 91.5 & 71.5 & 85.1 & 73.1 & 91.1 & 87.5 & 84.7 & 84.2 \\
\citet{zhang2020deep}
& 91.1 & 89.6 & 88.6 & 90.9 & 76.4 & 94.2 & 90.4 & 87.5 & 89.2 \\
CycleSegNet \cite{zhang2021cyclesegnet}
& \textbf{92.8} & \textbf{94.1} & 89.1 & 91.6 & \textbf{85.2} & 95.1 & 94.8 & 94.1 & 92.1 \\
\midrule
\cc \textbf{\modelname{}}
& \cc \underline{91.5} & \cc \underline{93.0} & \cc \textbf{93.3} & \cc \textbf{93.7} & \cc 81.7 & \cc \textbf{96.6} & \cc \textbf{96.2} & \cc \textbf{95.9} & \cc \textbf{92.7} \\
\bottomrule
\end{tabular}
}
\label{tab:suppl:icoseg_groups}
\end{table*}

\subsection{Evaluation Metrics}

\textbf{Mean Intersection over Union (mIoU)}~\cite{kirillov2023segment}: mIoU is a standard segmentation metric that quantifies how accurately the model predicts object boundaries by averaging the overlap between predicted segmentation masks and ground truth masks across all classes. \\
\textbf{Recall}~\cite{rasheed2024glamm}: Recall measures the model's ability to retrieve all relevant instances by computing the ratio of true positive predictions to all actual positive instances. Following recent work~\cite{rasheed2024glamm}, we obtain true positives by (1) identifying predictions with IoU values above a 0.5 threshold and then (2) among these, selecting the grounded phrase with the highest IoU that also achieves a SentenceBERT~\cite{reimers2019sentence} semantic similarity score above 0.5, thereby qualifying it as a text--mask match, \ie a true positive. \\
\textbf{SS}, \textbf{S-IoU}~\cite{conti2023vocabulary}: 
Following prior work~\cite{yuan2024osprey, conti2023vocabulary}, we compute Semantic Similarity (SS) as cosine between predicted and ground truth phrases in the SentenceBERT~\cite{reimers2019sentence} embedding space, and Semantic IoU (S-IoU) as their word overlap. \\
\textbf{I-SS}, \textbf{I-SIoU}: Because SS and S-IoU only take into account text labels, we additionally use IoU-matched SS (I-SS) and IoU-matched S-IoU (I-SIoU) as stricter metrics to jointly evaluate segmentation and text generation. For these metrics, as in Recall, we only consider labels whose corresponding masks exceed a 0.5 IoU threshold with the ground truth, and then compute the average SS and S-IoU over this subset, yielding our I-SS and I-SIoU scores. \\
\textbf{METEOR}: We also report the standard METEOR score, computed using the official \texttt{cocoevalcap} library. METEOR evaluates the quality of the generated text based on token-level matches, incorporating stemming and synonymy, and is computed over the entire output sentence. Whereas SS and S-IoU focus on the grounded phrases associated with specific segments, METEOR assesses the generated sentence as a whole, providing an additional metric for holistically evaluating our models. 

\begin{table*}[t]
\caption{\textbf{MIG-Bench Performance with 7B-Scale LVLMs.}}
\centering
\vspace{-0.2cm}
\resizebox{\textwidth}{!}{
    \begin{tabular}{lccc  cccccccc}
    \toprule
    \multirow{4}{*}{\textbf{Models}} & \multicolumn{3}{c}{\textbf{Spontaneous Grounding }} & \multicolumn{7}{c}{\textbf{ Referential Grounding }} & \multirow{4}{*}{\textbf{Avg.}} \\
    \cmidrule(lr){2-4} \cmidrule(lr){5-11}

    \multicolumn{1}{c}{} & \multicolumn{2}{c}{\textbf{Difference}} & \textbf{Similarity} & \multicolumn{4}{c}{\textbf{Visual Reference}} & \textbf{Textual} & \multicolumn{2}{c}{\textbf{Visual+Textual}} & \\
    \cmidrule(lr){2-3} \cmidrule(lr){4-4} \cmidrule(lr){5-8} \cmidrule(lr){9-9} \cmidrule(lr){10-11}

    \multicolumn{1}{c}{} & \textbf{Static} & \textbf{Robust} & \multicolumn{1}{c}{\textbf{Common}} & \textbf{OT} & \textbf{MV} & \textbf{Region} & \textbf{Refer} & \multicolumn{1}{c}{\textbf{GG}} & \textbf{Reason} & \multicolumn{1}{c}{\textbf{Co-Re}} & \\

    \midrule
    Mantis & 1.52 & 0.00 & 3.31 & 12.18 & 2.08 & 1.00 & 1.01 & 10.02 & 0.00 & 0.85 & 3.20 \\
    LLaVA-OV-7B & 6.06 & 3.19 & 3.43 & 0.18 & 1.04 & 1.08 & 9.09 & 15.43 & 6.93 & 0.85 & 4.73 \\
    Minicpm2.6 & 14.58 & 2.13 & 14.34 & 9.82 & 6.25 & 1.75 & 11.11 & 10.02 & 2.97 & 2.56 & 7.55 \\
    mPLUG-Owl3 & 18.56 & 6.38 & 34.93 & 8.55 & 7.64 & 2.41 & 7.07 & 22.85 & 9.09 & 5.98 & 12.35 \\
    InternVL2-8B & 6.92 & 7.45 & 25.49 & 20.73 & 9.72 & 3.49 & 28.28 & 30.26 & 17.82 & 9.40 & 15.96 \\
    Qwen2-VL-7B & 27.84 & 38.30 & 19.36 & 20.73 & 11.81 & 25.95 & 23.23 & 58.52 & 48.51 & 11.97 & 28.62 \\
    mPLUG-Owl3$_{+CoT}$ & 16.29 & 8.51 & 55.39 & 44.36 & 25.35 & 19.04 & 36.36 & 30.86 & 18.81 & 10.26 & 26.52 \\
    InternVL2-8B$_{+CoT}$ & 14.58 & 7.45 & 72.54 & 40.91 & 27.78 & 28.60 & 67.68 & 44.49 & 41.58 & 11.97 & 35.76 \\
    Qwen2-VL-7B$_{+CoT}$ & 23.48 & 40.43 & 63.85 & 62.73 & 42.71 & 24.85 & 54.55 & 43.29 & 51.49 & 30.77 & 43.82 \\
    Migician & 65.15 & 46.81 & 84.19 & 70.73 & 60.07 & 74.31 & 76.77 & 66.53 & 59.41 & 34.19 & 63.82 \\
\midrule 
\cc \textbf{\modelname{}}
& \cc 12.69 & \cc 24.47 & \cc 64.95 & \cc 54.36 & \cc 10.42
& \cc 10.22 & \cc 41.41 & \cc 42.89 & \cc 18.81 & \cc 12.82
& \cc 29.30 \\

    \bottomrule
    \end{tabular}
}

\label{tab:suppl_migbench}
\end{table*}

\begin{table}[t]
  \centering
  \caption{\textbf{MMR zero-shot results.}}
  \vspace{-0.2cm}
  \resizebox{0.98\linewidth}{!}{
  \begin{tabular}{lcccccccc}
    \toprule
    \multirow{3}{*}{\textbf{Method}}
      & \multicolumn{2}{c}{\textbf{Validation}}
      & \multicolumn{6}{c}{\textbf{Test}} \\
    \cmidrule(lr){2-3}\cmidrule(lr){4-9}
      & \multicolumn{2}{c}{Obj. \& Part}
      & \multicolumn{2}{c}{Obj.}
      & \multicolumn{2}{c}{Part}
      & \multicolumn{2}{c}{Obj. \& Part} \\
    \cmidrule(lr){2-3}\cmidrule(lr){4-5}\cmidrule(lr){6-7}\cmidrule(lr){8-9}
      & gIoU & cIoU & gIoU & cIoU & gIoU & cIoU & gIoU & cIoU \\
    \midrule
    LISA-7B                          &13.8 &18.3 &23.5 &25.1 &6.6 &7.9  &14.5 &17.9 \\
    GSVA-7B                          &14.6 &25.1 &26.4 &34.3 &6.0 &11.6 &15.5 &24.8 \\
    GLaMM                            &12.6 &19.2 &23.7 &31.9 &3.9 &6.4  &13.3 &18.7 \\
    LISA-Llama2-13B                    &15.4 &20.0 &26.1 &27.9 &7.4 &8.4 &16.1 &19.8 \\
    \midrule
    \cc \textbf{\modelname{}}                   &\cc \textbf{20.3} &\cc \textbf{42.1} &\cc \textbf{32.9} &\cc \textbf{49.0} &\cc \textbf{7.5} &\cc \textbf{21.0} &\cc \textbf{22.1} &\cc \textbf{38.9} \\
    \bottomrule
  \end{tabular}}
  \label{tab:suppl_mmr}
\end{table}

\subsection{Implementation Details}
\label{appendix:implementation}
We implement and optimize \modelname{} in PyTorch, leveraging DeepSpeed for efficient data-parallel training. Our LVLM backbone is initialized from the GLaMM checkpoint~\cite{rasheed2024glamm}\footnote{\url{https://huggingface.co/MBZUAI/GLaMM-FullScope}.}; the \modulename{} vision encoder is initialized from EVA-CLIP \cite{sun2023eva}; and the global attention module and text embedder are initialized from BLIP-2’s Q-Former \cite{li2023blip}.
Because \modulename{} replaces GLaMM’s original CLIP-based visual interface with components initialized from EVA-CLIP and Q-Former, its output tokens are not initially aligned with the embedding space of GLaMM’s pretrained LLM. Following VPG-C~\cite{li2024fine}, we therefore perform an alignment stage before training on \benchmarkname{}. Specifically, we optimize only \modulename{}’s projection layer, which maps its visual tokens to the LLM embedding dimension, using the standard autoregressive language-modeling objective on image–text pairs from GranD~\cite{rasheed2024glamm}. This stage uses neither \benchmarkname{} examples nor segmentation-mask supervision. The resulting projection initializes the subsequent joint training stage on \benchmarkname{}. Training is conducted on four NVIDIA A100 GPUs (40GB each) for 1 epoch with an effective batch size of 160. We apply LoRA to the LLM's attention query and value projections with rank 8 and scaling factor 16. Optimization uses AdamW with a peak learning rate of $3\times10^{-4}$, a 100-step linear warmup followed by cosine decay, and $\beta$ coefficients of 0.9 and 0.95. We train with batch size 4 and gradient accumulation over 10 steps.

\section{Other Benchmark Results}\label{sec:other_benchmarks}
\subsection{Object Cosegmentation}
In addition to our main results on the \benchmarkname{} benchmark for multi-image pixel-grounded reasoning, we evaluate \modelname{} on object cosegmentation as another type of multi-image pixel-grounding task.
Object cosegmentation is a classic cross-image segmentation task, first introduced by \cite{rother2006cosegmentation}, where the objective is to extract the shared foreground object from a set of unordered, unlabeled images. Unlike multi-image pixel-grounded reasoning, this setting requires no textual output. 
To adapt \modelname{} to cosegmentation, we input each image set with the prompt: ``Can you segment the common object between these images?'' and collect the predicted masks for every image. Following \cite{rubinstein2013unsupervised}, we evaluate on standard subsets of these datasets. The MSRC subset contains 7 classes (bird, car, cat, cow, dog, plane, sheep), and the iCoseg subset comprises 8 categories (bear2, brown\_bear, cheetah, elephant, helicopter, hot\_balloon, panda1, panda2).
We report mean precision ($\mathcal{P}$) and Jaccard index ($\mathcal{J}$) against ground-truth masks (\Cref{tab:suppl:cosegm} in main paper), consistent with prior work. Additionally, Figures \ref{tab:suppl:cosegm_qualitatives1} and \ref{tab:suppl:cosegm_qualitatives2} show qualitative results on MSRC and iCoseg, respectively. Across sets of five (5) images, \modelname{} consistently predicts clean common-object masks over a wide variety of categories, including unseen classes such as hot balloons and cheetahs. In \Cref{tab:suppl:icoseg_groups}, we further compare per-category results on iCoseg against prior work, highlighting \modelname{}’s strong generalization to both rigid and non-rigid object classes.\looseness-1

\subsection{Multi-Image Bounding-Box Grounding}
To understand whether \modelname{} transfers beyond \benchmarkname{} to a
related multi-image localization setting, we additionally evaluate it on
MIG-Bench \cite{li2025migician}. While \benchmarkname{} requires pixel-level masks, MIG-Bench
evaluates multi-image grounding using bounding boxes: given a collection of
images and a query, a model must identify the relevant image(s) and localize
the queried objects. MIG-Bench spans ten tasks, including
common-object, referential, group, and tracking-based grounding, and is
scored using Acc\@0.5.
We evaluate \modelname{} on all tasks, converting native masks to
tight bounding boxes for evaluation. As shown in \Cref{tab:suppl_migbench}, our model reaches 29.30 average
Acc\@0.5, the best result among general-purpose non-CoT 7B-scale
models: +0.68 over Qwen2-VL-7B, +13.34 over InternVL2-8B, and +16.95 over
mPLUG-Owl3. It also exceeds mPLUG-Owl3$_{+\mathrm{CoT}}$ by 2.78 points;
among 7B-scale systems, only CoT-enhanced InternVL2 and Qwen2-VL and the
MIG-specialized Migician score higher.
The largest gains occur on common-object grounding (64.95) and object
tracking (54.36), respectively +30.02 and +33.63 points over the strongest general-purpose
non-CoT 7B baseline. \modelname{} also performs strongly on referential
(41.41) and group grounding (42.89). This pattern aligns with
\benchmarkname{} supervision: comparative questions encourage
cross-image correspondence and image-specific assignment, while native
mask prediction provides localization that transfers directly to
MIG-Bench boxes.

\subsection{Single-Image Reasoning Segmentation}

MMR~\cite{jang2025mmr} is a large-scale single-image benchmark for
multi-target and multi-granularity reasoning segmentation, containing 194,398 complex, implicit
question--answer pairs spanning object- and part-level targets. A question may
require masks for multiple objects, fine-grained object parts, or both, and the
test set separately reports object-only, part-only, and mixed object/part
settings. This makes MMR a useful retention test for \modelname{}:
although our model is specialized for reasoning across multiple images, its
native $N{=}1$ interface should still localize multiple fine-grained targets in
a single image.
Following the official MMR codebase, we evaluate the complete validation and
test splits without MMR-specific
training. For the
direct transfer comparison, we focus on LISA, GSVA, and GLaMM checkpoints that
also receive no MMR training. MMR reports gIoU, the mean IoU over target
masks, and cIoU, the ratio of cumulative intersection to cumulative union.
Because cIoU is dominated more strongly by large regions, gIoU is particularly
informative for the part-only split.
Across all MMR validation and test examples, \modelname{} obtains 19.58/40.23
gIoU/cIoU with 100\% mask coverage. On the mixed object/part test, its gIoU is
22.1, compared with 14.5 for non-retrained LISA-7B, 16.1 for LISA-13B, and
13.3 for GLaMM. \modelname{} is the strongest checkpoint
without MMR training on every validation and test column in
\Cref{tab:suppl_mmr}. The part-only split remains the most difficult:
gIoU is 7.5, compared with 32.9 on object-only examples.
Therefore, the native single-image interface and multi-target decoding remain
functional after multi-image specialization.

\section{Additional Experiments}\label{app:additional_exps}

\begin{table}[t]
\centering
\caption{\textbf{Image-assignment diagnostics.}
Presence-based metrics measure source-image selection; exact count and
sequence additionally require correct mask multiplicity and ordering.}
\label{tab:suppl_image_assignment_diagnostics}
  \vspace{-0.2cm}
\setlength{\tabcolsep}{4pt}
\small
\begin{tabular}{lrrr}
\toprule
\textbf{Metric (\%)} & \textbf{All} & \textbf{2 img.} & \textbf{3 img.} \\
\midrule
Assignment F1 & 90.79 & 95.63 & 81.17 \\
Exact image set & 73.30 & 84.26 & 50.00 \\
All answer images covered & 81.00 & 90.59 & 60.63 \\
Order-aligned assignment & 75.80 & 81.84 & 62.66 \\
Exact per-image count & 37.90 & 46.32 & 20.00 \\
Exact identifier sequence & 34.70 & 43.09 & 16.88 \\
\bottomrule
\end{tabular}
\end{table}

\subsection{Image Assignment and Mask Cardinality}
Conventional mask metrics do not reveal whether an error comes from
selecting the wrong source image or predicting the wrong number of masks in an
otherwise correct image. We therefore evaluate image assignment and per-image
cardinality separately on all test outputs by extracting
the image identifier attached to every generated \texttt{[SEG](IMAGE$i$)} token.
The reference identifier sequence follows the unique grounded entities in the
annotated answer. For presence-based metrics (\ie identifying whether the response covers an image), repeated identifiers are
collapsed into an image set: precision, recall, and F1 measure whether selected
images contain answer targets; exact image-set accuracy requires the predicted
and reference sets to be identical; and required-image coverage asks whether
every reference image is selected. Exact per-image count is stricter, requiring
every image to receive exactly the correct number of \texttt{[SEG]} tokens, so
one extra or missing prediction makes the entire example incorrect.
Order-aligned assignment compares identifiers at corresponding positions up to
the shorter sequence, whereas exact identifier sequence additionally requires
the same length and order.

Assignment precision/recall/F1 are 93.67/88.09/90.79\%, exact image-set
accuracy is 73.30\%, and all required images are covered in 81.00\%
(\Cref{tab:suppl_image_assignment_diagnostics}). These presence-based results show that
\modelname{} usually identifies the correct source images. The stricter exact
per-image count and exact identifier-sequence accuracies are 37.90\% and
34.70\%, respectively. The gap widens with more images:
assignment F1 is 95.63\% on two-image prompts and 81.17\% on
three-image prompts, while exact per-image count is 46.32\% and
20.00\%, respectively.
To locate this cardinality bottleneck, we stratify the same outputs by question
type (\Cref{tab:assignment_breakdown}). Counting prompts attain 99.22\% assignment
F1 and 81.60\% exact per-image count. Functional, open-ended, and spatial
prompts retain assignment F1 of 90.25\%, 88.49\%, and 84.02\%, respectively,
but their exact counts are 28.00\%, 21.20\%, and 20.80\%. Therefore, the
model generally identifies which images contain the answer, while exact mask
multiplicity remains difficult for open-ended questions.\looseness-1

\begin{table}[t]
\centering
\caption{\textbf{Question-type breakdown of image assignment and cardinality.}}
\label{tab:assignment_breakdown}
  \vspace{-0.2cm}
\setlength{\tabcolsep}{5pt}
\small
\resizebox{0.98\linewidth}{!}{
\begin{tabular}{lrrrr}
\toprule
\textbf{Metric} & \textbf{Counting} & \textbf{Functional} & \textbf{Open-ended} & \textbf{Spatial} \\
\midrule
Assignment F1         & 99.22 & 90.25 & 88.49 & 84.02 \\
Exact image set       & 98.40 & 66.00 & 67.20 & 61.60 \\
All answer images covered & 98.40 & 77.20 & 77.60 & 70.80 \\
Exact per-image count & 81.60 & 28.00 & 21.20 & 20.80 \\
\bottomrule
\end{tabular}
}

\end{table}

\subsection{Cross-Image Reasoning}
\label{sec:suppl_counterfactual}

\begin{figure*}[t]
    \centering
    \newcommand{\counterfactualCI}[3]{%
        \draw[black,line width=0.85pt]
            (axis cs:#1,#2) -- (axis cs:#1,#3);
        \draw[black,line width=0.85pt]
            ([xshift=-3pt]axis cs:#1,#2) --
            ([xshift=3pt]axis cs:#1,#2);
        \draw[black,line width=0.85pt]
            ([xshift=-3pt]axis cs:#1,#3) --
            ([xshift=3pt]axis cs:#1,#3);
    }
    \newcommand{\counterfactualValue}[4]{%
        \node[
            font=\scriptsize,
            text=#4,
            anchor=west,
            inner sep=0pt,
            xshift=7pt
        ] at (axis cs:#1,#2) {#3};
    }
    \pgfplotsset{
        counterfactualaxis/.style={
            width=\linewidth,
            height=5.0cm,
            grid=major,
            grid style={gray!18},
            axis x line*=bottom,
            axis y line*=left,
            tick label style={font=\scriptsize},
            x tick label style={align=center},
            ylabel style={font=\small},
            enlarge x limits=0.24,
            scaled y ticks=false,
            clip=false,
        }
    }
    \begin{subfigure}[t]{0.48\textwidth}
        \centering
        \begin{tikzpicture}
        \begin{axis}[
            counterfactualaxis,
            ymin=0,
            ymax=38,
            ytick={0,10,20,30},
            ylabel={Difference (percentage points)},
            symbolic x coords={assignment,text},
            xtick=data,
            xticklabels={
                Image-assignment\\change,
                Response-text\\change
            },
        ]

        \counterfactualCI{assignment}{17.52}{32.48}
        \counterfactualCI{text}{8.82}{16.51}

        \addplot[
            only marks,
            mark=*,
            mark size=2.7pt,
            color=DarkSageGreen,
            mark options={
                fill=DarkSageGreen,
                draw=DarkSageGreen
            }
        ] coordinates {
            (assignment,25.18)
            (text,12.65)
        };

        \counterfactualValue{assignment}{25.18}{25.18}{DarkSageGreen}
        \counterfactualValue{text}{12.65}{12.65}{DarkSageGreen}

        \end{axis}
        \end{tikzpicture}
        \caption{Image replacement.}
        \label{fig:suppl_counterfactual_replacement}
    \end{subfigure}
    \hfill
    \begin{subfigure}[t]{0.48\textwidth}
        \centering
        \begin{tikzpicture}
        \begin{axis}[
            counterfactualaxis,
            ymin=0,
            ymax=45,
            ytick={0,10,20,30,40},
            ylabel={Difference (percentage points)},
            symbolic x coords={assignment,count},
            xtick=data,
            xticklabels={
                Image-assignment\\change,
                Per-image mask-count\\change
            },
        ]

        \counterfactualCI{assignment}{22.26}{37.23}
        \counterfactualCI{count}{20.07}{35.40}

        \addplot[
            only marks,
            mark=*,
            mark size=2.7pt,
            color=DarkSageGreen,
            mark options={
                fill=DarkSageGreen,
                draw=DarkSageGreen
            }
        ] coordinates {
            (assignment,29.56)
            (count,27.74)
        };

        \counterfactualValue{assignment}{29.56}{29.56}{DarkSageGreen}
        \counterfactualValue{count}{27.74}{27.74}{DarkSageGreen}

        \end{axis}
        \end{tikzpicture}
        \caption{Image removal.}
        \label{fig:suppl_counterfactual_removal}
    \end{subfigure}

    \vspace{-0.12cm}
    \begin{subfigure}[t]{0.48\textwidth}
        \centering
        \begin{tikzpicture}
        \begin{axis}[
            counterfactualaxis,
            ymin=-5,
            ymax=62,
            ytick={0,10,20,30,40,50,60},
            ylabel={mIoU change (points) / preserved (\%)},
            symbolic x coords={miou,sequence,count},
            xtick=data,
            xticklabels={
                Strict mIoU\\change,
                Assignment sequence\\preserved,
                Mask counts\\preserved
            },
            extra y ticks={0},
            extra y tick style={
                grid=major,
                grid style={black!45,dashed}
            },
        ]

        \counterfactualCI{miou}{-0.22}{1.69}
        \counterfactualCI{sequence}{28.90}{34.50}
        \counterfactualCI{count}{47.97}{54.08}

        \addplot[
            only marks,
            mark=*,
            mark size=2.7pt,
            color=DarkRed,
            mark options={
                fill=DarkRed,
                draw=DarkRed
            }
        ] coordinates {
            (miou,0.72)
            (sequence,31.67)
            (count,50.96)
        };

        \counterfactualValue{miou}{0.72}{0.72}{DarkRed}
        \counterfactualValue{sequence}{31.67}{31.67}{DarkRed}
        \counterfactualValue{count}{50.96}{50.96}{DarkRed}

        \end{axis}
        \end{tikzpicture}
        \caption{Image reordering.}
        \label{fig:suppl_counterfactual_permutation}
    \end{subfigure}
    \hfill
    \begin{subfigure}[t]{0.48\textwidth}
        \centering
        \begin{tikzpicture}
        \begin{axis}[
            counterfactualaxis,
            ymin=0,
            ymax=18,
            ytick={0,5,10,15},
            ylabel={Gain (percentage points)},
            symbolic x coords={miou,recall,set,f1},
            xtick=data,
            xticklabels={
                Strict\\mIoU,
                Recall\\@0.5,
                Image-set\\accuracy,
                Image\\F1
            },
        ]

        \counterfactualCI{miou}{6.30}{8.54}
        \counterfactualCI{recall}{6.43}{9.26}
        \counterfactualCI{set}{7.30}{13.10}
        \counterfactualCI{f1}{3.04}{5.41}

        \addplot[
            only marks,
            mark=*,
            mark size=2.7pt,
            color=DarkSageGreen,
            mark options={
                fill=DarkSageGreen,
                draw=DarkSageGreen
            }
        ] coordinates {
            (miou,7.41)
            (recall,7.85)
            (set,10.20)
            (f1,4.23)
        };

        \counterfactualValue{miou}{7.41}{7.41}{DarkSageGreen}
        \counterfactualValue{recall}{7.85}{7.85}{DarkSageGreen}
        \counterfactualValue{set}{10.20}{10.20}{DarkSageGreen}
        \counterfactualValue{f1}{4.23}{4.23}{DarkSageGreen}

        \end{axis}
        \end{tikzpicture}
        \caption{Cross-image update.}
        \label{fig:suppl_counterfactual_relational}
    \end{subfigure}
  \vspace{-0.2cm}
    \caption{\textbf{Controlled tests of cross-image reasoning.}
    (a) Replacing and (b) removing an answer image produce larger changes than modifying a non-answer image.
    (c) Reordering affects individual predictions but leaves aggregate strict mIoU largely unchanged.
    (d) Enabling \modulename{} improves segmentation and image selection.
    Dots show estimates; whiskers indicate 95\% bootstrap CIs.
    }
    \label{fig:suppl_counterfactual}

\end{figure*}

We further investigate how \modelname{} uses information across images by replacing or removing individual images, changing their order, and disabling the learned cross-image update in \modulename{}. These experiments demonstrate whether the model responds to changes in relevant visual evidence, how sensitive it is to image order, and how much it relies on the cross-image interactions learned during training.

We perform these experiments on the full \benchmarkname{} test set. First, we define an \emph{answer image} as an image containing at least one annotated answer target, and a \emph{non-answer image} as one without annotated targets. This distinction is based on the ground-truth annotations and does not imply that non-answer images are irrelevant to the question, as they may still provide useful context. Image removal is then evaluated by removing one answer image from each prompt. For the paired replacement and removal experiments, we use the 274 examples containing both answer and non-answer images. We uniformly sample one image of each type per prompt with a fixed random seed, using the same selected positions for both experiments. Counting questions are excluded from these paired comparisons because every input image contains an annotated target. All experiments use the same trained model with identical generation settings.
We report 95\% confidence intervals (CIs) using 20,000 bootstrap resamples of the test prompts. Each resample preserves the original sample size, with paired conditions and all reordered versions of a prompt sampled together to account for their dependence.

\noindent\textbf{Replacing and removing images.}
We first examine whether predictions depend more strongly on images containing annotated targets than on those without. We measure changes in image identifiers associated with generated masks, the number of predicted masks per image, and the generated response text. For example, the output \texttt{[SEG]\,(IMAGE2)}, \texttt{[SEG]\,(IMAGE1)}, \texttt{[SEG]\,(IMAGE2)} corresponds to the image-assignment sequence $(2,1,2)$. Mask counts are obtained directly from text predictions, while response-text change is measured as one minus the token sequence similarity between tokenized, lowercased responses.\looseness-1

For image replacement, we separately replace selected answer and non-answer images with the same distractor image. Distractors are sampled from the same source dataset, excluding images present in the input and those whose source annotations overlap with the original target labels. Question and image identifiers remain unchanged, allowing us to compare the effect of replacing different image types while controlling for replacement content.
Replacing an answer image changes the image-assignment sequence in 78.47\% of examples, compared with 53.28\% when replacing a non-answer image. This corresponds to a difference of 25.18 percentage points (\Cref{fig:suppl_counterfactual_replacement}). The same pattern appears in the generated responses, where the average text change is 35.82\% versus 23.17\%, a difference of 12.65 points. Per-image mask counts also change more frequently when answer images are replaced (73.72\% versus 51.46\%), with a difference of 22.26 points. These differences remain positive when we account for prompts sharing input or distractor images, with statistically significant CIs for both assignment changes and response-text changes.

We observe a similar pattern when removing images. Unlike replacement, removal changes the input image count, requiring us to renumber the remaining images and update numeric or ordinal references in the question.  To avoid counting the disappearance of predictions associated with the removed image itself, we discard those predictions from the baseline output and compare only the assignments and mask counts of the surviving images.
Removing an answer image changes the remaining image-assignment sequence in 80.66\% of examples \textit{vs.} 51.09\% when removing a non-answer image, a difference of 29.56 points (\Cref{fig:suppl_counterfactual_removal}). Per-image mask counts change in 77.01\% versus 49.27\% of examples, a 27.74-point difference.

Overall, replacing or removing answer images has a substantially larger effect on the model's predictions than modifying non-answer images, suggesting that \modelname{} responds strongly to relevant images carrying crucial visual evidence.

\noindent\textbf{Shuffling image order.}
We next examine whether \modelname{} produces consistent predictions when the same images are presented in different orders (\Cref{fig:suppl_counterfactual_permutation}). We evaluate every alternative ordering for each test example, updating numeric and ordinal image references in the question accordingly. The 680 two-image prompts each have one alternative ordering, while the 320 three-image prompts each have five, giving 2,280 reordered inputs in total. Predictions are mapped back to the original image positions before evaluation. Within each image, predicted and ground-truth masks are matched to maximize their total IoU, and strict mIoU averages over all ground-truth masks.

Across all reordered inputs, strict mIoU changes from 35.23 to 35.95, an increase of 0.72 points (95\% CI [-0.22,+1.69]). Because three-image prompts contribute five reordered inputs each, these pooled scores give them greater weight, although the confidence intervals resample complete prompts. When each prompt is weighted equally, the difference is smaller at +0.28 points; neither comparison indicates a clear change in aggregate segmentation performance.
However, the individual predictions are less consistent. The exact image-assignment sequence is preserved in only 31.67\% of reordered inputs, while per-image mask counts remain unchanged in 50.96\%. Although assignment-sequence equality also depends on the order in which predictions are generated, mask-count equality does not. Thus, image order affects the model's individual outputs even though its aggregate strict mIoU remains similar.

\noindent\textbf{Disabling cross-image updates.}
Finally, we investigate how much the trained model relies on the cross-image information provided by \modulename{} (\Cref{fig:suppl_counterfactual_relational}). We disable its learned additive update at inference by setting it to zero, while keeping the global-query pathway, visual token count, model weights, and generation settings unchanged. This isolates the effect of the learned update within the trained model, complementing the separately trained Pooling and Projection variants in \Cref{tab:ablation}.

Enabling the cross-image update improves strict mIoU by 7.41 points and Recall by 7.85 points. We also observe improvements in image selection, with exact image-set accuracy increasing by 10.20 points and image-selection F1 by 4.23 points. Both metrics compare the image identifiers associated with generated masks against the images containing annotated targets. All four improvements have CIs above zero, suggesting statistically significant gains.

Overall, replacing or removing answer images affects predictions much more
than changing non-answer images, while consistent image reordering produces no
clear overall decrease in spatial localization. Most importantly, disabling
the model's cross-image update directly reduces target coverage, showing that
\modelname{} benefits from processing the input images jointly via \modulename{}.

\section{Additional Qualitative Results}
\label{suppl:more-qualitatives}
\paragraph{\benchmarkname{} results.}
Figures \ref{tab:suppl:qualitatives1}, \ref{tab:suppl:qualitatives2}, and \ref{tab:suppl:qualitatives3} present additional qualitative results for \modelname{} and our baselines GLaMM~\cite{rasheed2024glamm} and LISA~\cite{lai2024lisa} alongside the Ground Truth, with each question prefixed by its \benchmarkname{} sample type. Following \Cref{tab:config_comparison}, we highlight grounded phrases in the generated text using the same colors as their corresponding masks for clarity. In the rare event that a model produces an ungrounded phrase (\eg phrase with no \texttt{[SEG]} token), we mark it with \colorbox{black}{\color{white} black} highlight. Overall, \modelname{} consistently produces crisp masks that align with the correct grounded phrases, whereas baselines more often miss target objects, generate fragmented or incomplete masks, or hallucinate incorrect groundings. \Cref{tab:suppl:qualitatives3} also includes a failure case: while \modelname{} correctly recognizes that the car’s wheels are partially submerged, it segments only one wheel rather than all four. Even in this case, \modelname{} maintains coherent comparative reasoning that matches the ground-truth logic, whereas the baselines struggle to compare across images, arrive at incorrect answers, and hallucinate visual details.\looseness-1

\noindent \textbf{\modelname{} in the wild.}
To assess robustness, we additionally provide examples of images collected from the web that \modelname{} has not seen during training. In \Cref{tab:suppl:in_the_wild}, we observe that \modelname{} demonstrates strong generalization to these unseen images. Notably, \modelname{} generates consistently superior segmentation masks compared to baselines, as exemplified by the fine-grained details of the hammers in Example 3 and the dining chairs in Example 4. Moreover, we observe that the baselines are more prone to hallucinations, \eg ``pizza is shown with multiple slices'' in the first example (GLaMM), ``watch'' in the second example (both GLaMM and LISA). Beyond visual recognition, these examples underscore \modelname{}'s capability to retain compositional reasoning capabilities when transferring to diverse and noisy real-world images.\looseness-1  

\begin{table*}[t!]
    \centering
    \resizebox{0.99\linewidth}{!}{
    \begin{tabulary}{\textwidth}{
        >{\centering\arraybackslash}p{.04\textwidth}@{}
        >{\centering\arraybackslash}p{.19\textwidth}
        >{\centering\arraybackslash}p{.19\textwidth}
        >{\centering\arraybackslash}p{.19\textwidth}
        >{\centering\arraybackslash}p{.19\textwidth}
        >{\centering\arraybackslash}p{.19\textwidth}
    }
        \toprule

        {\centering\rotatebox{90}{\textbf{sheep}}} &
        \includegraphics[height=1.6cm]{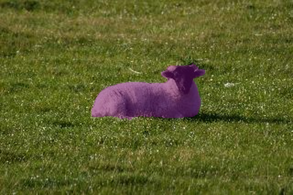} &
        \includegraphics[height=1.6cm]{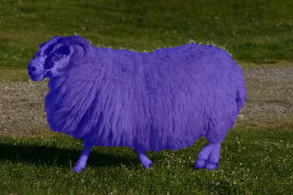} &
        \includegraphics[height=1.6cm]{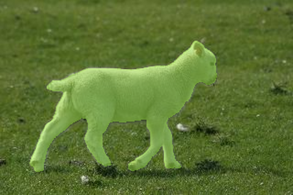} &
        \includegraphics[height=1.6cm]{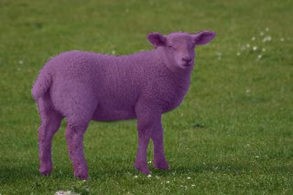} &
        \includegraphics[height=1.6cm]{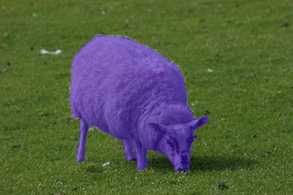} \\
        \midrule

        {\centering\rotatebox{90}{\textbf{bird}}} &
        \includegraphics[height=1.6cm]{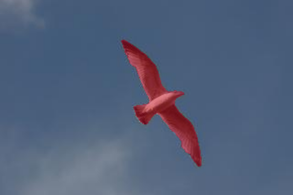} &
        \includegraphics[height=1.6cm]{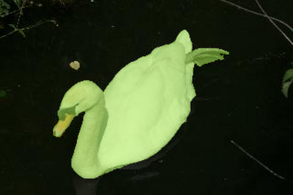} &
        \includegraphics[height=1.6cm]{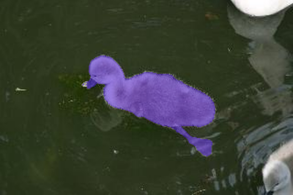} &
        \includegraphics[height=1.6cm]{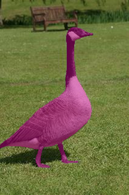} &
        \includegraphics[height=1.6cm]{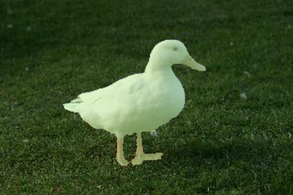} \\
        \midrule

        {\centering\rotatebox{90}{\textbf{car}}} &
        \includegraphics[height=1.6cm]{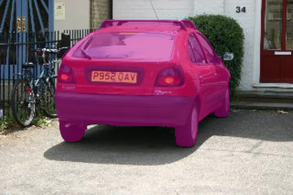} &
        \includegraphics[height=1.6cm]{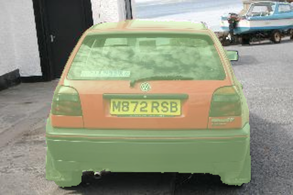} &
        \includegraphics[height=1.6cm]{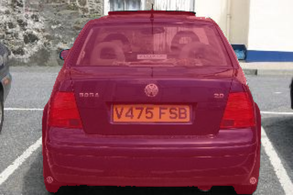} &
        \includegraphics[height=1.6cm]{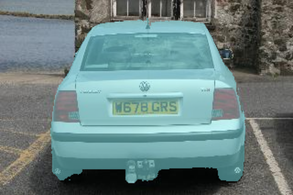} &
        \includegraphics[height=1.6cm]{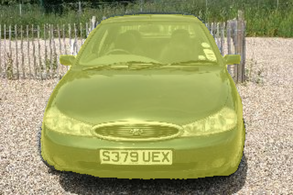} \\
        \midrule

        {\centering\rotatebox{90}{\textbf{cat}}} &
        \includegraphics[height=1.6cm]{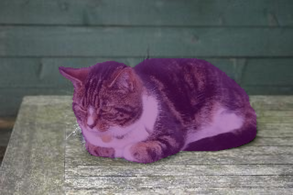} &
        \includegraphics[height=1.6cm]{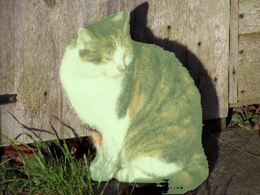} &
        \includegraphics[height=1.6cm]{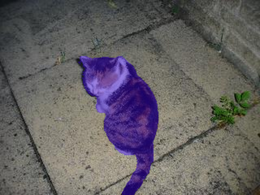} &
        \includegraphics[height=1.6cm]{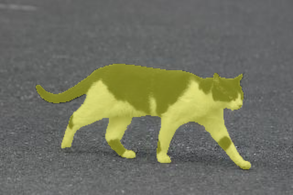} &
        \includegraphics[height=1.6cm]{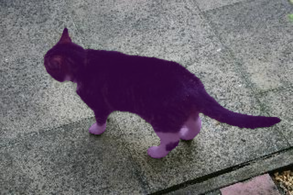} \\
        \midrule

        {\centering\rotatebox{90}{\textbf{cow}}} &
        \includegraphics[height=1.6cm]{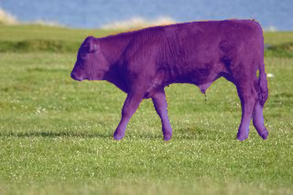} &
        \includegraphics[height=1.6cm]{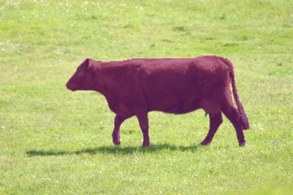} &
        \includegraphics[height=1.6cm]{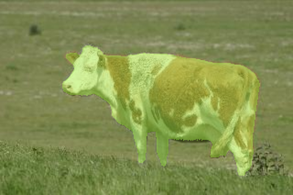} &
        \includegraphics[height=1.6cm]{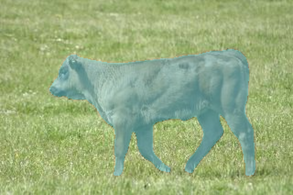} &
        \includegraphics[height=1.6cm]{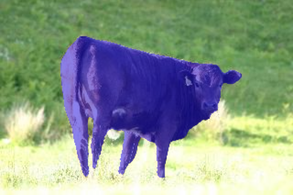} \\
        
        \bottomrule
    \end{tabulary}
    }

    \captionof{figure}{\textbf{\modelname{} object cosegmentation qualitative results on MSRC.}}
    \label{tab:suppl:cosegm_qualitatives1}
\end{table*}

\begin{table*}[t!]
    \centering
    \resizebox{0.99\linewidth}{!}{
    \begin{tabulary}{\textwidth}{
        >{\centering\arraybackslash}p{.04\textwidth}@{}
        >{\centering\arraybackslash}p{.19\textwidth}
        >{\centering\arraybackslash}p{.19\textwidth}
        >{\centering\arraybackslash}p{.19\textwidth}
        >{\centering\arraybackslash}p{.19\textwidth}
        >{\centering\arraybackslash}p{.19\textwidth}
    }
        \toprule

        {\centering\rotatebox{90}{\textbf{HotBalloon}}} &
        \includegraphics[height=1.6cm]{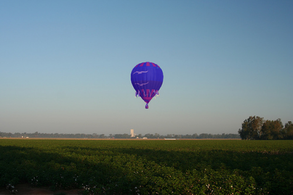} &
        \includegraphics[height=1.6cm]{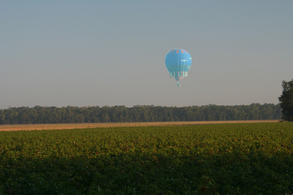} &
        \includegraphics[height=1.6cm]{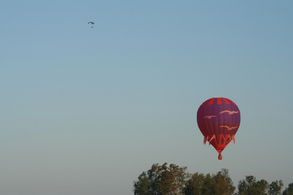} &
        \includegraphics[height=1.6cm]{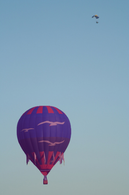} &
        \includegraphics[height=1.6cm]{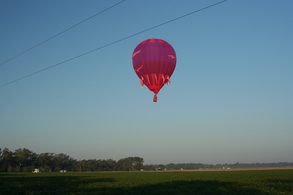} \\
        \midrule

        {\centering\rotatebox{90}{\textbf{brown\_bear}}} &
        \includegraphics[height=1.6cm]{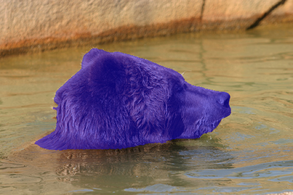} &
        \includegraphics[height=1.6cm]{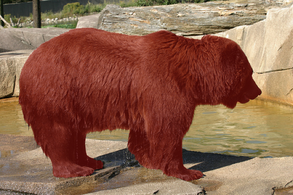} &
        \includegraphics[height=1.6cm]{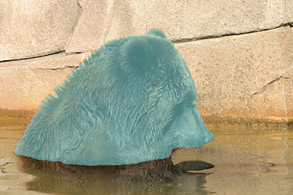} &
        \includegraphics[height=1.6cm]{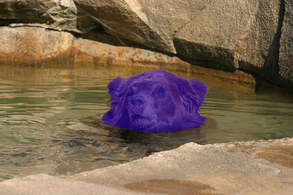} &
        \includegraphics[height=1.6cm]{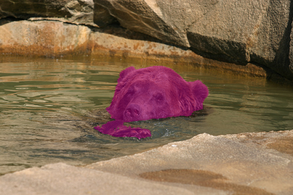} \\
        \midrule

        {\centering\rotatebox{90}{\textbf{cheetah}}} &
        \includegraphics[height=1.6cm]{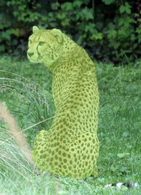} &
        \includegraphics[height=1.6cm]{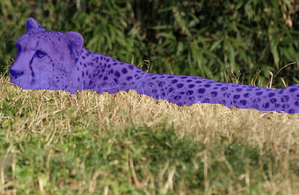} &
        \includegraphics[height=1.6cm]{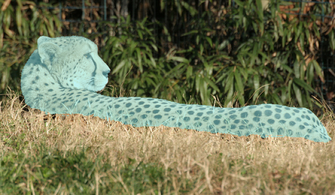} &
        \includegraphics[height=1.6cm]{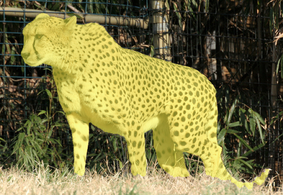} &
        \includegraphics[height=1.6cm]{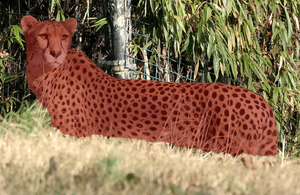} \\
        \midrule

        {\centering\rotatebox{90}{\textbf{panda1}}} &
        \includegraphics[height=1.6cm]{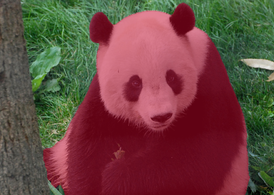} &
        \includegraphics[height=1.6cm]{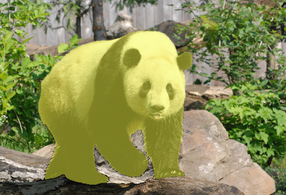} &
        \includegraphics[height=1.6cm]{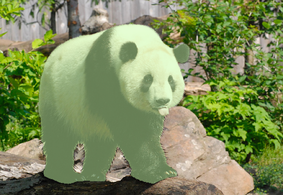} &
        \includegraphics[height=1.6cm]{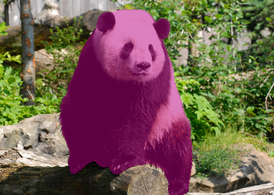} &
        \includegraphics[height=1.6cm]{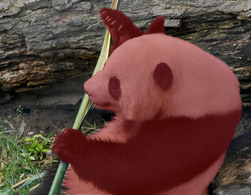} \\
        \midrule

        {\centering\rotatebox{90}{\textbf{elephant}}} &
        \includegraphics[height=1.6cm]{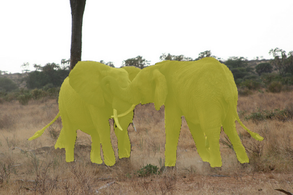} &
        \includegraphics[height=1.6cm]{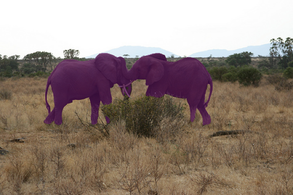} &
        \includegraphics[height=1.6cm]{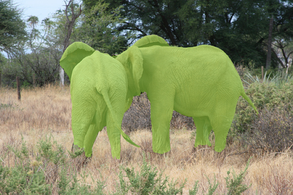} &
        \includegraphics[height=1.6cm]{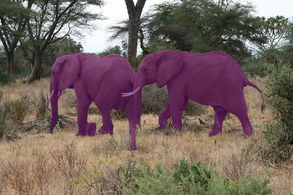} &
        \includegraphics[height=1.6cm]{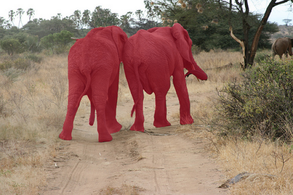} \\
        
        \bottomrule
    \end{tabulary}
    }

    \captionof{figure}{\textbf{\modelname{} object cosegmentation qualitative results on iCoseg.}}
    \label{tab:suppl:cosegm_qualitatives2}
    \vspace{-0.3cm}
\end{table*}

\begin{table*}[t!]
    \centering
    \resizebox{0.99\linewidth}{!}{
        \begin{tabulary}{\textwidth}{>{\centering\arraybackslash}p{.04\textwidth}@{} >{\centering\arraybackslash}p{.3\textwidth} >{\centering\arraybackslash}p{.3\textwidth} >{\centering\arraybackslash}p{.04\textwidth}@{} >{\centering\arraybackslash}p{.3\textwidth} >{\centering\arraybackslash}p{.3\textwidth}}
        \toprule
            \multicolumn{6}{c}{\parbox{1.28\linewidth}{
    \centering\textbf{\colorbox{BackgroundBlue}{\color{TextBlue}{\textit{Open-Ended}}}\quad Q: In a dimly lit environment, which type of lighting fixture might provide more
direct illumination across both spaces?}}}\\
        \midrule
            {\centering\rotatebox{90}{\textbf{Ground Truth}}} & 
            \includegraphics[width=\linewidth]{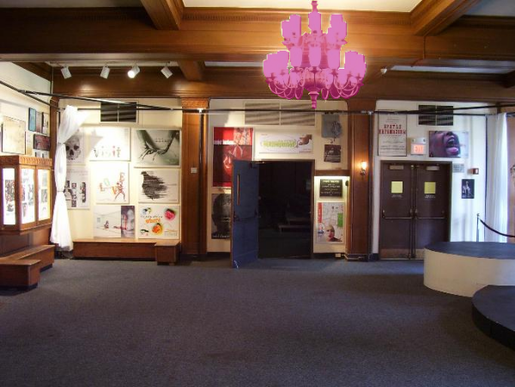} & 
            \includegraphics[width=\linewidth]{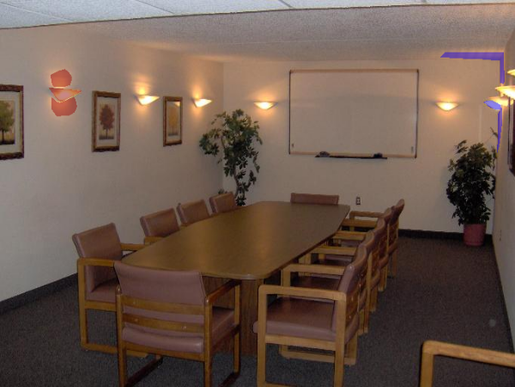} & 
            {\centering\rotatebox{90}{\textbf{\modelname{}}}} & 
            \includegraphics[width=\linewidth]{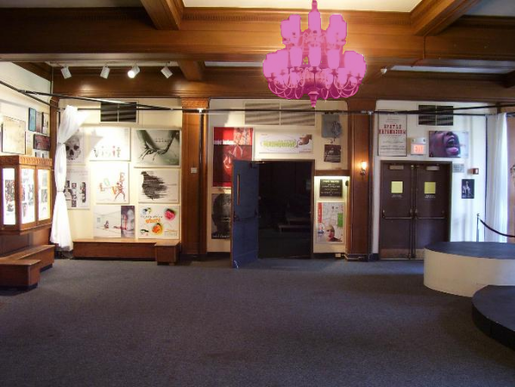} & 
            \includegraphics[width=\linewidth]{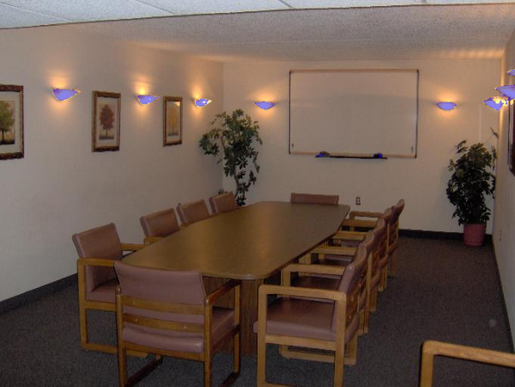} \\
            & \multicolumn{2}{p{.6\textwidth}}{
            In a dimly lit environment, the sconce fixtures in Image 2 (such as \colorbox{gt62!50}{sconce}
and \colorbox{gt63!50}{sconce}) might provide more direct illumination compared to the
\colorbox{gt61!50}{chandelier} in Image 1. This is because sconces typically provide
concentrated light against the wall, enhancing ambient lighting, whereas a
chandelier often disperses light across a larger area and may need additional
fixtures to create focused lighting.
            } 
            && \multicolumn{2}{p{.6\textwidth}}{
            In a dimly lit environment, the \colorbox{prima61!50}{chandelier} in Image 1 might provide more direct illumination due to its multiple light sources, compared to the \colorbox{prima62!50}{sconce} in Image 2, which is designed for localized lighting.
            } \\
        \midrule
            {\centering\rotatebox{90}{\textbf{GLaMM}}} & 
            \includegraphics[width=\linewidth]{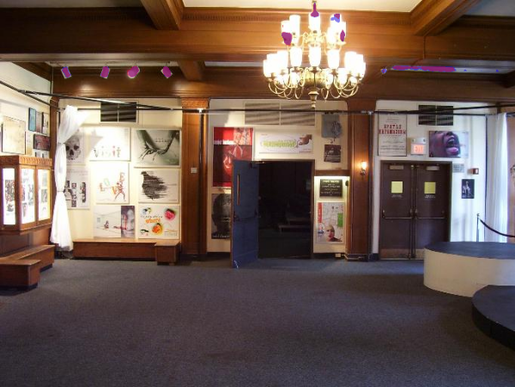} & 
            \includegraphics[width=\linewidth]{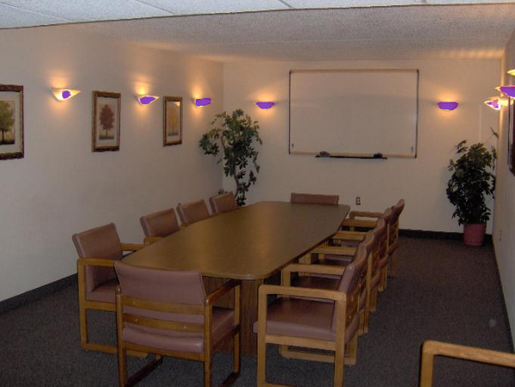} & 
            {\centering\rotatebox{90}{\textbf{LISA}}} & 
            \includegraphics[width=\linewidth]{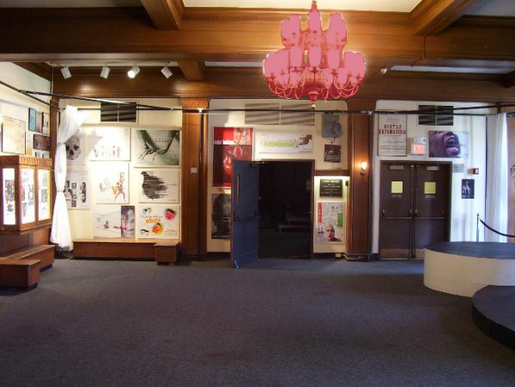} & 
            \includegraphics[width=\linewidth]{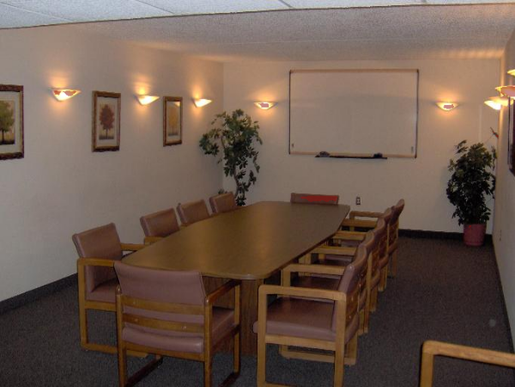} \\
            & \multicolumn{2}{p{.6\textwidth}}{
            In a dimly lit environment, the lights from \colorbox{glamm61!50}{light} and \colorbox{glamm62!50}{light} in Image 1 might provide more direct illumination due to their positioning and size, compared to the smaller lights like \colorbox{glamm63!50}{light} and \colorbox{glamm64!50}{light} in Image 2.
            } 
            && \multicolumn{2}{p{.6\textwidth}}{
            In Image 1, the \colorbox{lisa61!50}{chandelier} provides more direct illumination due to its central position and multiple light sources. In Image 2, the \colorbox{lisa62!50}{lamp} offers a more focused lighting option due to its smaller size and likely direct lighting design.
            } \\
            \midrule
            \multicolumn{6}{c}{\parbox{1.28\linewidth}{
    \centering
    \textbf{
      \colorbox{BackgroundGreen}{\color{TextGreen}{\textit{Functional}}}\quad Q: Which object would allow for better maneuverability in crowded areas?
    }
  }}\\
        \midrule
            {\centering\rotatebox{90}{\textbf{Ground Truth}}} & 
            \includegraphics[height=3cm]{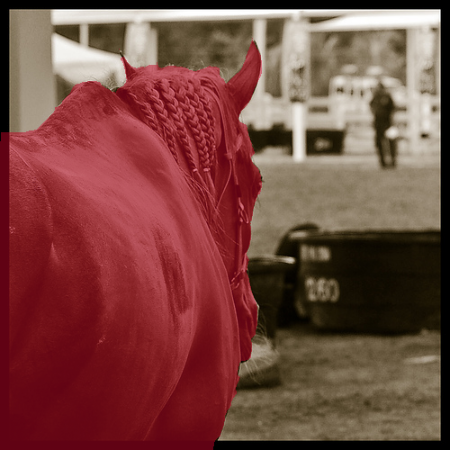} &
            \includegraphics[height=3cm]{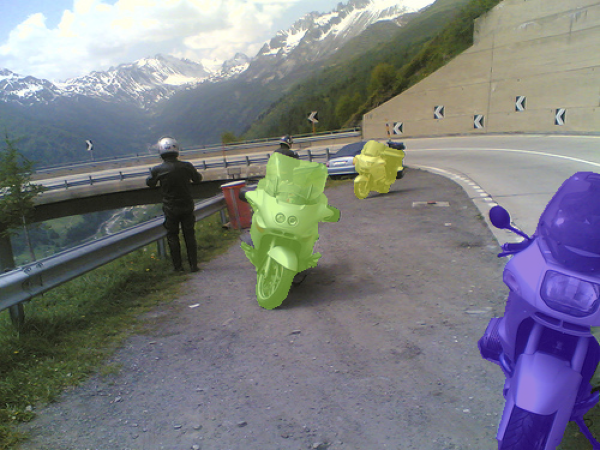} &
            \centering\rotatebox{90}{\textbf{\modelname{}}} & 
            \includegraphics[height=3cm]{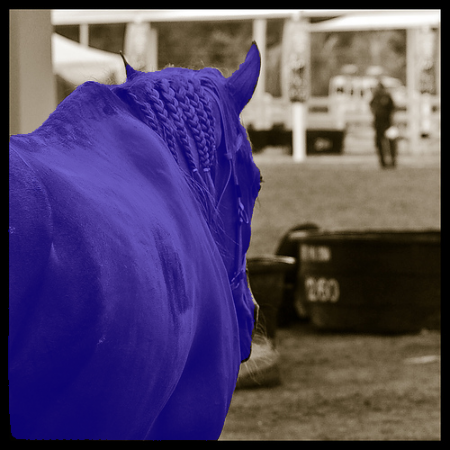} &
            \includegraphics[height=3cm]{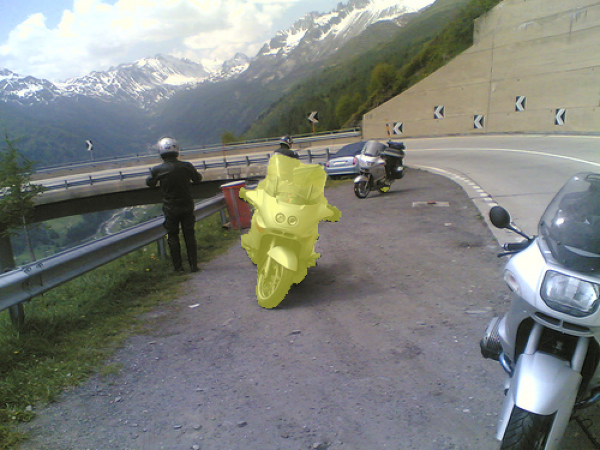} \\
            & \multicolumn{2}{p{.6\textwidth}}{
            In crowded areas, \colorbox{gt21!50}{horse} may allow for better maneuverability compared to
            \colorbox{gt22!50}{motorbike}, \colorbox{gt23!50}{motorbike}, and \colorbox{gt24!50}{motorbike}, as the horse can adjust its
            path more flexibly in tight spaces.
            }
            && \multicolumn{2}{p{.6\textwidth}}{
            The \colorbox{prima21!50}{motorbike} in Image 2 would allow for better maneuverability in crowded areas compared to the \colorbox{prima22!50}{horse} in Image 1.
            } \\
        \midrule
            \centering\rotatebox{90}{\textbf{GLaMM}} & 
            \includegraphics[height=3cm]{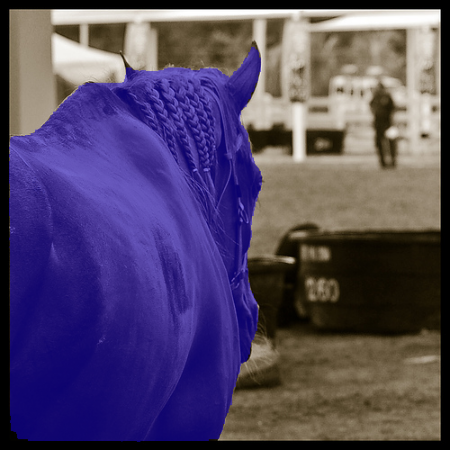} &
            \includegraphics[height=3cm]{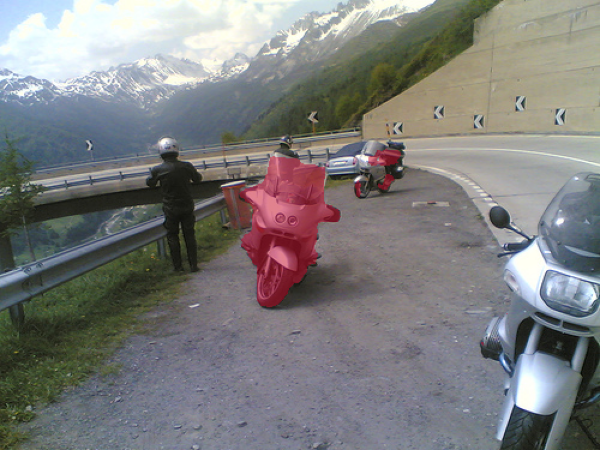} &
            \centering\rotatebox{90}{\textbf{LISA}} & 
            \includegraphics[height=3cm]{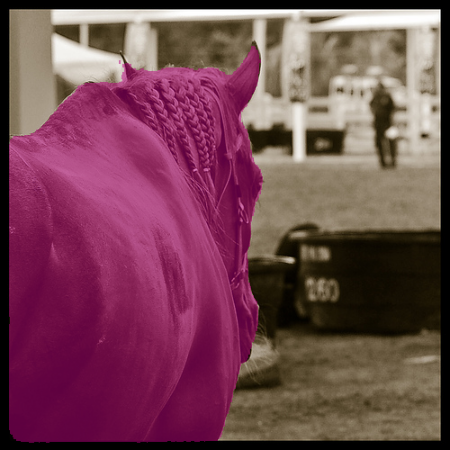} &
            \includegraphics[height=3cm]{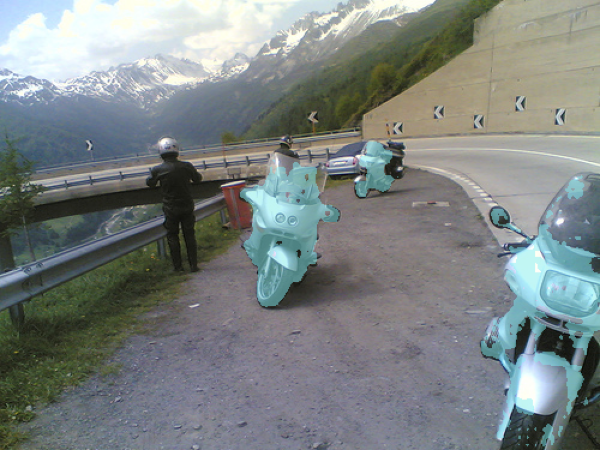} \\
            & \multicolumn{2}{p{.6\textwidth}}{
            The \colorbox{glamm21!50}{motorbike} in Image 2 would allow for better maneuverability in crowded areas compared to the \colorbox{glamm22!50}{truck} in Image 1.
            } 
            && \multicolumn{2}{p{.6\textwidth}}{
            The \colorbox{lisa21!50}{motorbike} in Image 2 would allow for better
            maneuverability in crowded areas compared to the
            \colorbox{lisa22!50}{truck} in Image 1.
            } \\
        \bottomrule
        \end{tabulary}
    }
    \captionof{figure}{\textbf{Additional Qualitative Results.} In the first example, \modelname{} generates accurate masks for the lighting fixtures in both images, whereas the baselines either miss the target objects or produce partial masks.
    In the second example, all models select the motorbike instead of the horse, but \modelname{} grounds the compared objects correctly while both baselines hallucinate a truck.
    }
    \label{tab:suppl:qualitatives1}
    \vspace{-0.3cm}
\end{table*}

\begin{table*}[t!]
    \centering
    \resizebox{0.99\linewidth}{!}{
        \begin{tabulary}{\textwidth}{>{\centering\arraybackslash}p{.04\textwidth}@{} >{\centering\arraybackslash}p{.3\textwidth} >{\centering\arraybackslash}p{.3\textwidth} >{\centering\arraybackslash}p{.04\textwidth}@{} >{\centering\arraybackslash}p{.3\textwidth} >{\centering\arraybackslash}p{.3\textwidth}}
        \toprule
 
            \multicolumn{6}{c}{\parbox{0.99\linewidth}{
    \centering
    \textbf{
      \colorbox{BackgroundCoffee}{\color{TextCoffee}{\textit{Numerical}}}\quad Q: Which image has the largest count of sheep?
    }
  }}\\
        \midrule
            {\centering\rotatebox{90}{\textbf{Ground Truth}}} & 
            \multicolumn{2}{c}{
              \begin{tabular}[b]{@{}ccc@{}}
                \includegraphics[width=.105\linewidth]{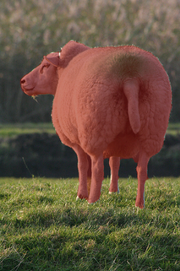}
                \includegraphics[width=.24\linewidth]{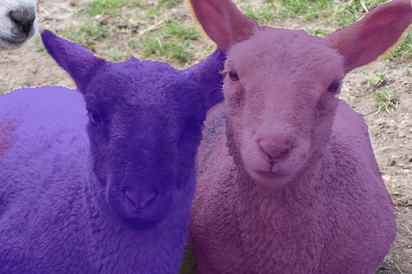}
                \includegraphics[width=.24\linewidth]{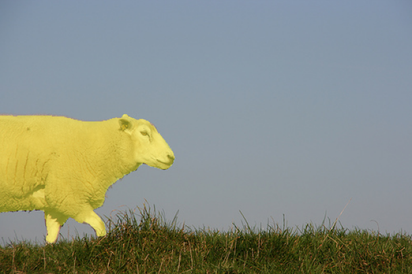}
              \end{tabular}
            } &
            {\centering\rotatebox{90}{\textbf{\modelname{}}}} & 
            \multicolumn{2}{c}{
              \begin{tabular}[b]{@{}ccc@{}}
                \includegraphics[width=.105\linewidth]{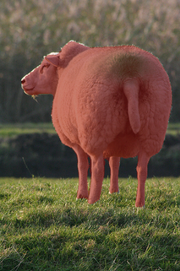}
                \includegraphics[width=.24\linewidth]{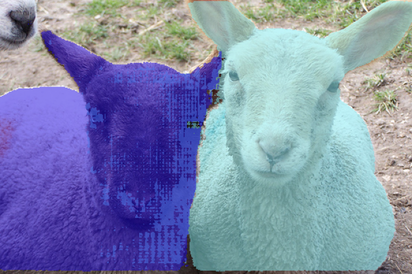}
                \includegraphics[width=.24\linewidth]{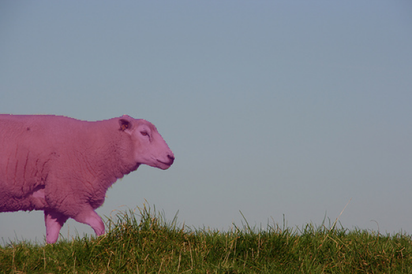}
              \end{tabular}
            } \\
            & \multicolumn{2}{p{.6\textwidth}}{
            There is one \colorbox{gt81!50}{sheep} in Image 1. And, 2 sheep (\colorbox{gt82!50}{sheep}, \colorbox{gt83!50}{sheep}) are present in Image 2. Finally, there is one \colorbox{gt84!50}{sheep} in Image 3. Therefore, the image with the largest count of sheep is Image 2.
}
            && \multicolumn{2}{p{.6\textwidth}}{
            Image 1 has one \colorbox{prima81!50}{sheep} in it. And, Image 2 has 2 sheep (\colorbox{prima82!50}{sheep}, \colorbox{prima83!50}{sheep}) in it. Finally, Image 3 includes one \colorbox{prima84!50}{sheep}. Thus, the image with the largest count of sheep is Image 2.
            } \\
        \midrule
            {\centering\rotatebox{90}{\textbf{GLaMM}}} & 
            \multicolumn{2}{c}{
              \begin{tabular}[b]{@{}ccc@{}}
                \includegraphics[width=.105\linewidth]{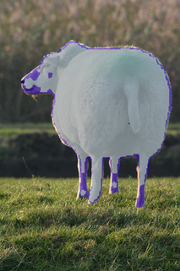}
                \includegraphics[width=.24\linewidth]{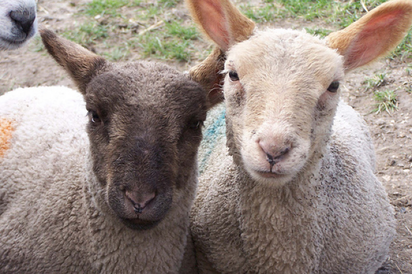}
                \includegraphics[width=.24\linewidth]{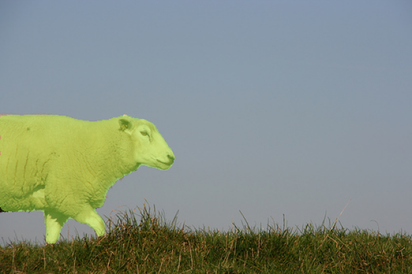}
              \end{tabular}
            } &
            {\centering\rotatebox{90}{\textbf{LISA}}} & 
            \multicolumn{2}{c}{
              \begin{tabular}[b]{@{}ccc@{}}
                \includegraphics[width=.105\linewidth]{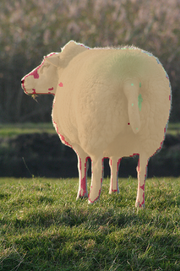}
                \includegraphics[width=.24\linewidth]{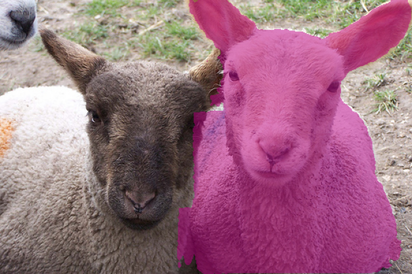}
                \includegraphics[width=.24\linewidth]{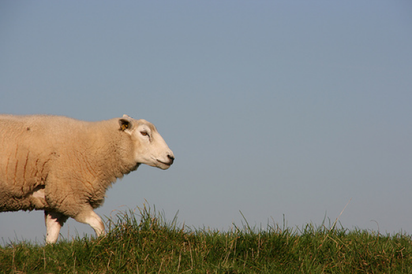}
              \end{tabular}
            } \\
            & \multicolumn{2}{p{.6\textwidth}}{
            Image 1 has 2 sheep (\colorbox{glamm81!50}{sheep}, \colorbox{glamm82!50}{sheep}) in it. And, Image 2 contains a single \colorbox{glamm83!50}{sheep}. Finally, Image 3 contains a single \colorbox{glamm84!50}{sheep}. Hence, the image with the largest count of sheep is Image 2.
            } 
            && \multicolumn{2}{p{.6\textwidth}}{
            Image 1 includes 2 sheep (\colorbox{lisa81!50}{sheep}, \colorbox{lisa82!50}{sheep}). And, Image 2 has one \colorbox{lisa83!50}{sheep} in it. That means the image with the largest count of sheep is Image 2.
            } \\
            \midrule 
            \multicolumn{6}{c}{\parbox{1.28\linewidth}{
    \centering
    \textbf{
      \colorbox{BackgroundGreen}{\color{TextGreen}{\textit{Functional}}}\quad Q: Which furniture item has a more practical storage solution for clothing or personal items, the chest of drawers in the second image or the drawer in the desk from the first image?
    }
  }}\\
        \midrule
            {\centering\rotatebox{90}{\textbf{Ground Truth}}} & 
            \includegraphics[height=3cm]{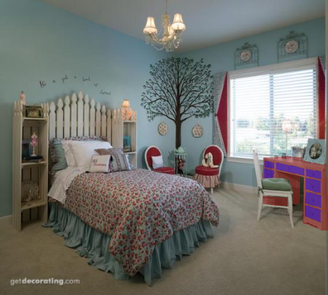} &
            \includegraphics[height=3cm]{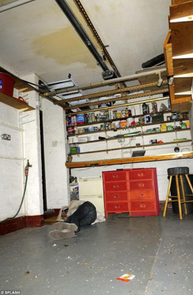} &
            \centering\rotatebox{90}{\textbf{\modelname{}}} & 
            \includegraphics[height=3cm]{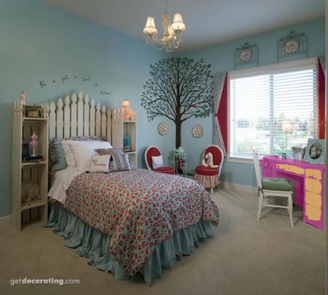} &
            \includegraphics[height=3cm]{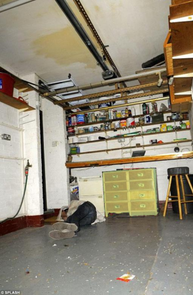} \\
            & \multicolumn{2}{p{.6\textwidth}}{
            The \colorbox{gt103!50}{chest of drawers} from the second image has a more practical storage solution for clothing or personal items compared to the \colorbox{gt102!50}{drawer} in the \colorbox{gt101!50}{desk} from the first image. The chest of drawers typically offers more space with multiple compartments, making it suitable for organizing larger quantities or types of items.
            }
            && \multicolumn{2}{p{.6\textwidth}}{
            The \colorbox{prima103!50}{chest of drawers} in the second image offers a more practical storage solution for clothing or personal items due to its multiple drawers, which provide organized storage space. In contrast, the \colorbox{prima102!50}{desk's drawer} on the \colorbox{prima101!50}{desk} in the first image is relatively small and offers limited capacity, making it less suitable for storing larger or multiple items.
            } \\
        \midrule
            \centering\rotatebox{90}{\textbf{GLaMM}} & 
            \includegraphics[height=3cm]{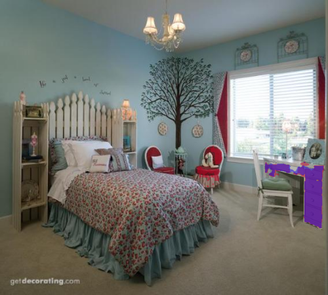} &
            \includegraphics[height=3cm]{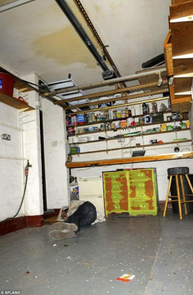} &
            \centering\rotatebox{90}{\textbf{LISA}} & 
            \includegraphics[height=3cm]{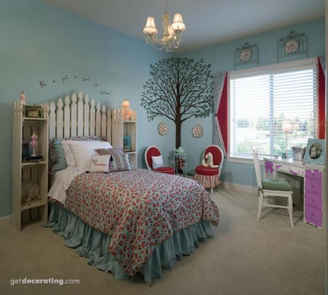} &
            \includegraphics[height=3cm]{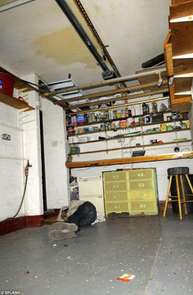} \\
            & \multicolumn{2}{p{.6\textwidth}}{
            The \colorbox{glamm103!50}{chest of drawers} in the second image has a more practical storage solution for clothing or personal items due to its multiple drawers, including \colorbox{glamm104!50}{chest of drawers's drawer}, which allows for organized storage. In contrast, the \colorbox{glamm102!50}{desk} in the first image has only a single \colorbox{glamm101!50}{desk's drawer}, which might not be as efficient for organizing multiple items.
            } 
            && \multicolumn{2}{p{.6\textwidth}}{
            The \colorbox{lisa103!50}{chest of drawers} in the second image provides a more practical storage solution for clothing or personal items due to its multiple drawers, which can be used for organizing various items efficiently. The \colorbox{lisa101!50}{desk} in the first image, while having a \colorbox{lisa102!50}{desk's drawer}, is more suited for office supplies and less for clothing storage.
            } \\
        \bottomrule
        \end{tabulary}
    }
    \captionof{figure}{\textbf{Additional Qualitative Results.}
    The first example highlights \modelname{}’s strong instance segmentation, which enables it to correctly count the sheep; in contrast, the baselines omit some sheep and hallucinate duplicate masks.
    The second example underscores its capacity for cross-context reasoning even when the relevant object (and its parts) appears at a smaller scale in the scene.}
    \label{tab:suppl:qualitatives2}
    \vspace{-0.3cm}
\end{table*}

\begin{table*}[t!]
    \centering
    \resizebox{0.99\linewidth}{!}{
        \begin{tabulary}{\textwidth}{>{\centering\arraybackslash}p{.04\textwidth}@{} >{\centering\arraybackslash}p{.35\textwidth} >{\centering\arraybackslash}p{.35\textwidth} >{\centering\arraybackslash}p{.04\textwidth}@{} >{\centering\arraybackslash}p{.35\textwidth} >{\centering\arraybackslash}p{.35\textwidth}}
        \toprule
            \multicolumn{6}{c}{\parbox{1.48\linewidth}{
    \centering \textbf{\colorbox{BackgroundPink}{\color{TextPink}{\textit{Spatial}}}\quad Q: Considering the arrangement of the sofa and its immediate surroundings, which sofa setup would allow for easier access to a side table from a seated position?}}}\\
        \midrule
            {\centering\rotatebox{90}{\textbf{Ground Truth}}} & 
            \includegraphics[width=\linewidth]{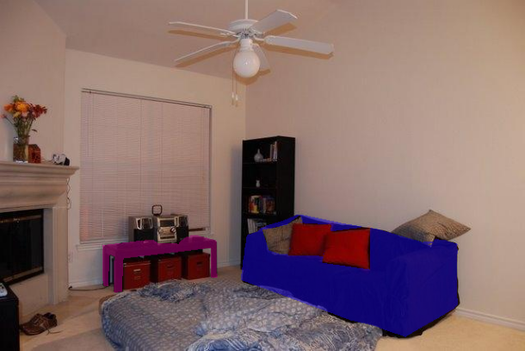} & 
            \includegraphics[width=\linewidth]{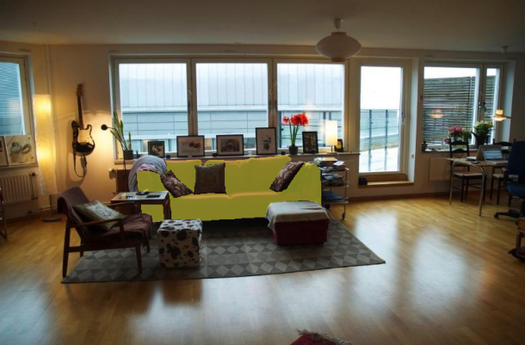} & 
            {\centering\rotatebox{90}{\textbf{\modelname{}}}} & 
            \includegraphics[width=\linewidth]{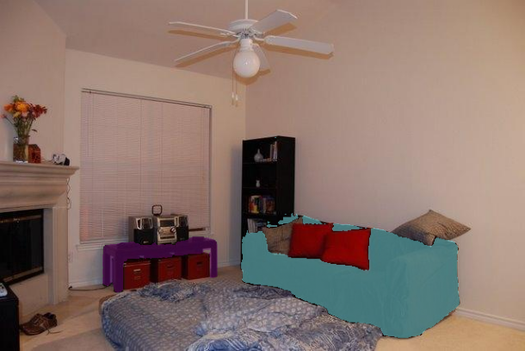} & 
            \includegraphics[width=\linewidth]{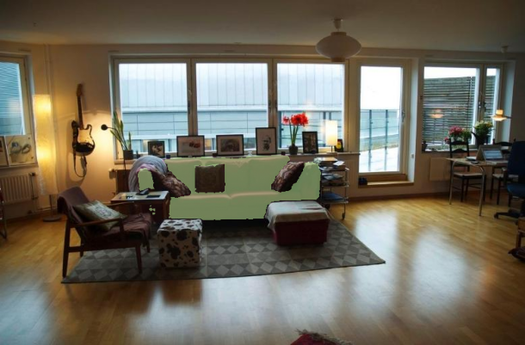} \\
            & \multicolumn{2}{p{.7\textwidth}}{
            The sofa in Image 1, \colorbox{gt111!50}{sofa}, allows for easier access to the side table, \colorbox{gt112!50}{table}, given its closer proximity. In Image 2, \colorbox{gt113!50}{sofa} is placed farther from the nearest table, making access less convenient.
            } 
            && \multicolumn{2}{p{.7\textwidth}}{
            In Image 1, the \colorbox{prima111!50}{sofa} is positioned near the \colorbox{prima112!50}{coffee table}, allowing for easy access to the table from a seated position. In Image 2, the \colorbox{prima113!50}{sofa} is not directly adjacent to a side table, making it less convenient for access.
            } \\
        \midrule
            {\centering\rotatebox{90}{\textbf{GLaMM}}} & 
            \includegraphics[width=\linewidth]{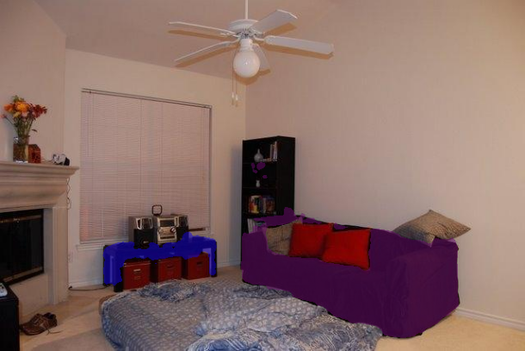} & 
            \includegraphics[width=\linewidth]{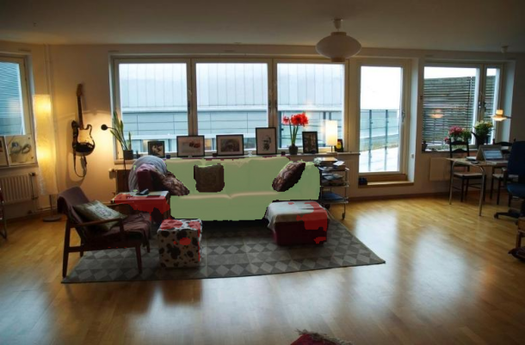} & 
            {\centering\rotatebox{90}{\textbf{LISA}}} & 
            \includegraphics[width=\linewidth]{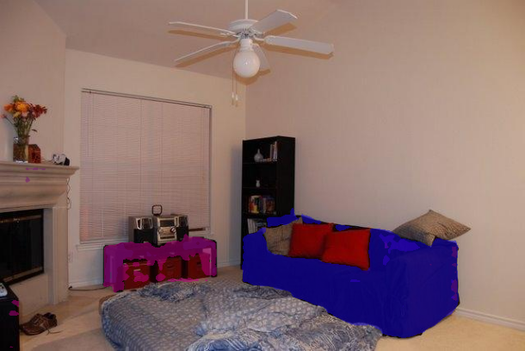} & 
            \includegraphics[width=\linewidth]{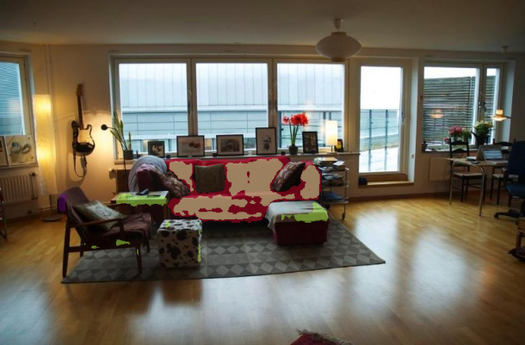} \\
            & \multicolumn{2}{p{.7\textwidth}}{
            In Image 1, the \colorbox{glamm111!50}{sofa} is positioned near the \colorbox{glamm112!50}{coffee table}, allowing for easier access to the table from a seated position. In Image 2, the \colorbox{glamm113!50}{sofa} is not directly adjacent to a side table, as the \colorbox{glamm114!50}{coffee table} is placed further away.
            } 
            && \multicolumn{2}{p{.7\textwidth}}{
            In Image 1, the \colorbox{lisa111!50}{sofa} is positioned with its \colorbox{lisa112!50}{sofa's arm} and \colorbox{lisa113!50}{sofa's back pillow} providing a clear path to the \colorbox{lisa114!50}{coffee table}. In Image 2, the \colorbox{lisa115!50}{sofa} is positioned with its \colorbox{lisa116!50}{sofa's arm} and \colorbox{lisa117!50}{sofa's back pillow} offering access to the \colorbox{lisa118!50}{coffee table}. Both setups allow for easy access to a side table from a seated position, but Image 2's \hallu{coffee table} is more centrally located, making it more convenient for seated individuals.
            } \\
        \midrule

            \multicolumn{6}{c}{\parbox{1.48\linewidth}{
    \centering
    \textbf{
      \colorbox{DarkRed}{\textcolor{white}{Failure case}}\quad\colorbox{BackgroundBlue}{\color{TextBlue}{\textit{Open-Ended}}}\quad Q: 
      Considering the various configurations of car parts seen across the images, 
      which car is positioned in a way that impedes typical access to its components
      and why?
    }
  }}\\
        \midrule
            \multirow{2}{*}{{\centering\rotatebox{90}{\textbf{Ground Truth}}}} & 
            \multicolumn{2}{c}{
              \begin{tabular}{@{}ccc@{}}
                \includegraphics[width=.23\linewidth]{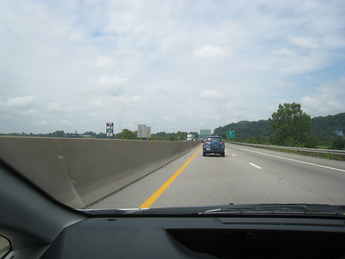}
                \includegraphics[width=.23\linewidth]{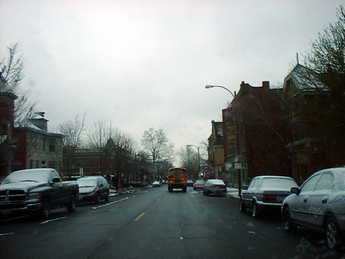}
                \includegraphics[width=.23\linewidth]{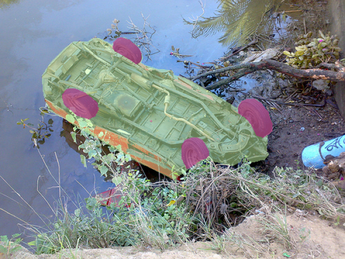}
              \end{tabular}
            } &
            \multirow{2}{*}{\centering\rotatebox{90}{\textbf{\modelname{}}}} & 
            \multicolumn{2}{c}{
              \begin{tabular}{@{}ccc@{}}
                \includegraphics[width=.23\linewidth]{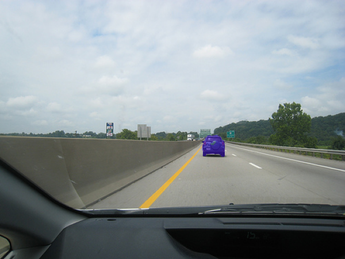}
                \includegraphics[width=.23\linewidth]{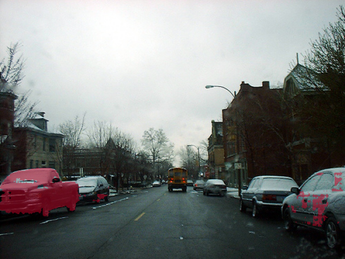}
                \includegraphics[width=.23\linewidth]{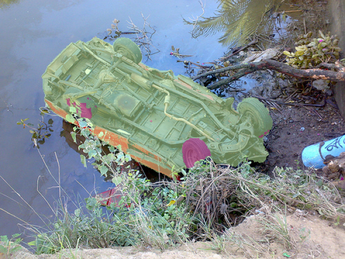}
              \end{tabular}
            } \\
            & \multicolumn{2}{p{.7\textwidth}}{
            The \colorbox{gt71!50}{car} in Image 3 is positioned in a way that impedes typical access to its
components. It is described as being upside down, making it difficult to reach
parts like \colorbox{gt72!50}{wheel} efficiently compared to the other cars, which are in
normal orientations in Images 1 and 2.}
            && \multicolumn{2}{p{.7\textwidth}}{
            In Image 3, the \colorbox{prima73!50}{car} is positioned in a way that impedes typical access to its components. The \colorbox{prima74!50}{car's wheel} is partially submerged in water, making it difficult to access without specialized equipment. This contrasts with the \colorbox{prima71!50}{car} in Image 1 and \colorbox{prima72!50}{car} in Image 2, where the wheels are more accessible due to their placement on the road.
            } \\
        \midrule
            \multirow{2}{*}{\centering\rotatebox{90}{\textbf{GLaMM}}} & 
            \multicolumn{2}{c}{
              \begin{tabular}{@{}ccc@{}}
                \includegraphics[width=.23\linewidth]{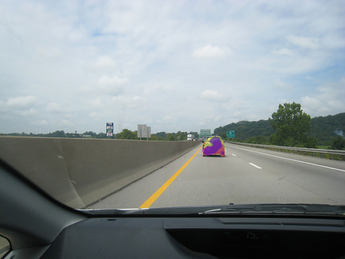}
                \includegraphics[width=.23\linewidth]{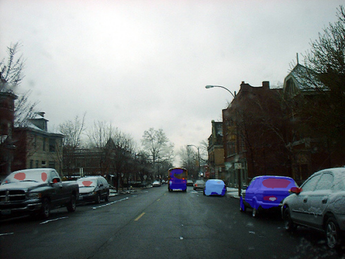}
                \includegraphics[width=.23\linewidth]{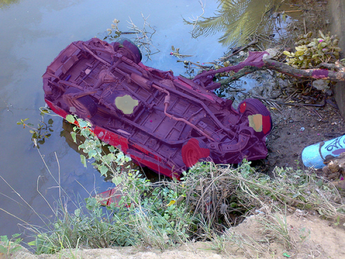}
              \end{tabular}
            } &
            \multirow{2}{*}{\centering\rotatebox{90}{\textbf{LISA}}} & 
            \multicolumn{2}{c}{
              \begin{tabular}{@{}ccc@{}}
                \includegraphics[width=.23\linewidth]{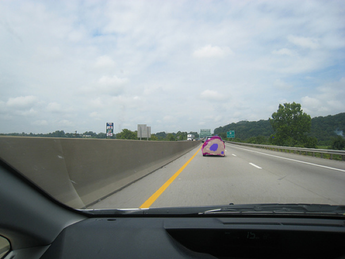}
                \includegraphics[width=.23\linewidth]{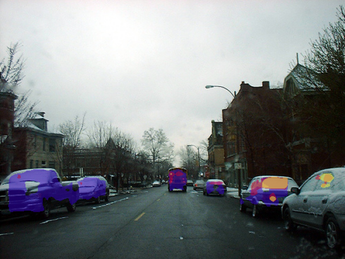}
                \includegraphics[width=.23\linewidth]{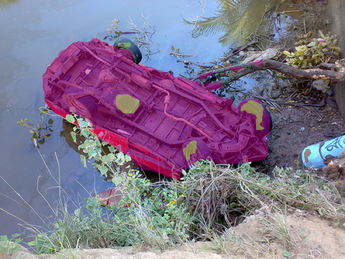}
              \end{tabular}
            } \\
            & \multicolumn{2}{p{.7\textwidth}}{
            In Image 1, the \colorbox{glamm71!50}{car} has its \colorbox{glamm72!50}{car's door} positioned in a way that might impede typical access due to its proximity to the \colorbox{glamm73!50}{car's window}, which could limit visibility and accessibility. In Image 2, \colorbox{glamm74!50}{car} has its \colorbox{glamm75!50}{car's door} positioned in a way that might impede typical access due to its close proximity to the \colorbox{glamm76!50}{car's window}, which could limit visibility and accessibility. In Image 3, \colorbox{glamm77!50}{car} has its \colorbox{glamm78!50}{car's door} positioned in a way that might impede typical access due to its close proximity to the \colorbox{glamm79!50}{car's}, which could limit visibility and accessibility.
            } 
            && \multicolumn{2}{p{.7\textwidth}}{
            In Image 1, the \colorbox{lisa71!50}{car} has its \colorbox{lisa72!50}{car's door} partially obstructed by the \colorbox{lisa73!50}{car's window}, making it less accessible compared to the other cars. In Image 2, \colorbox{lisa74!50}{car} has its \colorbox{lisa75!50}{car's door} partially covered by the \colorbox{lisa76!50}{car's window}, which might hinder access. In Image 3, \colorbox{lisa77!50}{car} has its \colorbox{lisa78!50}{car's window} partially obstructed by the \colorbox{lisa79!50}{car's door}, making it less accessible compared to the others.
            } \\
        \bottomrule
        \end{tabulary}
    }
    \captionof{figure}{\textbf{Additional Qualitative Results.} In the first example, \modelname{} demonstrates solid spatial reasoning without hallucination, in contrast to GLaMM, which hallucinates a coffee table in Image 2, and LISA, which hallucinates multiple objects, produces splotchy masks, and even outputs a grounded phrase without a \texttt{[SEG]} token (marked in \hallu{black}). The second example presents a failure case in which \modelname{} does not segment all four car wheels, which are small and embedded in a cluttered scene. Even so, \modelname{} still demonstrates strong logical and comparative reasoning that enables it to reach the correct answer.}
    \label{tab:suppl:qualitatives3}
    \vspace{-0.3cm}
\end{table*}
\begin{table*}[t!]
    \centering
    \resizebox{0.99\linewidth}{!}{
        \begin{tabulary}{\textwidth}{
            >{\centering\arraybackslash}p{.24\textwidth}
            >{\centering\arraybackslash}p{.24\textwidth}
            >{\centering\arraybackslash}p{.24\textwidth}
            >{\centering\arraybackslash}p{.24\textwidth}
            >{\centering\arraybackslash}p{.24\textwidth}
            >{\centering\arraybackslash}p{.24\textwidth}
        }
        \toprule
        
            \multicolumn{6}{c}{
            \parbox{1.44\linewidth}{\centering \textbf{Q: Which image shows a single serving portion rather than the entire dish?}}}\\
        \midrule
        \multicolumn{2}{c}{\textbf{\modelname{}}} &
            \multicolumn{2}{c}{\textbf{GLaMM}} &
            \multicolumn{2}{c}{\textbf{LISA}} \\
        \midrule
            \includegraphics[height=2cm]{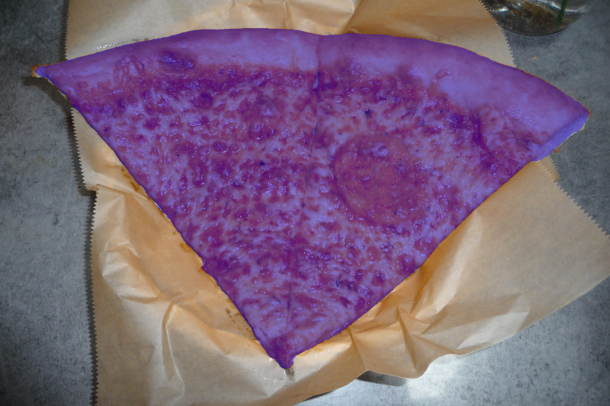} & 
            \includegraphics[height=2cm]{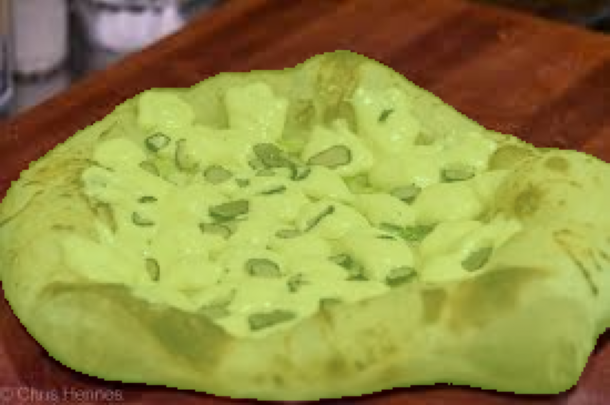} & 
            \includegraphics[height=2cm]{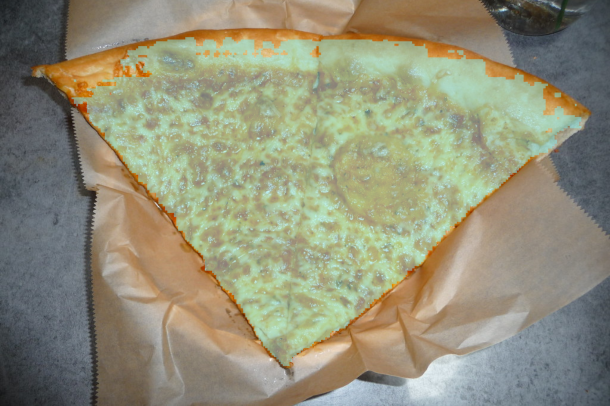} & 
            \includegraphics[height=2cm]{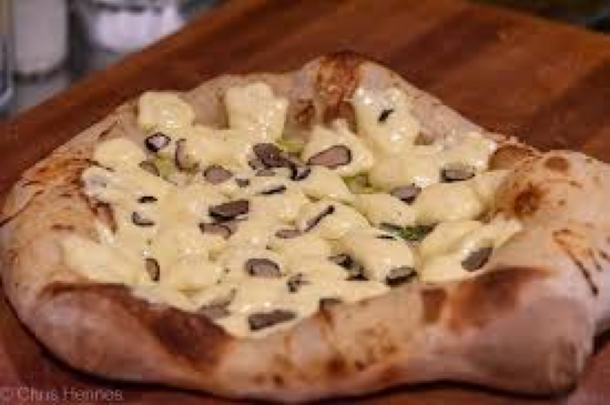} &
            \includegraphics[height=2cm]{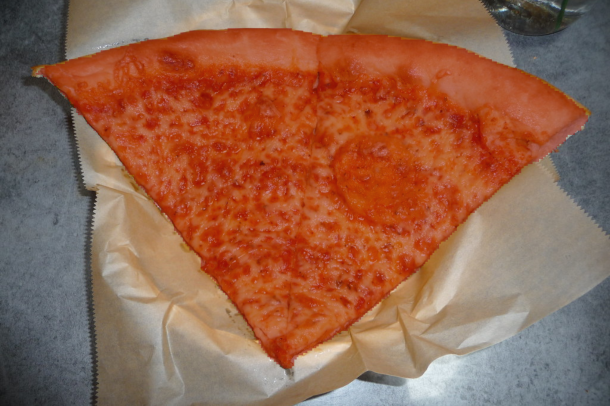} & 
            \includegraphics[height=2cm]{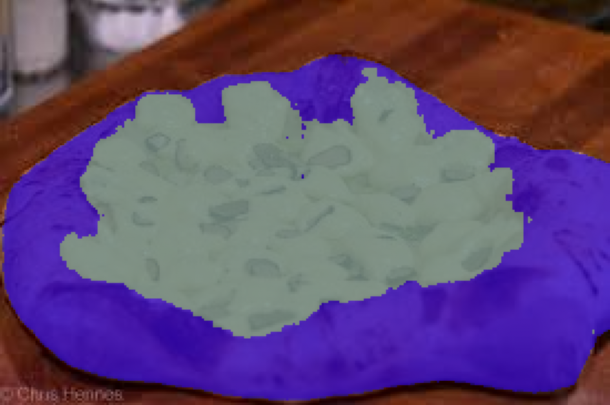} \\
            \multicolumn{2}{p{.48\textwidth}}{
            Image 1 shows a single serving portion, \colorbox{primaitw11!50}{pizza}, which is smaller compared to the larger dish in Image 2, \colorbox{primaitw12!50}{pizza}.
            } 
            &
            \multicolumn{2}{p{.48\textwidth}}{
            Image 2 shows a single serving portion of \colorbox{glammitw12!50}{pizza}, as it only includes a single slice, unlike Image 1, where \colorbox{glammitw11!50}{pizza} is shown with multiple slices.
            } 
            &
            \multicolumn{2}{p{.48\textwidth}}{
            Image 1 shows a single serving portion with \colorbox{lisaitw11!50}{pizza}, whereas Image 2 shows a dish with multiple portions, including \colorbox{lisaitw12!50}{pizza}, \colorbox{lisaitw13!50}{pizza}, and \colorbox{lisaitw14!50}{pizza}.
            } \\
        \midrule
        \multicolumn{6}{c}{
            \parbox{1.44\linewidth}{\centering \textbf{Q: Which accessory is more appropriate for carrying small personal items to a formal event?}}
  }\\
        \midrule
        \multicolumn{2}{c}{\textbf{\modelname{}}} &
            \multicolumn{2}{c}{\textbf{GLaMM}} &
            \multicolumn{2}{c}{\textbf{LISA}} \\
        \midrule
            \includegraphics[height=2cm]{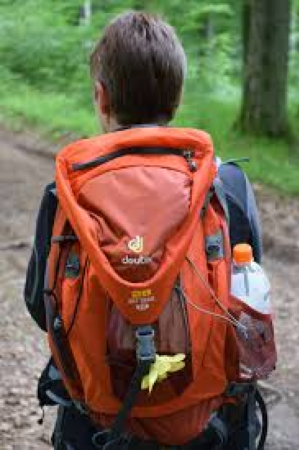} & 
            \includegraphics[height=2cm]{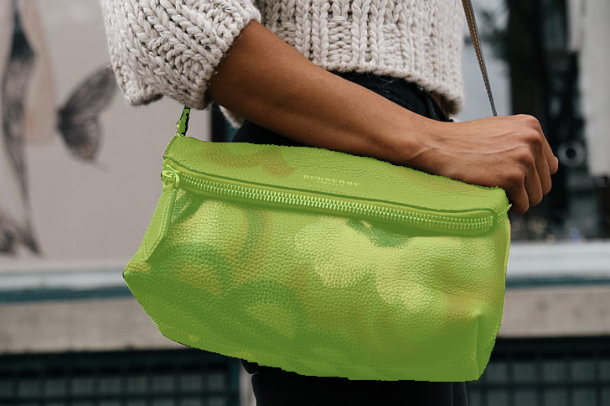} & 
            \includegraphics[height=2cm]{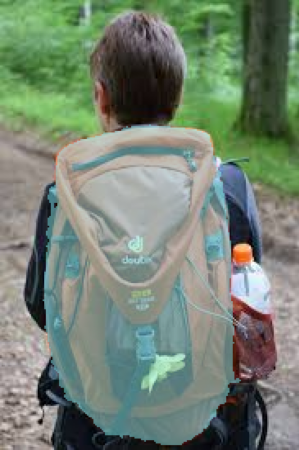} & 
            \includegraphics[height=2cm]{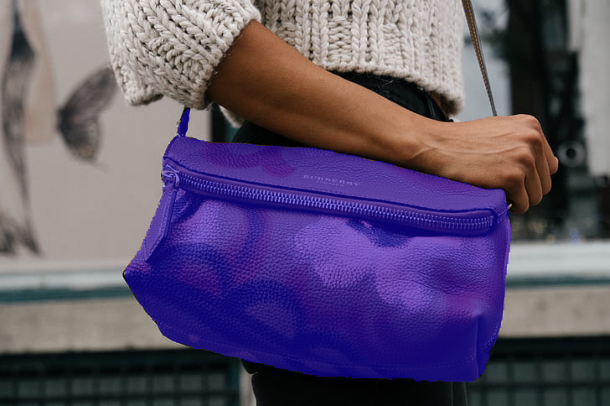} &
            \includegraphics[height=2cm]{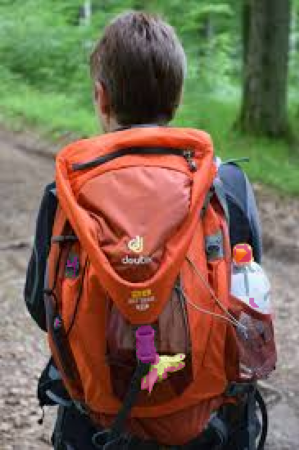} & 
            \includegraphics[height=2cm]{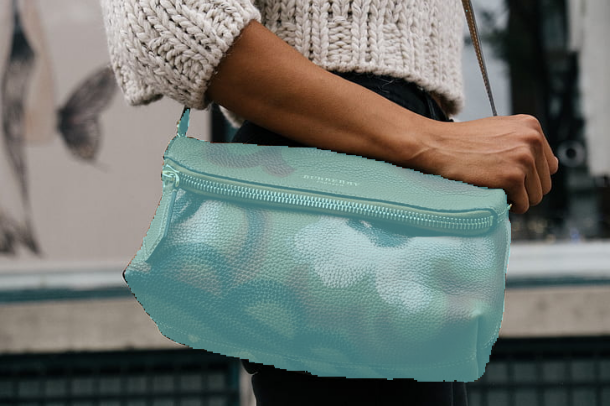} \\
            \multicolumn{2}{p{.48\textwidth}}{
            The \colorbox{primaitw51!50}{handbag} in Image 2 is more appropriate for carrying small personal items to a formal event due to its design and structure, which is typically used for carrying essentials securely.
            } 
            &
            \multicolumn{2}{p{.48\textwidth}}{
            The \colorbox{glammitw52!50}{handbag} in Image 2 is more appropriate for carrying small personal items to a formal event due to its design and size, compared to the \colorbox{glammitw51!50}{watch} in Image 1, which is more suitable for timekeeping rather than carrying items.
            } 
            &
            \multicolumn{2}{p{.48\textwidth}}{
            The \colorbox{lisaitw51!50}{watch} in Image 1 is more appropriate for carrying small personal items to a formal event due to its compact size and formal appearance, compared to the \colorbox{lisaitw52!50}{handbag} in Image 2, which is more casual and less formal.
            } \\

            \midrule
            \multicolumn{6}{c}{
            \parbox{1.44\linewidth}{\centering \textbf{Q: Which tool has a flat face used for striking objects?}}
  }\\
            \midrule
            \multicolumn{2}{c}{\textbf{\modelname{}}} &
            \multicolumn{2}{c}{\textbf{GLaMM}} &
            \multicolumn{2}{c}{\textbf{LISA}} \\
        \midrule
            \includegraphics[height=2cm]{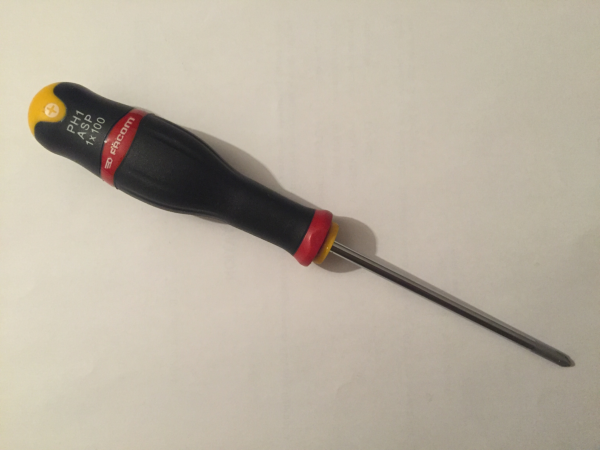} & 
            \includegraphics[height=2cm]{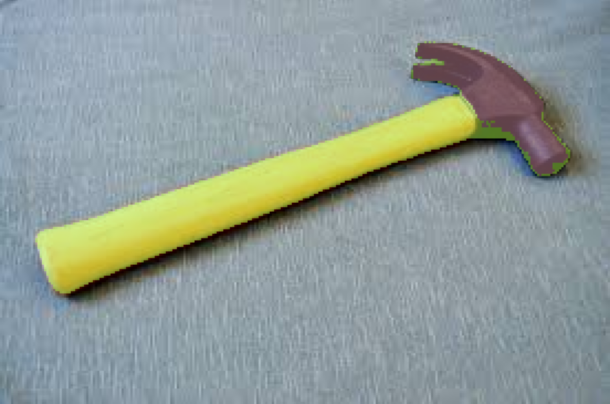} & 
            \includegraphics[height=2cm]{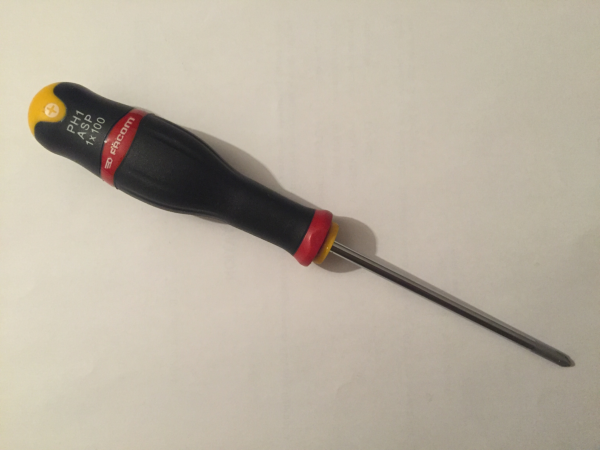} & 
            \includegraphics[height=2cm]{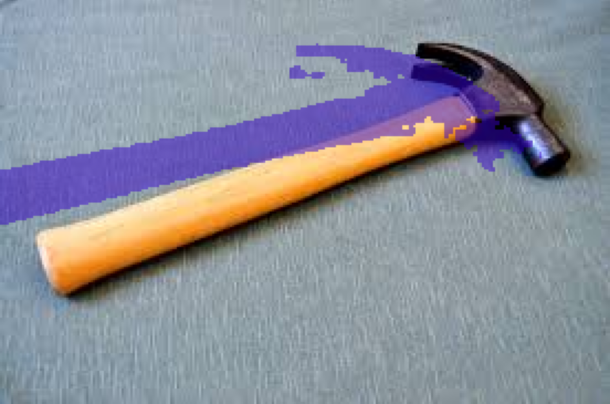} &
            \includegraphics[height=2cm]{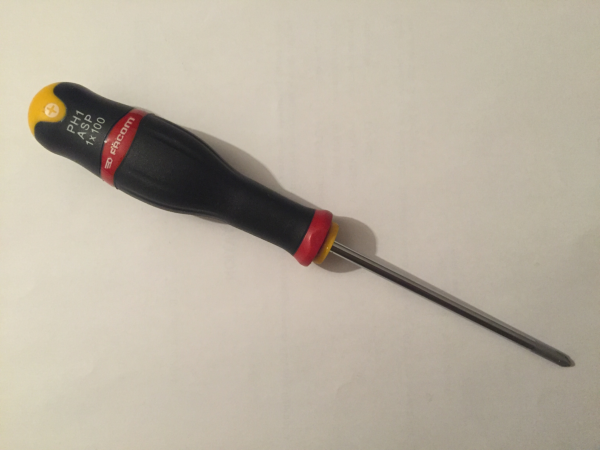} & 
            \includegraphics[height=2cm]{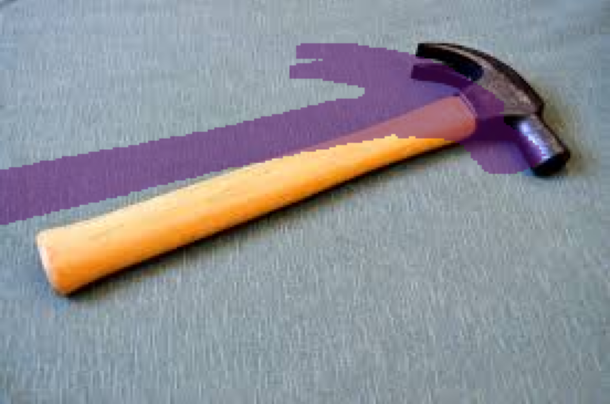} \\
            \multicolumn{2}{p{.48\textwidth}}{
            The \colorbox{primaitw31!50}{hammer} in Image 2 has a flat face, specifically the \colorbox{primaitw32!50}{hammer's face}, which is used for striking objects.
            } 
            &
            \multicolumn{2}{p{.48\textwidth}}{
            The \colorbox{glammitw31!50}{hammer} in Image 2 has a flat face, \colorbox{glammitw32!50}{hammer's face}, which is used for striking objects.
            } 
            &
            \multicolumn{2}{p{.48\textwidth}}{
            The \colorbox{lisaitw31!50}{hammer} in Image 2 has a flat face, specifically the \colorbox{lisaitw32!50}{hammer's face}, used for striking objects.
            } \\
        \midrule
        \multicolumn{6}{c}{\parbox{1.44\linewidth}{%
    \centering
    \textbf{%
      \textbf{Q: Which furniture piece is designed for a single person rather than accommodating multiple people?}
    }%
  }}\\
        \midrule
        \multicolumn{2}{c}{\textbf{\modelname{}}} &
            \multicolumn{2}{c}{\textbf{GLaMM}} &
            \multicolumn{2}{c}{\textbf{LISA}} \\
        \midrule
            \includegraphics[height=2cm]{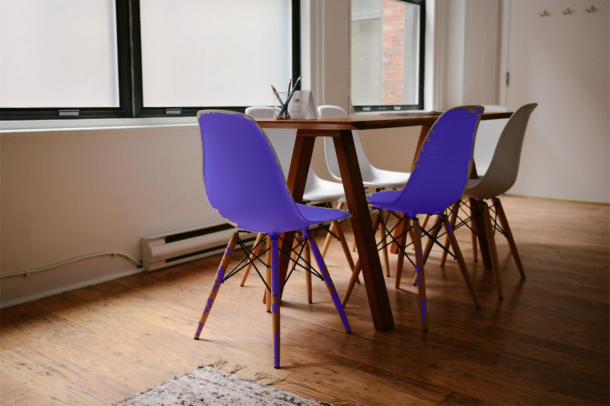} & 
            \includegraphics[height=2cm]{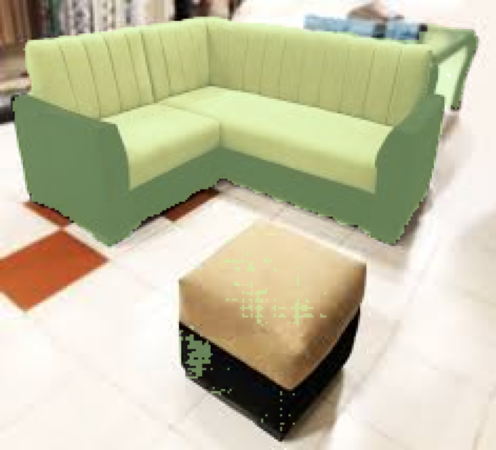} & 
            \includegraphics[height=2cm]{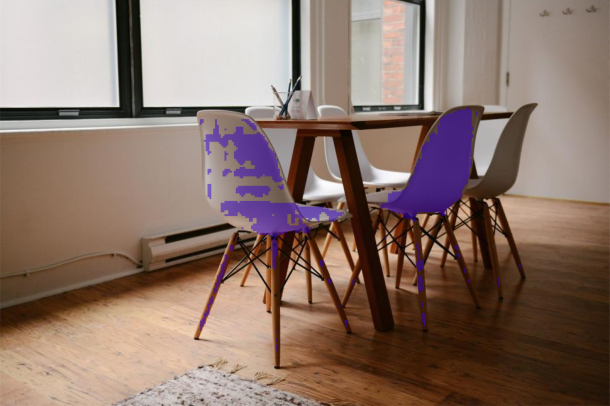} & 
            \includegraphics[height=2cm]{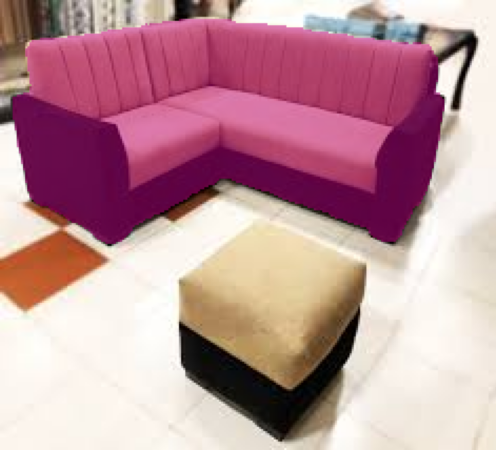} &
            \includegraphics[height=2cm]{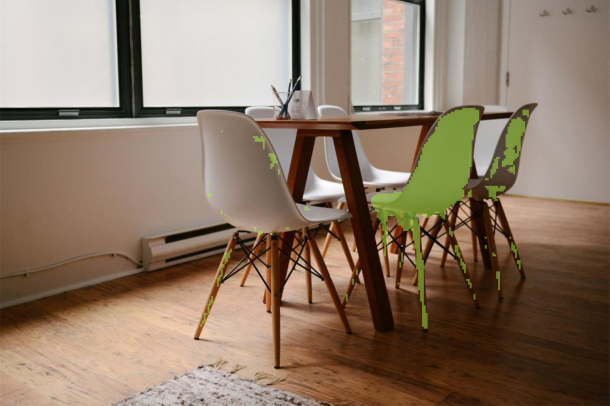} & 
            \includegraphics[height=2cm]{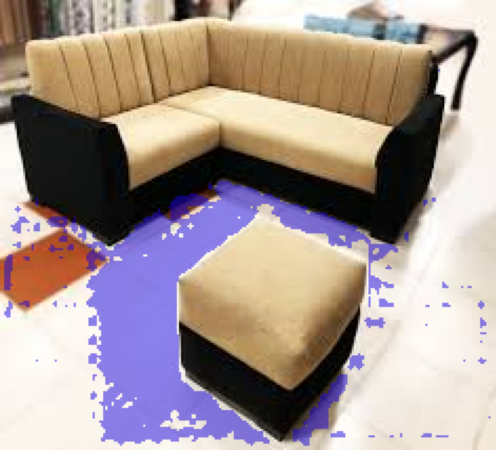} \\
            \multicolumn{2}{p{.48\textwidth}}{
            The \colorbox{primaitw41!50}{chair} in Image 1 is designed for a single person, as it is a single-seat chair, unlike the \colorbox{primaitw42!50}{sofa} in Image 2, which is designed for multiple people to sleep on.
}
            &
            \multicolumn{2}{p{.48\textwidth}}{
            The \colorbox{glammitw41!50}{chair} in Image 1 is designed for a single person, as it is a single-seat chair, unlike the \colorbox{glammitw42!50}{sofa} in Image 2, which is designed for multiple people to sit together.
            } 
            &
            \multicolumn{2}{p{.48\textwidth}}{
            The \colorbox{lisaitw41!50}{chair} in Image 1 is designed for a single person, whereas the \colorbox{lisaitw42!50}{table} in Image 2 is designed to accommodate multiple people.
            } \\
            \midrule
            \multicolumn{6}{c}{\parbox{1.44\linewidth}{%
    \centering
    \textbf{%
      \textbf{Q: Which vehicle is designed to transport a large group of public passengers simultaneously?}
    }%
  }}\\
        \midrule
        \multicolumn{2}{c}{\textbf{\modelname{}}} &
            \multicolumn{2}{c}{\textbf{GLaMM}} &
            \multicolumn{2}{c}{\textbf{LISA}} \\
        \midrule
            \includegraphics[height=2cm]{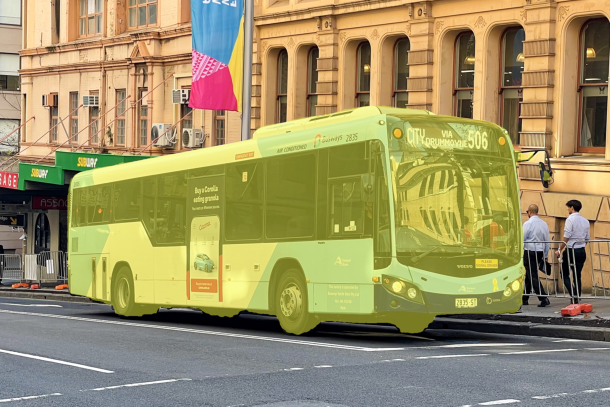} & 
            \includegraphics[height=2cm,width=\linewidth]{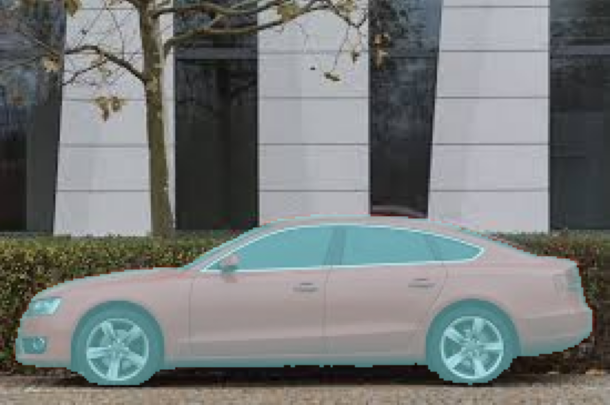} & 
            \includegraphics[height=2cm]{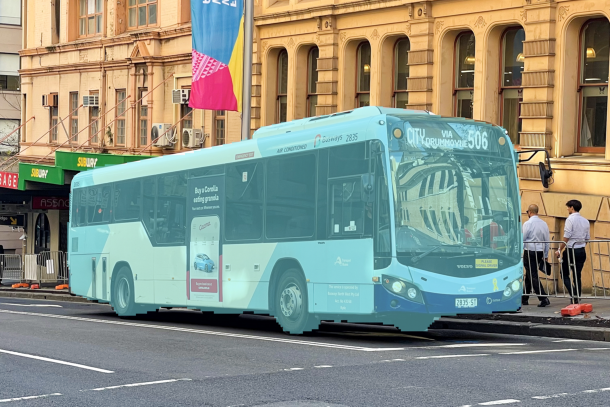} & 
            \includegraphics[height=2cm,width=\linewidth]{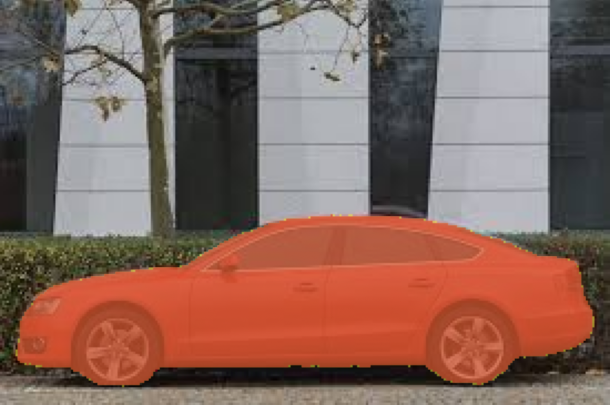} &
            \includegraphics[height=2cm]{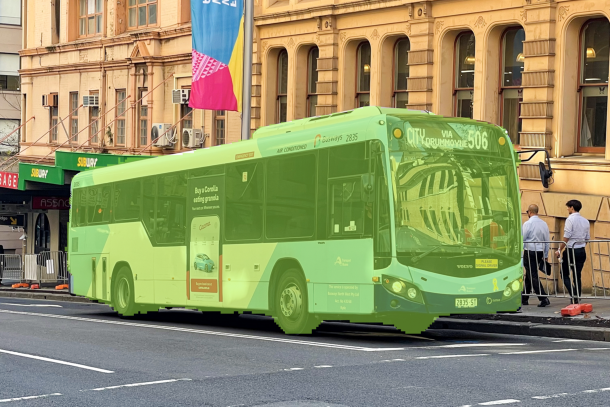} & 
            \includegraphics[height=2cm,width=\linewidth]{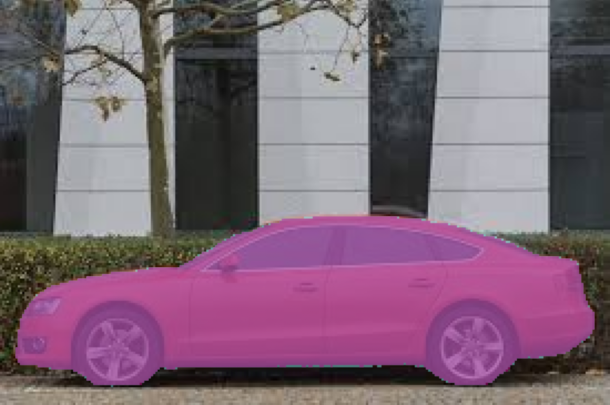} \\
            \multicolumn{2}{p{.48\textwidth}}{
            The \colorbox{primaitw61!50}{bus} in Image 1 is designed to transport a large group of public passengers simultaneously, unlike the \colorbox{primaitw62!50}{car} in Image 2, which is designed for individual or small group transportation.
}
            &
            \multicolumn{2}{p{.48\textwidth}}{
            The \colorbox{glammitw61!50}{bus} is designed to transport a large group of public passengers simultaneously, unlike the \colorbox{glammitw62!50}{car} and \colorbox{glammitw63!50}{car}, which are designed for individual or small group transport.
            } 
            &
            \multicolumn{2}{p{.48\textwidth}}{
            The \colorbox{lisaitw61!50}{bus} is designed to transport a large group of public passengers simultaneously, unlike the \colorbox{lisaitw62!50}{car} and \colorbox{lisaitw63!50}{car}.
            } \\
        \bottomrule
        \end{tabulary}
    }
    \captionof{figure}{\textbf{\modelname{} in-the-wild results on unseen web images requiring complex reasoning.} \modelname{} consistently generates precise segmentation masks and avoids language-driven hallucinations.}
    \label{tab:suppl:in_the_wild}
    \vspace{-0.3cm}
\end{table*}

\begin{table*}[t!]
    \centering
    \resizebox{0.99\linewidth}{!}{
        \begin{tabulary}{\textwidth}{
            >{\centering\arraybackslash}p{.3\textwidth}
            >{\centering\arraybackslash}p{.3\textwidth}|
            >{\centering\arraybackslash}p{.3\textwidth}
            >{\centering\arraybackslash}p{.3\textwidth}}
        \toprule
            \multicolumn{2}{c|}{
                \parbox{.6\textwidth}{\centering
                    \textbf{Q: Considering the presence of multiple dogs across the images, which scenario seems to involve active participation from the dogs in a task?}
                }
            } &
            \multicolumn{2}{c}{
                \parbox{.6\textwidth}{\centering
                    \textbf{Q: Which animal has a longer tail, possibly aiding balance or swatting insects?}
                }
            } \\
        \midrule
            \includegraphics[height=2cm]{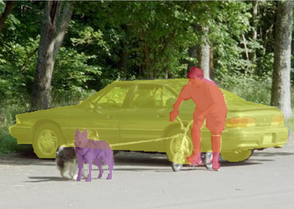} &
            \includegraphics[height=2cm]{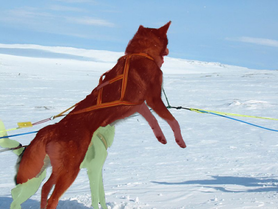} &
            \includegraphics[height=2cm]{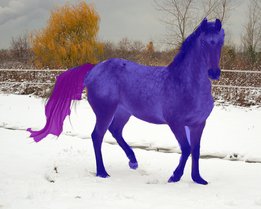} &
            \includegraphics[height=2cm]{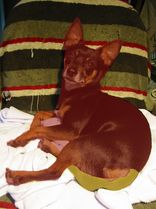} \\
            \multicolumn{2}{p{.6\textwidth}|}{
            In Image 2, the dogs (\colorbox{data44!50}{dog} and \colorbox{data45!50}{dog}) seem actively involved in a task, as they appear to be pulling or moving actively. In contrast, in Image 1, the \colorbox{data43!50}{dog} is stationary near the \colorbox{data41!50}{car} and \colorbox{data42!50}{person}.
            } &
            \multicolumn{2}{p{.6\textwidth}}{
            The \colorbox{data21!50}{horse} in Image 1 has a longer \colorbox{data22!50}{tail} compared to the \colorbox{data23!50}{dog}'s \colorbox{data24!50}{tail} in Image 2, aiding in balance and swatting insects.
            } \\
        \midrule
            \multicolumn{2}{c|}{
                \parbox{.6\textwidth}{\centering
                    \textbf{Q: Which image presents a more space-efficient sink and faucet arrangement for a compact room?}
                }
            } &
            \multicolumn{2}{c}{
                \parbox{.6\textwidth}{\centering
                    \textbf{Q: Which windowpane across all images provides the best view of the outside environment?}
                }
            } \\
        \midrule
            \multicolumn{2}{c|}{
              \begin{tabular}{@{}ccc@{}}
                \includegraphics[height=1.5cm]{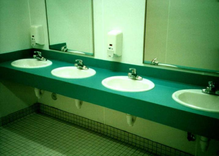}
                \includegraphics[height=1.5cm]{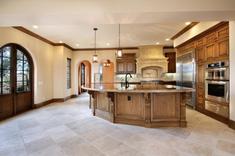}
                \includegraphics[height=1.5cm]{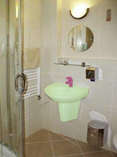}
              \end{tabular}
            } &
            \multicolumn{2}{c}{
              \begin{tabular}{@{}ccc@{}}
                \includegraphics[height=1.5cm]{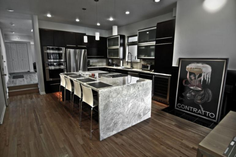}
                \includegraphics[height=1.5cm]{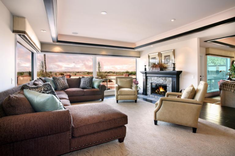}
                \includegraphics[height=1.5cm]{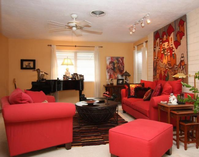}
              \end{tabular}
            } \\
            \multicolumn{2}{p{.6\textwidth}|}{
            Image 3 presents a more space-efficient arrangement with \colorbox{data31!50}{sink} and \colorbox{data32!50}{faucet}. The compact setup fits well into smaller spaces without compromising functionality, making it ideal for compact environments.
            } &
            \multicolumn{2}{p{.6\textwidth}}{
            The \colorbox{data11!50}{windowpane} in Image 2 provides the best view of the outside environment as it offers a broad expanse with a visible landscape.
            } \\
        \midrule
            \multicolumn{2}{c|}{
                \parbox{.6\textwidth}{\centering
                    \textbf{Q: Which object, cow or sheep, is more widely represented across the images?}
                }
            } &
            \multicolumn{2}{c}{
                \parbox{.6\textwidth}{\centering
                    \textbf{Q: Which object would require more maintenance in terms of cleaning?}
                }
            } \\
        \midrule
            \multicolumn{2}{c|}{
              \begin{tabular}{@{}ccc@{}}
                \includegraphics[height=2cm]{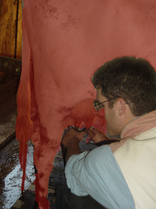}
                \includegraphics[height=2cm]{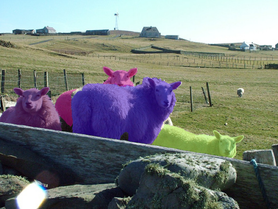}
                \includegraphics[height=2cm]{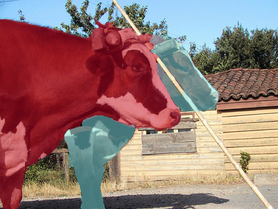}
              \end{tabular}
            } &
            \multicolumn{2}{c}{
              \begin{tabular}{@{}ccc@{}}
                \includegraphics[height=2cm]{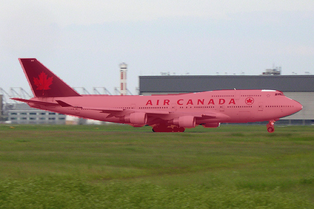}
                \includegraphics[height=2cm]{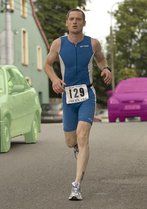}
                \includegraphics[height=2cm]{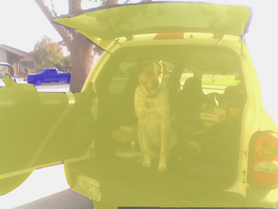}
              \end{tabular}
            } \\
            \multicolumn{2}{p{.6\textwidth}|}{
            Both cows and sheep are represented across the images, but sheep have more individual objects: \colorbox{data52!50}{sheep}, \colorbox{data53!50}{sheep}, \colorbox{data54!50}{sheep}, and \colorbox{data55!50}{sheep} in Image 2, compared to the cows in Images 1 and 3, which are \colorbox{data51!50}{cow}, \colorbox{data56!50}{cow}, and \colorbox{data57!50}{cow}.
            } &
            \multicolumn{2}{p{.6\textwidth}}{
            The \colorbox{data81!50}{airplane} from Image 1 would require more maintenance in terms of cleaning compared to \colorbox{data82!50}{car}, \colorbox{data83!50}{car} from Image 2, and \colorbox{data84!50}{car}, \colorbox{data85!50}{car} from Image 3. Airplanes have larger surfaces and interior spaces, necessitating extensive cleaning protocols due to more frequent and diverse passenger usage.
            } \\
        \midrule
            \multicolumn{2}{c|}{
                \parbox{.6\textwidth}{\centering
                    \textbf{Q: Which dog appears to be in a more relaxed setting based on their posture and surroundings?}
                }
            } &
            \multicolumn{2}{c}{
                \parbox{.6\textwidth}{\centering
                    \textbf{Q: Which computer setup is likely more suitable for collaborative work involving multiple people?}
                }
            } \\
        \midrule
            \includegraphics[height=2cm]{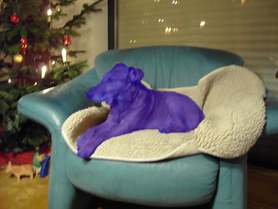} &
            \includegraphics[height=2cm]{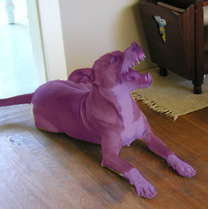} &
            \includegraphics[height=2cm]{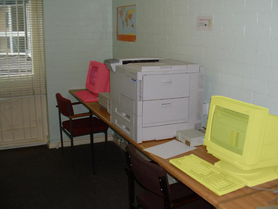} &
            \includegraphics[height=2cm]{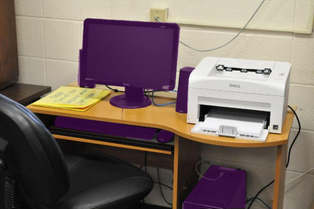} \\
            \multicolumn{2}{p{.6\textwidth}|}{
            The dog in image 1, \colorbox{data61!50}{dog}, appears to be in a more relaxed setting as it is lying on a cushioned chair. In contrast, \colorbox{data62!50}{dog} in image 2 is on a wooden floor and seems more alert or active.
            } &
            \multicolumn{2}{p{.6\textwidth}}{
            The setup with \colorbox{data71!50}{computer} and \colorbox{data72!50}{computer} from Image 1 is likely more suitable for collaborative work involving multiple people, as it features multiple computers in the same space. In contrast, Image 2 features a single \colorbox{data73!50}{computer}, which is more suited for individual use.
            } \\
        \bottomrule
        \end{tabulary}
    }
    \captionof{figure}{\textbf{Examples from \benchmarkname{}: \colorbox{BackgroundBlue}{\color{TextBlue}{\textit{Open-Ended}}} Questions.}}
    \label{tab:suppl:dataset_examples1}
    \vspace{-0.3cm}
\end{table*}

\begin{table*}[t!]
    \centering
    \resizebox{0.99\linewidth}{!}{
        \begin{tabulary}{\textwidth}{
            >{\centering\arraybackslash}p{.3\textwidth}
            >{\centering\arraybackslash}p{.3\textwidth}|
            >{\centering\arraybackslash}p{.3\textwidth}
            >{\centering\arraybackslash}p{.3\textwidth}}
        \toprule
            \multicolumn{2}{c|}{
                \parbox{.6\textwidth}{\centering
                    \textbf{Q: Which vehicle is more suitable for transporting a large group of people over long distances?}
                }
            } &
            \multicolumn{2}{c}{
                \parbox{.6\textwidth}{\centering
                    \textbf{Q: In a scenario where a single person needs to travel with minimal waiting time, which object provides the best solution?}
                }
            } \\
        \midrule
            \includegraphics[height=2cm]{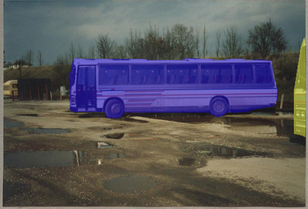} &
            \includegraphics[height=2cm]{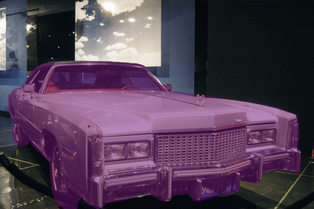} &
            \includegraphics[height=2cm]{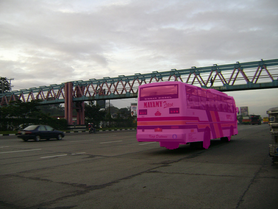} &
            \includegraphics[height=2cm]{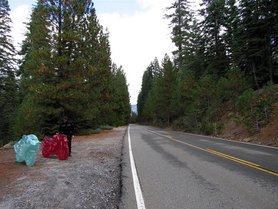} \\
            \multicolumn{2}{p{.6\textwidth}|}{
            The \colorbox{data111!50}{bus} and \colorbox{data112!50}{bus} in Image 1 are more suitable for transporting a large group of people over long distances due to their larger passenger capacity compared to the \colorbox{data113!50}{car} in Image 2, which is designed for fewer passengers.
            } &
            \multicolumn{2}{p{.6\textwidth}}{
            In this scenario, the motorbikes \colorbox{data122!50}{motorbike} and \colorbox{data123!50}{motorbike} from Image 2 provide the best solution for a single person to travel with minimal waiting time, as they are ready for immediate use and can weave through traffic more easily than the \colorbox{data121!50}{bus} in Image 1, which operates on a fixed schedule and route.
            } \\
        \midrule
            \multicolumn{2}{c|}{
                \parbox{.6\textwidth}{\centering
                    \textbf{Q: Which object would be more effective at removing cooking odors in these kitchens?}
                }
            } &
            \multicolumn{2}{c}{
                \parbox{.6\textwidth}{\centering
                    \textbf{Q: Which type of vehicle, across the images, is likely more suitable for carrying larger cargo?}
                }
            } \\
        \midrule
            \multicolumn{2}{c|}{
              \begin{tabular}{@{}ccc@{}}
                \includegraphics[height=1.5cm]{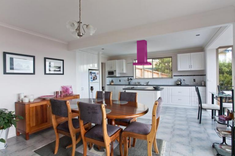}
                \includegraphics[height=1.5cm]{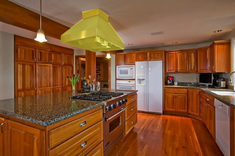}
                \includegraphics[height=1.5cm]{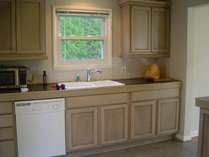}
              \end{tabular}
            } 
            &
            \multicolumn{2}{c}{
              \begin{tabular}{@{}ccc@{}}
                \includegraphics[height=1.5cm]{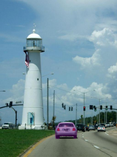}
                \includegraphics[height=1.5cm]{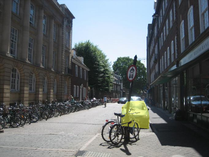}
                \includegraphics[height=1.5cm]{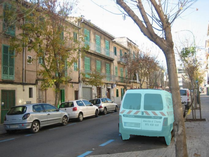}
              \end{tabular}
            } \\
            \multicolumn{2}{p{.6\textwidth}|}{
            The \colorbox{data91!50}{hood} in Image 1 and \colorbox{data92!50}{hood} in Image 2 would be more effective at removing cooking odors compared to Image 3, which lacks a hood.
            } &
            \multicolumn{2}{p{.6\textwidth}}{
            The vehicle type more suitable for carrying larger cargo is the \colorbox{data103!50}{van}, found in Image 3. Vans are generally designed with more space and capacity for transporting goods compared to a typical car, such as \colorbox{data101!50}{car} in Image 1 or \colorbox{data102!50}{car} in Image 2.
            } \\
        \midrule
            \multicolumn{2}{c|}{
                \parbox{.6\textwidth}{\centering
                    \textbf{Q: Are the objects used for similar purposes across the two images?}
                }
            } &
            \multicolumn{2}{c}{
                \parbox{.6\textwidth}{\centering
                    \textbf{Q: In an urban setting with narrow streets, which object is more maneuverable and suitable for quick, individual travel?}
                }
            } \\
        \midrule
            \includegraphics[height=2cm]{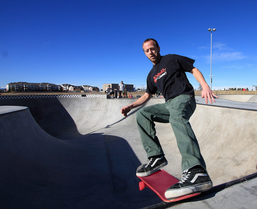} &
            \includegraphics[height=2cm]{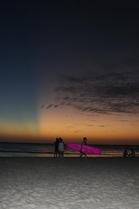}
            &
            \includegraphics[height=2cm]{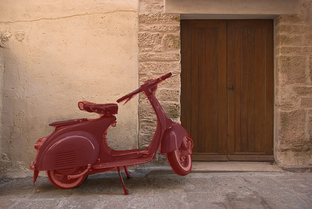} &
            \includegraphics[height=2cm]{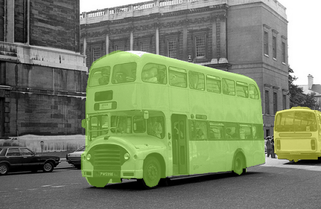} \\
            \multicolumn{2}{p{.6\textwidth}|}{
            No, the skateboards in Image 1, represented by \colorbox{data141!50}{skateboard}, are used for activities on land and in skate parks, while the surfboard in Image 2, exemplified by \colorbox{data142!50}{surfboard}, is used for surfing on water.
            } &
            \multicolumn{2}{p{.6\textwidth}}{
            The \colorbox{data151!50}{motorbike} in Image 1 is more maneuverable and suitable for quick, individual travel through narrow streets compared to the larger vehicles like \colorbox{data152!50}{bus} or \colorbox{data153!50}{bus} in Image 2.
            } \\
        \midrule
            \multicolumn{2}{c|}{
                \parbox{.6\textwidth}{\centering
                    \textbf{Q: Which category of animals is more suitable for agricultural work such as herding on open terrain?}
                }
            } &
            \multicolumn{2}{c}{
                \parbox{.6\textwidth}{\centering
                    \textbf{Q: Imagine a situation where you need to transport people through difficult terrain without access to fuel. Which object in the images would best fulfill this requirement?}
                }
            } \\
        \midrule
            \includegraphics[height=2cm]{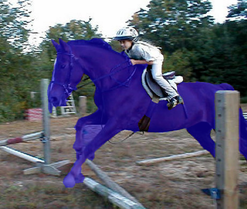} &
            \includegraphics[height=2cm]{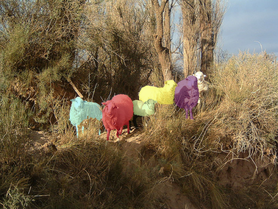} &
            \includegraphics[height=2cm]{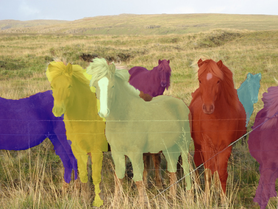} &
            \includegraphics[height=2cm]{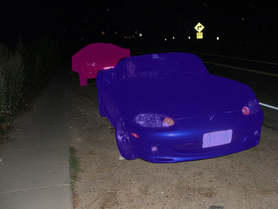} \\
            \multicolumn{2}{p{.6\textwidth}|}{
            For agricultural work such as herding on open terrain, sheep are typically more suitable. In Image 2, multiple sheep, {\colorbox{data132!50}{sheep}}, {\colorbox{data133!50}{sheep}}, {\colorbox{data134!50}{sheep}}, {\colorbox{data135!50}{sheep}}, and {\colorbox{data136!50}{sheep}}, are depicted in a natural setting, indicating their suitability for grazing and herding. In contrast, the horse in Image 1, {\colorbox{data131!50}{horse}}, although useful for riding and transportation, is less commonly used for herding tasks.
            } &
            \multicolumn{2}{p{.6\textwidth}}{
            For transporting people through difficult terrain without access to fuel, horses from Image 1, such as \colorbox{data161!50}{horse}, \colorbox{data162!50}{horse}, \colorbox{data163!50}{horse}, \colorbox{data164!50}{horse}, \colorbox{data165!50}{horse}, \colorbox{data166!50}{horse}, and \colorbox{data167!50}{horse}, would be most suitable. Horses are not dependent on fuel and can navigate uneven and challenging terrain effectively, making them ideal for situations where vehicles like \colorbox{data168!50}{car} and \colorbox{data169!50}{car} from Image 2 cannot operate due to lack of fuel or suitable paths.
            } \\
        \bottomrule
        \end{tabulary}
    }
    \captionof{figure}{\textbf{Examples from \benchmarkname{}: \colorbox{BackgroundGreen}{\color{TextGreen}{\textit{Functional}}} Questions.}}
    \label{tab:suppl:dataset_examples2}
    \vspace{-0.3cm}
\end{table*}

\begin{table*}[t!]
    \centering
    \resizebox{0.95\linewidth}{!}{
        \begin{tabulary}{\textwidth}{
            >{\centering\arraybackslash}p{.25\textwidth}
            >{\centering\arraybackslash}p{.25\textwidth}|
            >{\centering\arraybackslash}p{.25\textwidth}
            >{\centering\arraybackslash}p{.25\textwidth}}
        \toprule
            \multicolumn{2}{c|}{
                \parbox{.5\textwidth}{\centering
                    \textbf{Q: Analyzing the spatial arrangement across images, which boat appears more isolated from other objects, and what could this imply about its usability in a crowded marina?}
                }
            } &
            \multicolumn{2}{c}{
                \parbox{.5\textwidth}{\centering
                    \textbf{Q: If you wanted to move an item outside the door and required ample space for the maneuver, which door would you choose based on surrounding furniture?}
                }
            } \\
        \midrule
            \includegraphics[height=2cm]{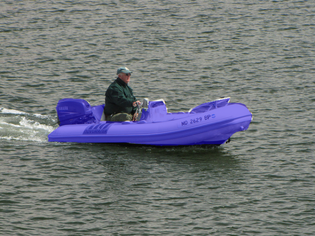} &
            \includegraphics[height=2cm]{media/dataset_examples/data_171_masked.png} &
            \includegraphics[height=2cm]{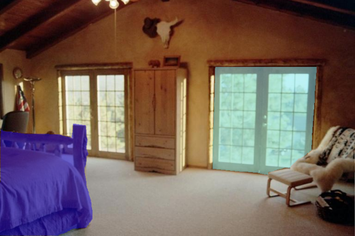} &
            \includegraphics[height=2cm]{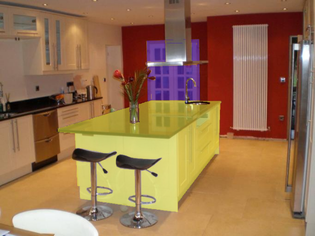} \\
            \multicolumn{2}{p{.5\textwidth}|}{
            \colorbox{data171!50}{Boat} in Image 1 appears more isolated from other objects, as it is on open water without nearby boats. This implies it has more immediate maneuverability and usability compared to the boats in Image 2, which are closely docked and restricted by their surroundings.
            } &
            \multicolumn{2}{p{.5\textwidth}}{
            Based on the surrounding furniture, the \colorbox{data211!50}{door} in Image 1 provides ample space for maneuvering an item outside, as the area is more open with only the \colorbox{data212!50}{bed} further away from the door. In contrast, the \colorbox{data214!50}{door} in Image 2 is surrounded by the \colorbox{data213!50}{kitchen island} and various cabinets, limiting available space for easy movement.
            } \\
        \midrule
            \multicolumn{2}{c|}{
                \parbox{.5\textwidth}{\centering
                    \textbf{Q: In which image are more swivel chairs positioned symmetrically along a line, and how does that arrangement affect potential usability in each space?}
                }
            } &
            \multicolumn{2}{c}{
                \parbox{.5\textwidth}{\centering
                    \textbf{Q: Which sink seems to provide the most counter space for placing items, considering the surrounding spatial arrangement across the images?}
                }
            } \\
        \midrule
            \includegraphics[height=2cm]{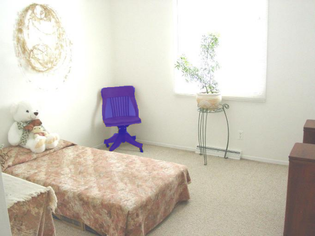} &
                \includegraphics[height=2cm]{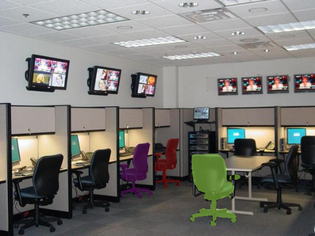}
            &
            \multicolumn{2}{c}{
              \raisebox{5.5ex}{\begin{tabular}{@{}ccc@{}}
                \includegraphics[height=2cm]{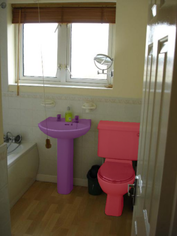}
                \includegraphics[height=2cm]{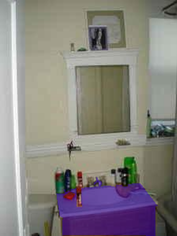}
                \includegraphics[height=2cm]{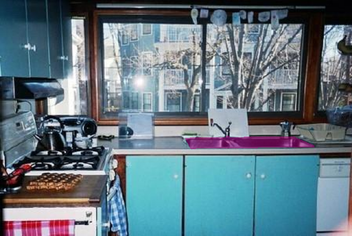}
              \end{tabular}}
            } \\
            \multicolumn{2}{p{.5\textwidth}|}{
            In Image 2, more swivel chairs are positioned symmetrically along a line, specifically the chairs \colorbox{data192!50}{swivel chair}, \colorbox{data193!50}{swivel chair}, and \colorbox{data194!50}{swivel chair}, among others, forming a row. This arrangement is conducive to usability in a workspace setting, as it suggests a structured environment that could enhance accessibility and movement along the stations. In contrast, Image 1 only includes one swivel chair, \colorbox{data191!50}{swivel chair}, which does not contribute to a symmetric arrangement in that context.
            } &
            \multicolumn{2}{p{.5\textwidth}}{
            The sink in Image 3, \colorbox{data204!50}{sink}, appears to provide the most counter space due to its positioning within the kitchen counter, which extends on both sides. The sinks in Image 1, \colorbox{data202!50}{sink}, and Image 2, \colorbox{data203!50}{sink}, are less likely to offer additional space as they are situated closer to other fixtures like the \colorbox{data201!50}{toilet} and bathroom items, respectively.
            } \\
        \bottomrule
        \end{tabulary}
    }
    \captionof{figure}{\textbf{Examples from \benchmarkname{}: \colorbox{BackgroundPink}{\color{TextPink}{\textit{Spatial}}} Questions.}}
    \label{tab:suppl:dataset_examples3}
    \vspace{-0.3cm}
\end{table*}

\begin{table*}[t!]
    \centering
    \resizebox{0.99\linewidth}{!}{
        \begin{tabulary}{\textwidth}{
            >{\centering\arraybackslash}p{.25\textwidth}
            >{\centering\arraybackslash}p{.25\textwidth}|
            >{\centering\arraybackslash}p{.25\textwidth}
            >{\centering\arraybackslash}p{.25\textwidth}}
        \toprule
            \multicolumn{2}{c|}{
                \parbox{.5\textwidth}{\centering
                    \textbf{Q: Which of the two images has a greater number of sandwiches and by how many?}
                }
            } &
            \multicolumn{2}{c}{
                \parbox{.5\textwidth}{\centering
                    \textbf{Q: Between the two images, which one contains more zebras and what is the difference?}
                }
            } \\
        \midrule
            \includegraphics[height=2cm]{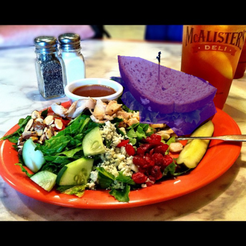} &
            \includegraphics[height=2cm]{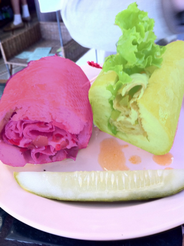} &
            \includegraphics[height=2cm]{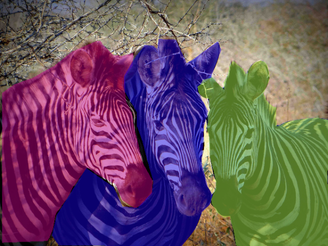} &
            \includegraphics[height=2cm]{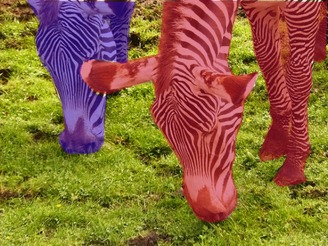} \\
            \multicolumn{2}{p{.5\textwidth}|}{
            In Image 1, there is one \colorbox{data221!50}{sandwich}. On the other hand, Image 2 has 2 sandwiches (\colorbox{data222!50}{sandwich}, \colorbox{data223!50}{sandwich}) in it. Therefore, the image with a greater number of sandwich is Image 2, with (2 - 1) = 1 more sandwich.
            } &
            \multicolumn{2}{p{.5\textwidth}}{
            Image 1 features 3 zebras (\colorbox{data231!50}{zebra}, \colorbox{data232!50}{zebra}, \colorbox{data233!50}{zebra}). In contrast, Image 2 features 2 zebras (\colorbox{data234!50}{zebra}, \colorbox{data235!50}{zebra}). Hence, between the two images, Image 1 contains (3 - 2) = 1 more zebra compared to the other image.
            } \\
        \midrule
            \multicolumn{2}{c|}{
                \parbox{.5\textwidth}{\centering
                    \textbf{Q: If we count all the dogs from all images, what is the sum?}
                }
            } &
            \multicolumn{2}{c}{
                \parbox{.5\textwidth}{\centering
                    \textbf{Q: Which of the two images has a smaller number of windowpanes and by how many?}
                }
            } \\
        \midrule
            \includegraphics[height=2cm]{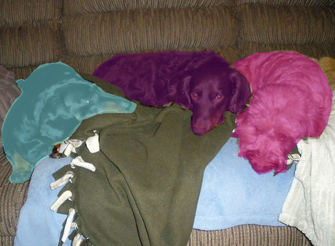} &
            \includegraphics[height=2cm]{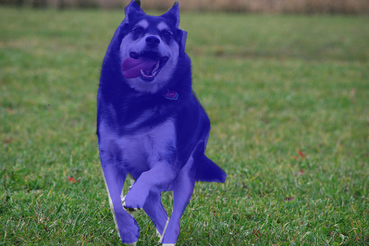} &
            \includegraphics[height=2cm]{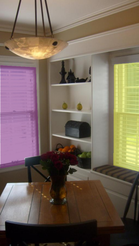} &
            \includegraphics[height=2cm]{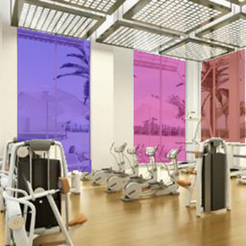} \\
            \multicolumn{2}{p{.5\textwidth}|}{
            Image 1 contains 3 dogs (\colorbox{data251!50}{dog}, \colorbox{data252!50}{dog}, \colorbox{data253!50}{dog}). And, there is one \colorbox{data254!50}{dog} in Image 2. Thus, if we count all the dogs from all images, the sum is (3 + 1) = 4.
            } &
            \multicolumn{2}{p{.5\textwidth}}{
            Image 1 includes 2 windowpanes (\colorbox{data261!50}{windowpane}, \colorbox{data262!50}{windowpane}). But, Image 2 features 3 windowpanes (\colorbox{data263!50}{windowpane}, \colorbox{data264!50}{windowpane}, \colorbox{data265!50}{windowpane}). Therefore, the image with a smaller number of windowpane is Image 1, with (3 - 2) = 1 fewer windowpane.
            } \\
        \bottomrule
        \end{tabulary}
    }
    \captionof{figure}{\textbf{Examples from \benchmarkname{}: \colorbox{BackgroundCoffee}{\color{TextCoffee}{\textit{Numerical}}} Questions.}}
    \label{tab:suppl:dataset_examples4}
    \vspace{-0.3cm}
\end{table*}

\clearpage
\begin{figure*}
    \centering
    \begin{tcolorbox}[
        colback=gray!5!white,
        colframe=black!75!white,
        fonttitle=\bfseries,
        width=0.9\textwidth,
        boxsep=1pt,
        top=1pt,
        bottom=1pt,
        left=1pt,
        right=1pt,
    ]
\begin{lstlisting}[language=Python, basicstyle=\ttfamily\footnotesize,
    keywordstyle=\ttfamily\footnotesize,
    commentstyle=\ttfamily\footnotesize,
    stringstyle=\ttfamily\footnotesize,
            breaklines=true, 
            breakatwhitespace=true, 
    showstringspaces=false,
    breakindent=0pt,
    aboveskip=1pt,   
    belowskip=1pt]
[Images and Image Description]

Your task is to create question-answer pairs using the provided information about the objects and their parts. The goal is to create plausible, challenging questions and provide detailed answers that require multi-image reasoning using the information about objects and parts.

Only create a question if it is possible to generate a high-quality question-answer pair based on the image and the provided information. 

You must generate at least 3 to 5 naturally phrased and coherent question-answer pairs. Questions should be implicit, requiring logical inferences or connections that are not explicitly stated. Each question-answer pair must involve a comparative or relational aspect that incorporates elements from all the images. The objective is to create challenging questions that do not explicitly include visual content details. 

[Question Guidelines]
[Answer Guidelines]
[Format Requirements]

To reiterate, each generated pair must adhere to the guidelines and format requirements, incorporate information from all the provided images, and include references to objects and parts in the answers.
\end{lstlisting}
\end{tcolorbox}
\vspace{-0.3cm}
    \caption{\textbf{Prompt template for open-ended reasoning QA data generation}.}
    \label{fig:prompt_full}
\end{figure*}
\begin{figure*}
    \centering
\begin{tcolorbox}[colback=gray!5!white, colframe=black!75!white, fonttitle=\bfseries, width=0.9\textwidth]
\begin{lstlisting}[language=Python, basicstyle=\ttfamily\footnotesize,
    keywordstyle=\ttfamily\footnotesize,
    commentstyle=\ttfamily\footnotesize,
    stringstyle=\ttfamily\footnotesize,
            breaklines=true, 
            breakatwhitespace=true, 
    showstringspaces=false,
    breakindent=0pt]
You receive information about specific objects and parts across multiple images, including their category identifiers and detailed locations as bounding boxes. Objects and parts will be represented in the '{name}_{id}' format, where '{id}' is an identifier for each object and part. The bounding boxes are represented as (x1, y1, x2, y2). These values correspond to the top-left coordinates (x1, y1) and the bottom-right coordinates (x2, y2). The category names and bounding box coordinates of objects and their parts in each image are as follows:

### Image Description Start ###
Image 1:
Height: 512
Width: 683
Objects:
- Object "table_101 [584, 281, 673, 338]"
- Part "drawer_10101 [609, 304, 662, 323]"
- Object "sofa_102 [0, 252, 114, 466]"
- Part "arm_10201 [0, 258, 111, 409]"
- Part "back_10202 [1, 252, 29, 323]"
- Object "bed_103 [318, 199, 683, 512]"
- Part "footboard_10301 [361, 305, 682, 463]"
- Part "headboard_10302 [355, 199, 573, 271]"
- Part "leg_10303 [363, 453, 682, 509]"
Image 2:
Height: 400
Width: 600
Objects:
- Object "bed_201 [0, 174, 405, 399]"
- Part "headboard_20101 [2, 174, 220, 272]"
- Object "bench_202 [248, 273, 425, 399]"
- Part "leg_20201 [380, 304, 422, 347]"

[TRUNCATED]
### Image Description End ###
\end{lstlisting}
\end{tcolorbox}
\vspace{-0.3cm}
    \caption{\textbf{Prompt template for [Image Description]}.}
    \label{fig:prompt_image_description}
\end{figure*}
\begin{figure*}
    \centering
    \begin{tcolorbox}[
        colback=gray!5!white,
        colframe=black!75!white,
        fonttitle=\bfseries,
        width=0.9\textwidth,
        boxsep=1pt,
        top=0pt,
        bottom=0pt,
        left=0pt,
        right=0pt,
    ]
\begin{lstlisting}[language=Python, basicstyle=\ttfamily\scriptsize,
    keywordstyle=\ttfamily\scriptsize,
    commentstyle=\ttfamily\scriptsize,
    stringstyle=\ttfamily\scriptsize,
            breaklines=true, 
            breakatwhitespace=true, 
    showstringspaces=false,
    breakindent=0pt,
    aboveskip=1pt,   
    belowskip=1pt]
**Question Guidelines:**
1. Categories: Questions must address one or more of the following categories:
    a. **Real-World Contextualization:**  
        - Contextualize questions in real-world scenarios where comparative or relational reasoning is naturally required. 
        - Examples:
            - [Truncated]
    b. **Functional Suitability and Usage:**  
        - Focus on the functional role or suitability of objects for specific tasks or purposes.
        - Examples:
            - [Truncated]
    c. **Affordance and Interaction Reasoning:**  
        - Explore how objects afford certain actions or interactions, emphasizing their usability.
        - Examples:
            - [Truncated]
    d. **Implicit Comparative Reasoning:**  
        - Require implicit comparisons or evaluations rather than direct observations.
        - Examples:
            - [Truncated]
    e. **Multi-Image Consistency and Variation:**  
        - Ask about consistency or variations in object attributes or arrangements across images.
        - Examples:
            - [Truncated]
    f. **Purpose and Function Analysis:**  
        - Delve into the purpose or function of objects, considering how they contribute to a system or task.
        - Examples:
            - [Truncated]
    g. **Context-Dependent Reasoning:**  
        - Ask questions that require understanding of how object functionality depends on context or environment.
        - Examples:
            - [Truncated]
2. Avoid questions that simply ask for a list of objects or parts in the images.
3. Questions must require simultaneous reasoning across multiple images, ensuring that all answers reference all the images for comparison.
4. Do not ask a question unless you are confident in your ability to answer it. Only include questions with definite answers.
5. Ensure that the questions and answers are concise and easy to understand, avoiding overly complex or ambiguous sentences.
6. Do not mention the coordinates of a part or an object directly in the question.
7. Ask the question as if you are seeing the images, do not mention that the information is from text.
8. When mentioning an object or part in the question, use only the {name} without the {id}.
9. Ensure that the questions are naturally phrased and coherent, resembling typical human language patterns.
\end{lstlisting}
\end{tcolorbox}
\vspace{-0.3cm}
    \caption{\textbf{Prompt template for [Question Guidelines]}.}
    \label{fig:prompt_ques_guideline}
\end{figure*}
\begin{figure*}
    \centering
    \begin{tcolorbox}[
        colback=gray!5!white,
        colframe=black!75!white,
        fonttitle=\bfseries,
        width=0.9\textwidth,
        boxsep=1pt,
        top=0pt,
        bottom=0pt,
        left=0pt,
        right=0pt,
    ]
\begin{lstlisting}[language=Python, basicstyle=\ttfamily\footnotesize,
    keywordstyle=\ttfamily\footnotesize,
    commentstyle=\ttfamily\footnotesize,
    stringstyle=\ttfamily\footnotesize,
            breaklines=true, 
            breakatwhitespace=true, 
    showstringspaces=false,
    breakindent=0pt,
    aboveskip=1pt,   
    belowskip=1pt]
**Answer Guidelines**
1. Answers must compare objects and parts across images, using explicit references to the objects and parts in the correct format.
2. Do not mention that the information is provided via a text description. Always answer as if you are directly observing the images.
3. Answers must reference at least one object (and preferably a part) from each image, where relevant.
4. It is strictly prohibited to refer to objects or parts that are not provided in the image description.
5. When referring to an object or part in the answer, use the '{name}_{id}' format exactly as provided in the given information. 
6. Ensure that the answers are naturally phrased and coherent, resembling typical human language patterns.
\end{lstlisting}
\end{tcolorbox}
\vspace{-0.3cm}
    \caption{\textbf{Prompt template for [Answer Guidelines]}.}
    \label{fig:prompt_answer_guidelines}
\end{figure*}
\begin{figure*}
    \centering
    \begin{tcolorbox}[
        colback=gray!5!white,
        colframe=black!75!white,
        fonttitle=\bfseries,
        width=0.9\textwidth,
        boxsep=1pt,
        top=0pt,
        bottom=0pt,
        left=0pt,
        right=0pt,
    ]
\begin{lstlisting}[language=Python, basicstyle=\ttfamily\footnotesize,
    keywordstyle=\ttfamily\footnotesize,
    commentstyle=\ttfamily\footnotesize,
    stringstyle=\ttfamily\footnotesize,
            breaklines=true, 
            breakatwhitespace=true, 
    showstringspaces=false,
    breakindent=0pt,
    aboveskip=1pt,   
    belowskip=1pt]
**Format Requirements:**
1. Return the output as a JSON list containing dictionaries. Each dictionary must have the following keys:
    - "categories": A list of categories the question belongs to.
    - "question": The generated question as a string.
    - "answer": The corresponding answer as a string.
2. **Reference Format**: In the answer, the objects or parts must be referenced using this exact format: {object_name}_{object_id} and {part_name}_{part_id}. Do not use alternative formats such as "a part of object" or "object_1's part." These variations are strictly forbidden.
3. **Referring Across Multiple Images**: If you need to refer to an object or part that appears in multiple images, use the format {object_name}_{object_id1}_{object_id2} or {part_name}_{part_id1}_{part_id3}. Include the corresponding IDs from each image after the object or part name, separated by underscores. 
 - **Example**: If you are referring to a rocket that appears in two images with IDs 101 and 202, write `rocket_101_202`. Do not write "rockets (rocket_101 and rocket_202)" or any other variation.
4. **No Unnecessary Formatting**: Do not use any unnecessary asterisks, quotation marks, or special characters for emphasis or formatting. There should not be any asterisks or quotation marks around object names, part names, or bounding boxes.
\end{lstlisting}
\end{tcolorbox}
\vspace{-0.3cm}
    \caption{\textbf{Prompt template for [Format Requirements]}.}
    \label{fig:prompt_format_requirements}
\end{figure*}

\end{document}